\documentclass{article}

\usepackage{natbib}

\usepackage[final]{neurips_2025}

\usepackage[utf8]{inputenc} 
\usepackage[T1]{fontenc}    
\usepackage{hyperref}       
\usepackage{url}            
\usepackage{booktabs}       
\usepackage{amsfonts}       
\usepackage{nicefrac}       
\usepackage{microtype}      
\usepackage{xcolor}         
\usepackage{multirow}
\usepackage{multicol}
\usepackage{kotex}
\usepackage{amsmath}
\usepackage{amssymb}
\usepackage{graphicx} 
\usepackage{subcaption}
\usepackage{wrapfig}
\usepackage{relsize}
\usepackage{amsthm}
\usepackage{mathtools}
\definecolor{lightyellow}{RGB}{255, 255, 204}
\definecolor{softgray}{rgb}{1.0, 0.52, 0.32}
\newtheorem{theorem}{Theorem}[section]
\newtheorem{lemma}[theorem]{Lemma}
\usepackage[dvipsnames]{xcolor}
\usepackage[x11names]{xcolor}
\usepackage[table,xcdraw]{xcolor}
\definecolor{babyblue}{rgb}{0.32, 0.52, 1.0}
\usepackage{amsthm}     
\usepackage{enumitem}   

\newcommand\blfootnote[1]{%
  \begingroup
  \renewcommand\thefootnote{}%
  \NoHyper\footnote{#1}\endNoHyper%
  \addtocounter{footnote}{-1}%
  \endgroup
}

\title{On Epistemic Uncertainty of Visual Tokens for Object Hallucinations in Large Vision-Language Models}

\author{%
Hoigi Seo$^{1*}$ \quad\quad\quad Dong Un Kang$^{1*}$ \\ \textbf{Hyunjin Cho}$^1$ \quad\quad\quad \textbf{Joohoon Lee}$^2$ \quad\quad\quad \textbf{Se Young Chun}$^{1,2,3\dag}$ \\
$^1$Dept. of ECE \; $^2$IPAI \& $^3$INMC, Seoul National University, Republic of Korea\\
\texttt{\{seohoiki3215, qkrtnskfk23, jim0228, joohoonl, sychun\}@snu.ac.kr}
}

\usepackage{cleveref}

\begin{document}

\maketitle
\begin{abstract}
Large vision-language models (LVLMs), which integrate a vision encoder (VE) with a large language model, have achieved remarkable success across various tasks. However, there are still crucial challenges in LVLMs such as object hallucination, generating descriptions of objects that are not in the input image. Here, we argue that \emph{uncertain} visual tokens within the VE is a key factor that contributes to object hallucination. Our statistical analysis found that there are positive correlations between visual tokens with high \emph{epistemic} uncertainty and the occurrence of hallucinations. 
Furthermore, we show theoretically and empirically that visual tokens in early VE layers that exhibit large representation deviations under small adversarial perturbations indicate high epistemic uncertainty. 
Based on these findings, we propose a simple yet effective strategy to mitigate object hallucination by modifying the VE only. Our method comprises a proxy method with adversarial perturbations for identifying uncertain visual tokens efficiently and a method to mask these uncertain visual tokens during the self-attention process in the middle layers of the VE, suppressing their influence on visual encoding and thus alleviating hallucinations. Extensive experiments show that our method significantly reduces object hallucinations in LVLMs and can synergistically work with other prior arts.
\end{abstract}

\section{Introduction}
\label{intro}
\blfootnote{* equal contribution, $\dagger$ corresponding author.}
Large Vision-Language Models (LVLMs) have demonstrated impressive capabilities across a range of multi-modal tasks, including image captioning~\cite{agrawal2019nocaps, chen2024pali, li2023blip, lin2014microsoft, ye2023mplug}, visual-question answering (VQA)~\cite{chen2024pali, marino2019ok, wang2023image}, and multi-modal dialogue systems~\cite{dai2023instructblip, li2024llava, liu2024improved, liu2023visual, zhu2024minigpt}. Despite these notable advancements, recent studies~\cite{gunjal2024detecting, li2023evaluating, rohrbach2018object, wang2023evaluation} have reported that LVLMs are susceptible to hallucination, generating textual descriptions that do not align with the input image. In particular, object hallucination, where the model describes objects not present in the input image, significantly undermines the reliability and thus the practical utility of LVLMs~\cite{huang2024opera, jiang2024devils, kang2025see, leng2024mitigating}.

To mitigate object hallucination in LVLMs, recent works~\cite{an2024agla, che2025eazy, huang2024opera, jiang2024devils, kang2025see, leng2024mitigating, liu2024paying} have explored training-free approaches including modifying the decoding strategy of the language model~\cite{an2024agla, huang2024opera, leng2024mitigating, liu2024paying}, modulating attention mechanisms~\cite{jiang2024devils, kang2025see, liu2024paying}, or altering the input image~\cite{an2024agla} during inference. While these methods have shown effectiveness in reducing object hallucination, they often suffer limitations such as requiring multiple inferences of the large language model, which is the most computationally expensive component of LVLMs, or yielding relatively small performance gains. In contrast, approaches for object hallucination mitigation that directly target the vision encoder, a core component responsible for visual perception, have been relatively underexplored.

In this work, we investigate how visual information contributes to object hallucination in LVLMs, with a particular focus on the uncertainty of visual tokens introduced by the pre-trained vision encoder (\textit{i.e.,} epistemic uncertainty). Estimating this uncertainty typically requires intensive computation, such as Monte Carlo (MC) dropout~\cite{mukhoti2018evaluating}, which involves thousands of forward passes. To provide a more efficient alternative, we present a theoretical analysis showing that the deviation of visual token representations under adversarial perturbations is monotonically related to an upper bound of uncertainty for each visual token, particularly in the early layers of the vision encoder.
Empirically, we find that the norm of representation deviation in visual tokens caused by adversarial perturbations closely aligns with uncertainty estimates obtained via MC dropout, enabling a more efficient approximation of visual token uncertainty. Furthermore, we empirically demonstrate a strong positive correlation between visual token uncertainty and the occurrence of object hallucination of LVLMs.

Motivated by this observation, we propose a simple yet effective method to mitigate hallucination by intervening only in the vision encoder during inference. Specifically, we first identify \textit{uncertain visual tokens}, defined as those whose representations exhibit significant deviation under PGD-based adversarial perturbations~\cite{madry2018towards} which reflect high epistemic uncertainty. We then suppress their influence by masking these uncertain tokens in the self-attention layers of intermediate vision encoder blocks. This approach reduces the model’s dependence on uncertain visual features while preserving the global semantic structure of the image representation.

Extensive experiments demonstrate that our method effectively reduces object hallucination on benchmark datasets such as CHAIR~\cite{rohrbach2018object}, POPE~\cite{li2023evaluating}, and AMBER~\cite{wang2023llm}. We validate our approach across a range of LVLM architectures~\cite{chen2023shikra, liu2024improved, zhu2024minigpt}, incorporating diverse vision encoders, language models, and training regimes to ensure generalizability. Notably, because our method exclusively modifies the vision encoder, it can be seamlessly combined with existing methods that adjust decoding strategies or attention mechanisms within the language model.

Our contribution can be summarized as follows.
\begin{itemize}
    \item We theoretically and empirically demonstrate that the visual tokens exhibiting the representation deviations under adversarial perturbations indicate upper bound of epistemic uncertainty, which is strongly correlated with object hallucination in LVLMs.
    \item Motivated by this insight, we propose an efficient and effective method that mitigates hallucination by identifying uncertain visual tokens via adversarial perturbation and masking them in the self-attention layers of intermediate vision encoder blocks.
    \item Our method is validated across multiple benchmarks and LVLM architectures, and is easily compatible with existing mitigation methods, enabling synergistic gains in performance.
\end{itemize}

\section{Related Works}\label{rel_work}
\paragraph{Large Vision-Language Models.}
Large Vision-Language Models (LVLMs) integrate visual and textual inputs for multi-modal reasoning and generation. Modern LVLMs typically consist of a vision encoder~\cite{dosovitskiy2020vit, fang2023eva, ilharco_gabriel_2021_5143773,radford2021learning, zhai2023sigmoid}, a connector, and a language model~\cite{qwen, vicuna2023, touvron2023llama, qwen2}. Some use linear projections to align visual features with the language embedding space~\cite{chen2023shikra, liu2023visual}, while others adopt Q-Former modules~\cite{dai2023instructblip, li2023blip, zhu2024minigpt} that use learnable queries to extract and compress visual information. Despite their remarkable performance on multi-modal tasks, LVLMs exhibit hallucination, generating output misaligned with visual content, raising concerns about their reliability in real-world usage.

\paragraph{Mitigating hallucinations in LVLMs.}
Hallucination in LVLMs refers to the phenomenon in which the output contradicts the visual input by fabricating visual information~\cite{bai2024hallucination, liu2024survey}. Mitigation strategies fall into training-based and training-free categories. Training-based methods optimize the LVLMs~\cite{jiang2024hallucination, yue2024less} or incorporate auxiliary modules for output guidance~\cite{duan2025truthprint, lyu2024alleviating, zhou2024analyzing}, but are often computationally expensive. Training-free approaches modify logits of language models to suppress hallucination-prone text tokens~\cite{an2024agla, huang2024opera, huo2024self, leng2024mitigating, li2025mitigating, liu2024paying, wang2024mllm, zhu2024ibd}, adjust attention process~\cite{jiang2024devils, kang2025see, li2025mitigating, liu2024paying, xie2025tarac}, or modify inputs~\cite{an2024agla, mao2025through, zhang2025mllms}. However, the approaches overlook deficiencies in the vision encoder. We instead propose an orthogonal and training-free strategy: leverage adversarial attacks to identify uncertain visual tokens and suppress them, complementing language-level approaches.

\paragraph{Adversarial attack on LVLMs.}
Adversarial attack~\cite{goodfellow2015explaining, madry2018towards, szegedy2013intriguing} introduces imperceptible perturbations in images to induce incorrect predictions by a model. While early efforts focused on tasks such as classification~\cite{goodfellow2015explaining, madry2018deep} and object detection~\cite{cai2023ensemble, liang2021parallel, xie2017adversarial}, recent work has extended attacks to LVLMs~\cite{carlini2023aligned, qi2024visual, schlarmann2023adversarial, shayegani2024jailbreak, zhao2023evaluating} to improve the robustness of the models. In image-targeted attacks, where input is in a discrete pixel space, Projected Gradient Descent (PGD)~\cite{madry2018towards} remains a dominant strategy due to its effectiveness. The optimization process of PGD is formalized as follows.
\begin{equation}\label{eq:pgd}
    \quad \hat{x}_{i+1} = \mathlarger{\Pi}\Big(\hat{x}_i+\alpha\cdot\text{sign}\big(\nabla_{\hat{x}_i} \mathcal{L}(F(\hat{x}_i), F(x))\big)\Big),
\end{equation}
where $\alpha\in \mathbb{N}$ is the learning rate, $F$ is a target neural network, $x$ is the original image, $\hat{x}_i$ denotes the perturbed image at iteration $i$, and $\Pi$ projects onto the constraint set $\|\hat{x}_{i+1}-x\|_{\infty} \leq k$. LVLMs show strong multi-modal capabilities but remain vulnerable to adversarial attacks~\cite{carlini2023aligned, qi2024visual, wang2024break, zhao2023evaluating}, which can target the entire model~\cite{carlini2023aligned, qi2024visual, zhao2023evaluating} or specifically the vision encoder~\cite{wang2024break}.

\section{Method}
In this section, we present our approach for identifying uncertain visual tokens within the vision encoder using adversarial perturbations, as detailed in Sec.\ref{sec:adv_reveal}. We demonstrate that these tokens significantly contribute to object hallucination in LVLMs through statistical analysis. Based on these findings, we propose a masking strategy within the vision encoder to suppress the influence of uncertain tokens, resulting in a notable reduction in hallucinations, as described in Sec.\ref{sec:masking}.

\subsection{Adversarial Attack to Vision Encoder Reveals Uncertain Visual Tokens}\label{sec:adv_reveal}
\begin{figure}
    \centering
    \begin{subfigure}{0.35\linewidth}
        \includegraphics[width=\linewidth]{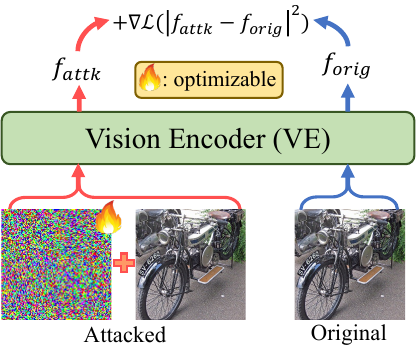}
        \caption{Adversarial attack process}
        \label{fig:attack}   
    \end{subfigure}
    \hspace{1em}
    \begin{subfigure}{0.55\linewidth}
        \includegraphics[width=\linewidth]{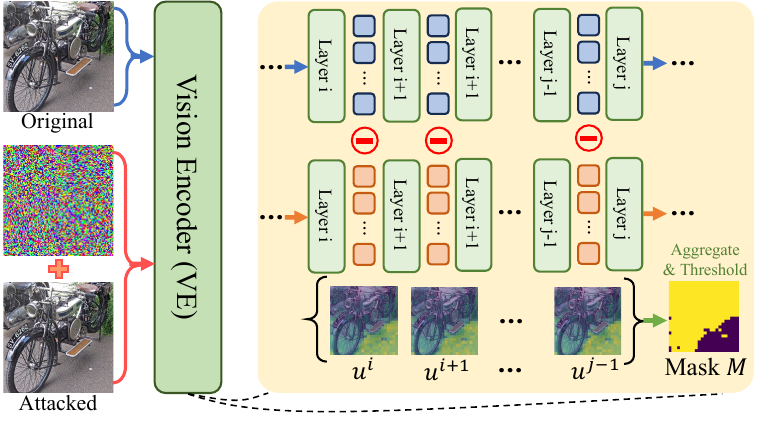}
        \caption{Uncertainty mask generation process}
        \label{fig:maskgen}
    \end{subfigure}
    \caption{\textbf{Overall illustration of the adversarial attack and uncertainty mask generation process.} \textbf{(a)} The original image is processed by the vision encoder (VE) to obtain features $f_{\text{orig}}$. An adversarial image is created by adding optimizable noise, which is then encoded to produce $f_{\text{attk}}$. The noise is optimized using Projected Gradient Descent (PGD) to maximize the mean squared error between $f_{\text{orig}}$ and $f_{\text{attk}}$, as described in Eq.~\ref{eq:pgd}. \textbf{(b)} From layers $i$ to $j-1$, we extract feature sets $\mathcal{F}_{\text{orig}}=\{f_{\text{orig}}^{i},\dots,f_{\text{orig}}^{j-1}\}$ and $\mathcal{F}_{\text{attk}}=\{f_{\text{attk}}^{i},\dots,f_{\text{attk}}^{j-1}\}$. The norm differences of corresponding features form layer-wise uncertainty maps $\mathcal{U}=\{u^{i},\dots,u^{j-1}\}$. These maps are min-max normalized, aggregated, and standardized to produce the final binary uncertainty mask $M$ using a threshold $\sigma_{\text{th}}$.}
    \label{fig:attack_and_maskgen}
\end{figure}
\subsubsection{Efficient uncertainty approximation of visual token with adversarial attack}\label{subsubsec:adv_theory_method}
Estimating uncertainty induced by deep neural networks (\textit{i.e.} epistemic uncertainty) is commonly approached by approximating Bayesian inference using Monte Carlo (MC) dropout~\cite{laves2020calibration, mukhoti2018evaluating}. However, the approximation process introduces substantial overhead as a result of thousands of forward passes. In this work, we find that the epistemic uncertainty of individual visual tokens differs from each other, as perceived by the vision encoder for a given image, and their upper bound can be efficiently estimated via adversarial attacks. To support this claim, we first introduce the following lemma.

\begin{lemma}[Approximate local Gaussianity under small perturbation] 
Let $f = \{f_t\}_{t=1}^L$ be a smooth $L$-layer neural network parameterized by $\theta$. For an input $x \in \mathbb{R}^{N \times 3}$, define the hidden state at layer $t$ as $z^{(t)} = f_t \circ \cdots \circ f_1(x)$. For a perturbed input $x+\epsilon$, with $\|\epsilon\|_\infty \le k$ for sufficiently small $k>0$, define the perturbed hidden state as $Z^{(t)} = f_t \circ \cdots \circ f_1(x+\epsilon)$. 
Then, under the assumption that the perturbation is small and $f \in C^2$, $Z^{(t)}$ can be locally approximated by a Gaussian centered at $z^{(t)}$, with a third-order remainder in the log-density.
\end{lemma}

The proof of Lemma~\ref{lem:local_gaussianity} can be found in the Appendix Sec.~\ref{append_proof:local_gaussianity}. The lemma implies that the hidden states exhibit Gaussianity under small perturbation, which allows us to prove the following theorem.

\begin{theorem}[Upper bound of differential entropy increases as hidden state deviation increases under adversarial attack]
Let \( x \) be an input image, and let \( \epsilon \) be a small adversarial perturbation. Define the perturbed input as \( X := x + \epsilon \). Let \( f = \{f_t\}_{t=1}^L \) be a smooth \( L \)-block transformer that processes a sequence of \( N \) input tokens. Let \( z^{(t)} := f_t \circ \cdots \circ f_1(x) \in \mathbb{R}^{N \times d} \) and \( Z^{(t)} := f_t \circ \cdots \circ f_1(X) \in \mathbb{R}^{N \times d} \) be the hidden states at layer \( t \) for the clean and perturbed inputs, respectively. Denote the \( i \)-th token representation at layer \( t \) as \( z_i^{(t)} \in \mathbb{R}^{d} \) and \( Z_i^{(t)} \in \mathbb{R}^{d} \).
If \( Z_i^{(t)} \) changes smoothly with small \( \epsilon \), then the upper bound of the differential entropy of \( Z_i^{(t)} \) increases as \( \mathbb{E}_{\epsilon}[\|Z_i^{(t)} - z_i^{(t)}\|_2^2] \) increases.
\end{theorem}

The proof of Theorem~\ref{thm:entropy}, provided in Appendix Sec.~\ref{append_proof:thm}, shows that under adversarial attack, the norm of hidden state deviation efficiently approximates the upper bound of visual token's entropy.

Leveraging this insight from Theorem~\ref{thm:entropy}, we aim to obtain a mask that identifies uncertain visual tokens with an adversarial attack. Specifically, given an image $x$ and a vision encoder $F_{V}$, we first obtain the feature $f_{\text{orig.}}=F_{V}(x) \in \mathbb{R}^{N \times d}$, where $N$ denotes the number of image tokens. We then generate an adversarially perturbed image $\hat{x}_0$ by adding small noise $\epsilon$ to $x$ such that $\|\epsilon\|_{\infty} \leq k$. We then extract feature of perturbed image with $f_{\text{attk.}}=F_V(\hat{x}_0)$. We define the adversarial objective as the mean squared error between $f_{\text{orig.}}$ and $f_{\text{attk.}}$, and optimize $\epsilon$ with PGD for $I$ iterations as specified in Eq.~\ref{eq:pgd} to obtain the final attacked image $\hat{x}\coloneqq\hat{x}_I$. This attack process is illustrated in Fig.~\ref{fig:attack}.
\begin{figure}
  \centering
  \vspace{-1em}
  \includegraphics[width=\linewidth]{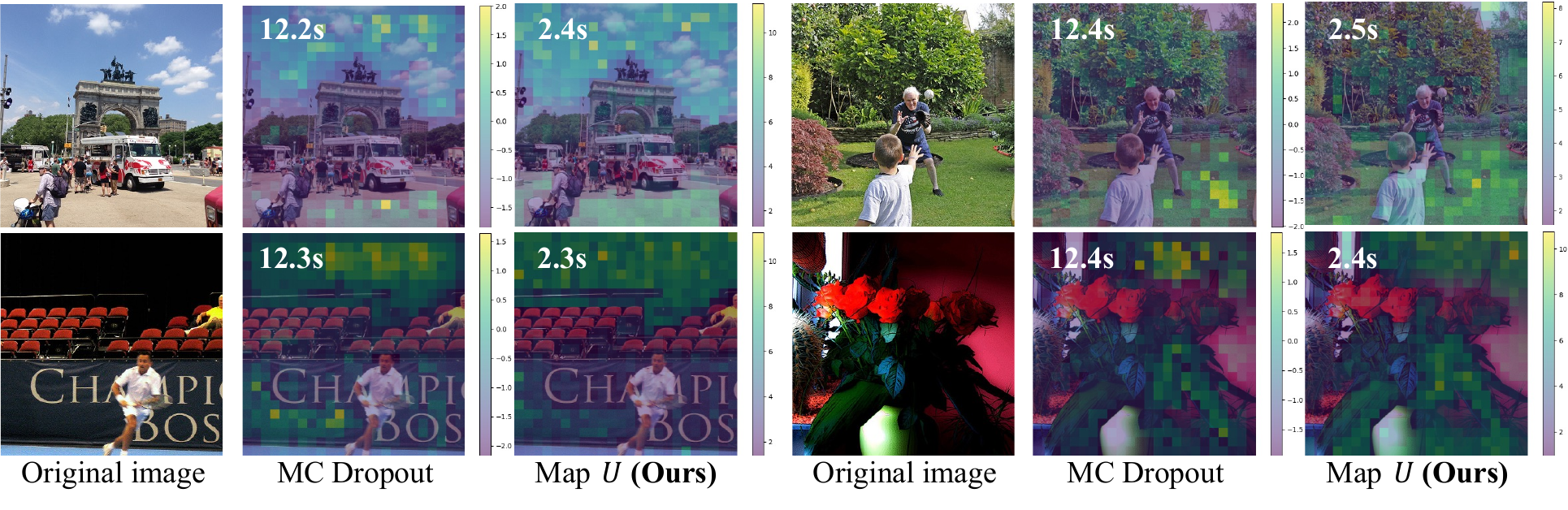}
  \caption{\textbf{Visual comparison of estimated uncertainty from MC dropout~\cite{mukhoti2018evaluating} and our method.} Our uncertainty map $U$ identifies uncertain regions similar to the uncertainty map obtained via MC dropout. MC dropout was applied to the residuals of the LLaVA-1.5 vision encoder with a dropout rate of $p=0.5$ and the variance of each token was estimated over 1,000 forward passes. For the adversarial attack, we applied 100 iterations of PGD with $k=3$. The MC-based uncertainty values were log-scaled for visualization clarity. The runtime for each example is shown in the top-left corner.}
  \label{fig:mc_vs_ours}
\end{figure}

Next, we extract the hidden states from each layer of the $F_{V}$ within the consecutive layer index set $\mathcal{S} = \{i, \dots, j-1\}$ for both the original image $x$ and the perturbed image $\hat{x}$. This results in two hidden states sets: $\mathcal{F}_{\text{orig}} = \{f_{\text{orig}}^i, \dots, f_{\text{orig}}^{j-1}\}$ from $x$ and $\mathcal{F}_{\text{attk}} = \{f_{\text{attk}}^i, \dots, f_{\text{attk}}^{j-1}\}$ from $\hat{x}$. For each layer $l \in \mathcal{S}$, we compute the norm of deviation between the corresponding hidden states defined as $u^l = \|f_{\text{attk}}^l - f_{\text{orig}}^l\|_2$, resulting in a set of layer-wise uncertainty maps $\mathcal{U} = \{u^i, \dots, u^{j-1} |\forall l\in\mathcal{S} \}$.

We then aggregate the layer-wise uncertainty maps in $\mathcal{U}$ to produce the aggregated uncertainty map $U$ by applying min-max normalization to each $u^l$ and averaging across layers, as defined below:
\begin{equation}\label{eq:uncertain_map}
U = \frac{1}{j - i} \sum_{l = i}^{j-1} \frac{u^l - u^l_{\text{min}}}{u^l_{\text{max}} - u^l_{\text{min}}}.
\end{equation}
Finally, we normalize the uncertainty map $U$ using its mean $\mu_{U}$ and standard deviation $\sigma_{U}$, and binarize it with a threshold parameter $\sigma_{\text{th}}$ to obtain the binary uncertainty mask $M$ as follows:
\begin{equation}\label{eq:mask_gen}
    M = 1-\frac{1}{2}\bigg\lfloor\text{sign}\bigg(\Big(\frac{U-\mu_U}{\sigma_U}\Big)-\sigma_{\text{th}} \bigg) +1 \bigg\rfloor \in \{0, 1\}^{N}.
\end{equation}

Here, a value of 0 in the mask $M$ indicates an ``uncertain'' visual token, while 1 denotes a relatively ``certain'' one.
Figure~\ref{fig:maskgen} illustrates the mask generation process, and examples of $M$ are shown in Appendix Sec.\ref{appendix:Mask_addi_examples}.
In Sec~\ref{sec:masking}, we describe how $M$ is used to mitigate object hallucination.

\subsubsection{Empirical study on extracting uncertainty with adversarial attack}\label{subsubsec:adv_emp}
\begin{wrapfigure}{r}{0.5\linewidth}
  \centering
  \vspace{-1.5em}
  \includegraphics[width=\linewidth]{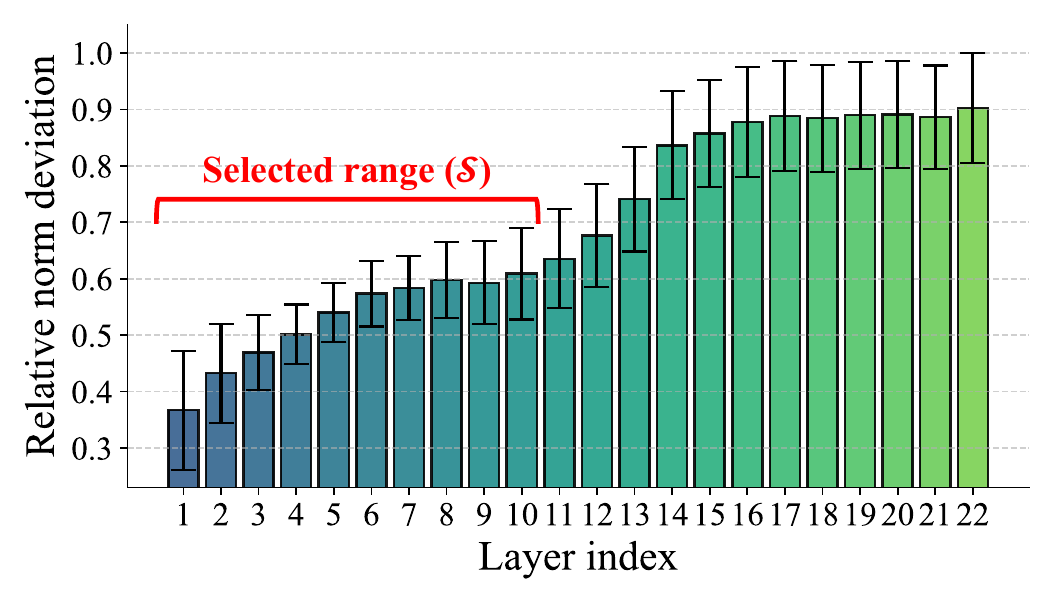}
  \caption{\textbf{Relative deviation between attacked and original features.} We used 500 images from the MS-COCO~\cite{lin2014microsoft} with LLaVA-1.5 vision encoder~\cite{liu2024improved}. Perturbations introduced through the vision encoder remain minimal in early layers but intensify in later ones. We extract the mask from early layers where feature deviations are comparatively small. Error bars denote the $2\sigma$ range.}
  \label{fig:layerwise_diff}
  \vspace{-1.0em}
\end{wrapfigure}

\paragraph{Comparison with uncertainty via MC dropout.} We compare our uncertainty map $U$ with MC dropout~\cite{mukhoti2018evaluating} to assess how well $U$ approximates epistemic uncertainty. As shown in Fig.\ref{fig:mc_vs_ours}, the results indicate that $U$ closely aligns with the uncertainty estimated via MC dropout, demonstrating that $U$ serves as an efficient approximation. On average, it is approximately 5 times faster than MC dropout in practice. Additional qualitative and computational cost comparisons are provided in Appendix Sec.\ref{append_sec:mc_vs_ours}.

\paragraph{The range of layer indices set $\mathcal{S}$ of vision encoder.}
As described in Sec.~\ref{subsubsec:adv_theory_method}, we extract hidden states from the consecutive layer index set $\mathcal{S}$. Our Lemma~\ref{lem:local_gaussianity} and Theorem~\ref{thm:entropy} rely on the assumption that adversarially induced norm of visual feature deviations are small, requiring that perturbations remain limited. Fig.\ref{fig:layerwise_diff} shows these deviations are minor in early layers but increase in later ones. To ensure consistency with both the theoretical assumptions and empirical observations, we construct $\mathcal{S}$ from early layers of vision encoder. Further analyses on $\mathcal{S}$, provided in Sec.\ref{subsec:ablations}, additionally support this theoretical and empirical alignment.

\subsection{Mitigating Object Hallucination of LVLMs via Uncertain Visual Tokens}\label{sec:masking}
Building on the identification of uncertain visual tokens through adversarial perturbations in Sec.~\ref{sec:adv_reveal}, we now investigate how these tokens can be utilized to reduce object hallucination in LVLMs. 

\subsubsection{Uncertain visual tokens contribute to object hallucination}\label{subsubsec:uncertain_hallu}
\begin{wrapfigure}{r}{0.5\linewidth}
  \centering
  \vspace{-2em}
  \includegraphics[width=\linewidth]{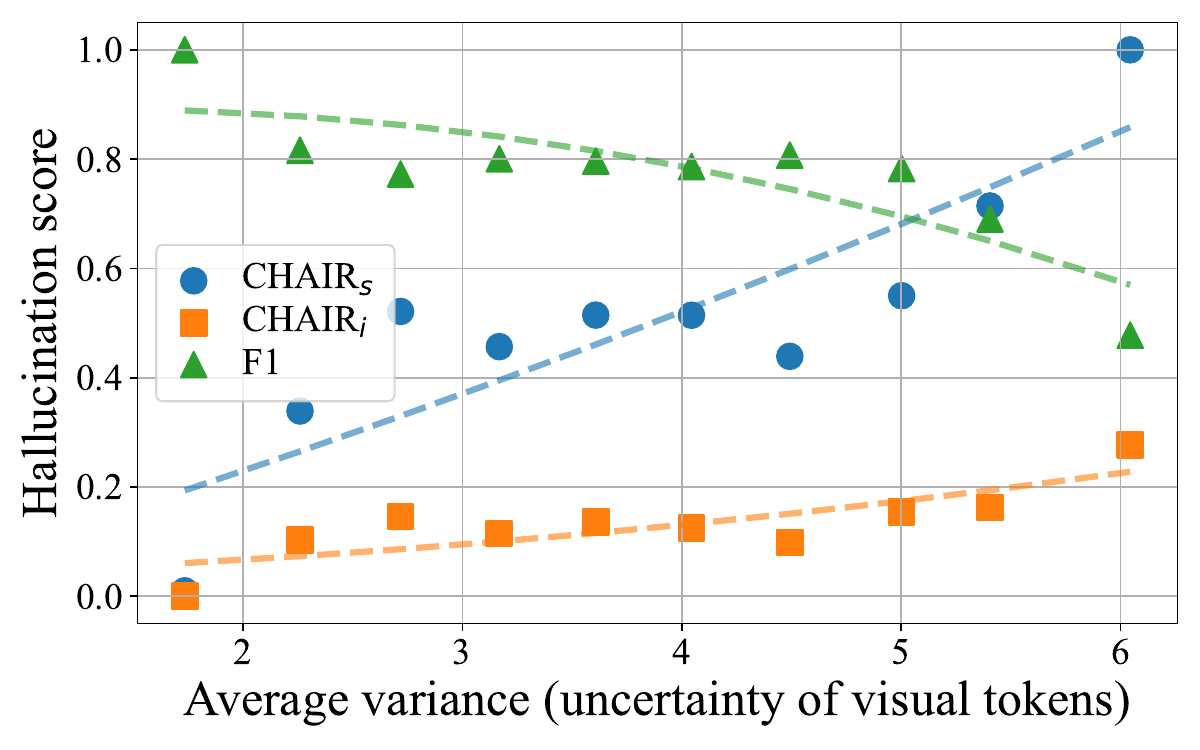}
  \caption{\textbf{Relationship between uncertain visual tokens and object hallucination.} The $x$-axis represents the average variance within each bin, while the $y$-axis shows the corresponding metric scores. The results indicate that higher uncertainty is associated with more object hallucination, with $p$-value~$<0.05$. The trend line was fitted with quadratic function. Note that \textit{higher values} of CHAIR$_s$ and CHAIR$_i$, and \textit{lower} F1 score indicate more \textit{severe object hallucinations}.}
  \label{fig:variance_vs_metric}
  \vspace{-1.0em}
\end{wrapfigure}

To assess the practical relevance of uncertain visual tokens in object hallucination, we conducted a preliminary study using LLaVA-1.5-7B~\cite{liu2024improved} on 1,000 randomly sampled images from MS-COCO~\cite{lin2014microsoft}. 
We estimate the uncertainty map of each visual token via Monte Carlo (MC) dropout, by computing the token-level variance.
Using Eq.~\ref{eq:mask_gen} and a threshold of $\sigma_{\text{th}}=1$, we generate an uncertainty mask and calculate the average variance across the uncertain visual tokens in each image. The resulting averaged variances are grouped into 10 bins, and for each bin, we evaluate the severity of hallucination using the CHAIR~\cite{rohrbach2018object} benchmark. 

The experimental results are presented in Fig.~\ref{fig:variance_vs_metric}. Fig.~\ref{fig:variance_vs_metric} shows that higher average uncertainty of visual tokens corresponds to more severe object hallucination. To statistically validate this monotonic trend, we performed Spearman’s rank correlation analysis between the average uncertainty (measured via token-level variance) and each hallucination metric. The resulting correlation coefficients were $\rho=0.794$ ($p\text{-value}=0.006$) for CHAIR$_s$, $\rho=0.733$ ($p\text{-value}=0.016$) for CHAIR$_i$, and $\rho=-0.745$ ($p\text{-value}=0.013$) for the F1 score, all statistically significant at $p\text{-value}<0.05$, and indicating \textit{strong monotonic relationships}~\cite{rovai2013social} ($|\rho| > 0.7$).  Through this statistical analysis, we confirm that uncertain visual tokens contribute to hallucination of LVLMs.

\subsubsection{Masking uncertain visual tokens for training-free hallucination mitigation}~\label{subsubsec:masking_method}
Building on the findings in Sec.~\ref{subsubsec:uncertain_hallu}, we propose a method to reduce object hallucination by leveraging the uncertainty mask $M$, which highlights uncertain visual tokens identified through adversarial perturbation. Instead of completely removing these tokens, we attenuate their influence during the self-attention process in the intermediate layers of the vision encoder. The intermediate layers of vision encoder was selected on the basis of empirical evidence that indicates its superior effectiveness in mitigating object hallucination. 

Let \( Q, K, V \in \mathbb{R}^{N \times d'} \) be the query, key, and value matrices in a self-attention layer, where \( N \) denotes the number of tokens and \( d' \) the dimensionality of the hidden states within self-attention process. Let \( M \in \{0,1\}^N \) be the binary uncertainty mask obtained from Eq.~\ref{eq:mask_gen}. Then, our masking strategy modifies the attention computation as follows:
\begin{equation}
\text{Attention}(Q, K, V, M) = \left( \text{Softmax}\left( \frac{QK^\top}{\sqrt{d'}} \right)V \right) \odot M
\end{equation}
\begin{wrapfigure}{r}{0.6\linewidth}
  \centering
  \vspace{-1.5em}
  \includegraphics[width=\linewidth]{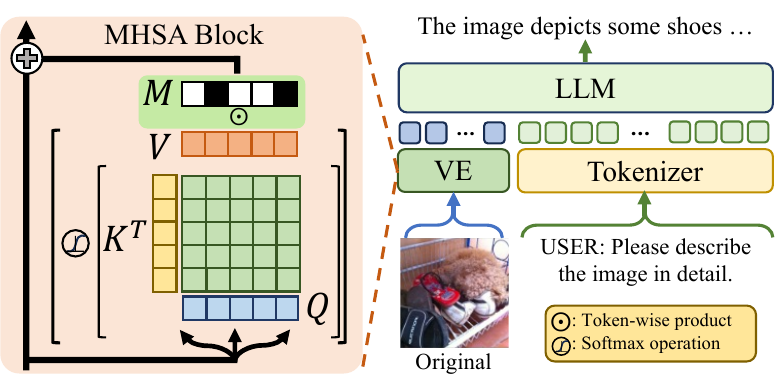}
  \caption{\textbf{Illustration of our attention masking method during inference.} In the intermediate multi-head self-attention layers of the vision encoder, we apply a binary uncertainty mask $M$ to the attention outputs. This token-wise masking reduces the influence of uncertain visual tokens, while preserving the meaningful visual representation.}
  \label{fig:masking_strat}
  \vspace{-1.0em}
\end{wrapfigure}
Here, $\odot$ denotes token-wise multiplication. This operation reduces the influence of uncertain tokens in the attention output while keeping the attention weights and other token interactions intact. Since the masking is applied within the residual connection structure, the model retains stable and meaningful visual representations while suppressing the contribution from uncertain visual tokens. We illustrate this masking strategy within the self-attention process of the vision encoder within LVLMs in Fig.~\ref{fig:masking_strat}.

Compared to masking strategies applied at the input or output of the vision encoder, intervening during self-attention computation in intermediate layers of the vision encoder offers a more balanced approach to reduce the effect of uncertain tokens without discarding potentially useful visual information, as shown in the ablation study in Sec.~\ref{subsec:ablations}.

\subsubsection{Does our method reduce uncertainty and mitigate object hallucination? Yes.}\label{subsubsec:does}
\begin{wrapfigure}{r}{0.5\linewidth}
  \centering
  \vspace{-1.5em}
  \includegraphics[width=\linewidth]{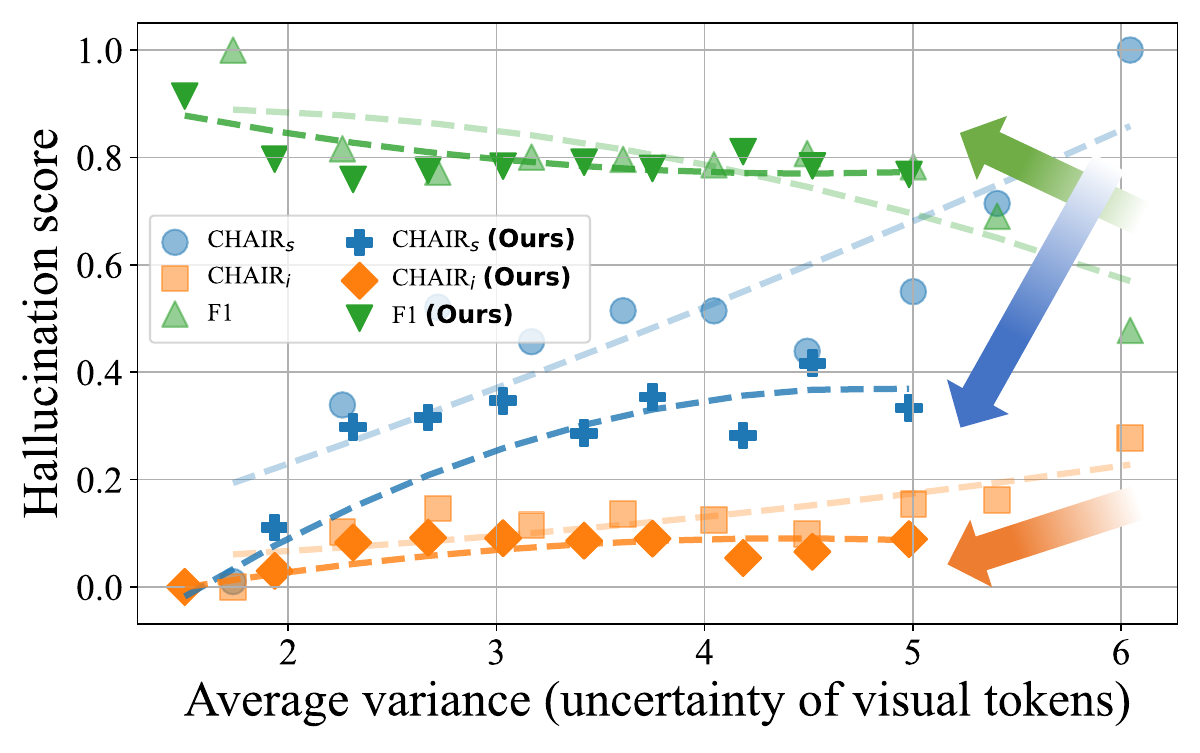}
  \vspace{-2em}
  \caption{\textbf{Impact of the proposed masking strategy on visual token uncertainty.} Average token-level variance estimated via MC dropout decreases after applying our method, indicating reduced uncertainty. This reduction correlates with improved performance on object hallucination metrics. The trend line was fitted with quadratic function.}
  \label{fig:variance_vs_metric_ours}
  \vspace{-1.5em}
\end{wrapfigure}

Based on the relationship between uncertainty of visual tokens and object hallucination discussed in Sec.~\ref{subsubsec:uncertain_hallu}, we mitigate object hallucination using the method introduced in Sec.~\ref{subsubsec:masking_method}. To evaluate the effectiveness of our method in reducing visual token uncertainty, we conducted the same experiment as shown in Fig.~\ref{fig:variance_vs_metric}.

The results in Fig.~\ref{fig:variance_vs_metric_ours} show that the average variance in the bin with the highest uncertainty decreases from 6.04 to 4.98, CHAIR$_s$ drops from 1.00 to 0.33, CHAIR$_i$ from 0.27 to 0.09, and the F1 score increases from 0.47 to 0.77. To evaluate statistical significance, we performed the Wilcoxon signed rank test~\cite{conover1999practical}, which confirmed significant reductions in average variance ($p = 0.002$), CHAIR$_s$ ($p = 0.002$), and CHAIR$_i$ ($p = 0.004$), all statistically significant at $p < 0.05$. The F1 score was preserved. These results demonstrate that the uncertainty of visual tokens contributes to object hallucination, and that our method effectively suppresses this uncertainty, thereby mitigating hallucinations in LVLMs.

\section{Experiments}
\subsection{Experimental Setup}
\paragraph{Baselines and implementation details.}~\label{subsec:baselines_imple_details}
We evaluate our method on diverse LVLMs differing in size, architecture, and vision encoders: LLaVA-1.5-7B~\cite{liu2024improved} with \texttt{CLIP-L/336px}~\cite{radford2021learning}, Shikra-7B~\cite{chen2023shikra} with \texttt{CLIP-L}, and MiniGPT-4 using \texttt{EVACLIP-g}~\cite{sun2023eva} and a Q-Former for image-text alignment. To assess compatibility, we integrate our method with hallucination mitigation methods including OPERA\cite{huang2024opera}, VCD~\cite{leng2024mitigating}, PAI~\cite{liu2024paying}, and Devils~\cite{jiang2024devils}. Adversarial attacks are run with $k=3$ and 200 PGD steps. The uncertainty masks $M$ are extracted from layers $\mathcal{S} = \{1, \dots, 10\}$ of the vision encoder, with masking applied to layers 13–17 for LLaVA-1.5 and Shikra, and 9–16 for MiniGPT-4. $\sigma_{\text{th}}$s are tuned per baseline-method pair. Further details are provided in Appendix Sec.~\ref{append_sec:imple}, and ~\ref{append_sec:baselines}.

\paragraph{Benchmarks.}
To measure object hallucination, we use three standard benchmarks. CHAIR~\cite{rohrbach2018object} measures sentence-level ($C_s \coloneqq \text{CHAIR}_s$) and instance-level ($C_i \coloneqq \text{CHAIR}_i$) hallucinations from generated descriptions with 500 prompts randomly sampled from COCO~\cite{lin2014microsoft}:
\begin{equation}\label{eq:chair}
    \text{CHAIR}_s=\frac{|\{\text{sentences with hallucinated objects}\}|}{|\{\text{all sentences}\}|},\quad
    \text{CHAIR}_i=\frac{|\{\text{hallucinated objects}\}|}{|\{\text{all mentioned objects}\}|}.
\end{equation}
POPE~\cite{li2023evaluating} evaluates hallucination through binary object presence queries across three splits (Random, Popular, Adversarial), total 9,000 prompts, reporting accuracy. AMBER~\cite{wang2023llm} comprehensively evaluates hallucination in two settings: a generative approach (Gen.) that assesses hallucination through image captioning and a discriminative approach (Disc.) that uses yes/no choices. To measure object hallucination with AMBER, we adopted the full set of Gen. and the ‘Existence’ subset of Disc., conducting with a total of 5,928 prompts.
See Appendix~\ref{append_sec:benchmark} and~\ref{append_sec:quanti} for more details and results. 
\begin{table}[!t]
  \centering
  \caption{\textbf{Quantitative results on CHAIR and POPE benchmarks.} Object hallucination is evaluated on the CHAIR and POPE benchmarks using three LVLMs and five decoding strategies, both with and without our method. POPE results are reported on three splits: Random, Popular, and Adversarial. The maximum token length is set to 512. $\Delta$\% denotes the relative difference in performance. $\uparrow/\downarrow$ indicate that higher/lower values are better. We highlight the best scores in \textbf{bold}.}
  \label{tab:main_quanti}
  \renewcommand{\arraystretch}{1.2}
  \setlength{\tabcolsep}{5pt}
  \resizebox{\textwidth}{!}{
  \begin{tabular}{cl|ccc|ccc|ccc|ccc|ccc}
    \toprule
     & \multirow{2}{*}{Method} & \multicolumn{3}{c}{Greedy} & \multicolumn{3}{c}{OPERA} & \multicolumn{3}{c}{VCD} & \multicolumn{3}{c}{PAI} & \multicolumn{3}{c}{Devils} \\
    \cmidrule(lr){3-5} \cmidrule(lr){6-8} \cmidrule(lr){9-11} \cmidrule(lr){12-14} \cmidrule(lr){15-17}
    & & Orig. & +Ours & $\Delta$\% & Orig. & +Ours& $\Delta$\% & Orig. & +Ours& $\Delta$\% & Orig. & +Ours& $\Delta$\% & Orig. & +Ours& $\Delta$\% \\
    \midrule
    \multirow{6}{*}{\rotatebox{90}{LLaVA-1.5-7B}} & $C_s$ $\downarrow$ & 
    47.4 & \cellcolor{lightyellow}29.2 & {\small\textcolor{babyblue}{$\downarrow$38.4\%}} &
    47.8 & \cellcolor{lightyellow}29.4 & {\small\textcolor{babyblue}{$\downarrow$38.5\%}} &
    53.8 & \cellcolor{lightyellow}35.2 & {\small\textcolor{babyblue}{$\downarrow$34.6\%}} &
    33.2 & \cellcolor{lightyellow}26.0 & {\small\textcolor{babyblue}{$\downarrow$21.7\%}} &
    27.0 & \cellcolor{lightyellow}\textbf{23.0} & {\small\textcolor{babyblue}{$\downarrow$14.8\%}} \\
    & $C_i$ $\downarrow$ &
    12.2 & \cellcolor{lightyellow}9.3 & {\small\textcolor{babyblue}{$\downarrow$23.8\%}} &
    12.8 & \cellcolor{lightyellow}9.5 & {\small\textcolor{babyblue}{$\downarrow$25.8\%}} &
    15.2 & \cellcolor{lightyellow}10.7 & {\small\textcolor{babyblue}{$\downarrow$29.6\%}} &
    8.5  & \cellcolor{lightyellow}7.9 & {\small\textcolor{babyblue}{$\downarrow$7.1\%}} &
    6.6  & \cellcolor{lightyellow}\textbf{5.6} & {\small\textcolor{babyblue}{$\downarrow$15.2\%}} \\
    & F1 $\uparrow$ &
    77.9 & \cellcolor{lightyellow}78.2 & {\small\textcolor{babyblue}{$\uparrow$0.4\%}} & 
    77.7 & \cellcolor{lightyellow}\textbf{78.4} & {\small\textcolor{babyblue}{$\uparrow$0.9\%}} &
    75.2 & \cellcolor{lightyellow}75.2 & {\small\textcolor{babyblue}{$\uparrow$0.0\%}} &
    78.3 & \cellcolor{lightyellow}77.2 & {\small\textcolor{softgray}{$\downarrow$1.4\%}} &
    78.3 & \cellcolor{lightyellow}78.0 & {\small\textcolor{softgray}{$\downarrow$0.4\%}}  \\
    \cmidrule(lr){2-17}
    & Rand. $\uparrow$ & 
    89.3 & \cellcolor{lightyellow}89.3 & {\small\textcolor{babyblue}{$\uparrow$0.0\%}} &
    89.2 & \cellcolor{lightyellow}88.6 & {\small\textcolor{softgray}{$\downarrow$0.7\%}} &
    84.6 & \cellcolor{lightyellow}86.2 & {\small\textcolor{babyblue}{$\uparrow$1.9\%}} &
    89.4 & \cellcolor{lightyellow}89.2 & {\small\textcolor{softgray}{$\downarrow$0.2\%}} &
    89.6 & \cellcolor{lightyellow}\textbf{90.0} & {\small\textcolor{babyblue}{$\uparrow$0.4\%}} \\
    & Pop. $\uparrow$ &
    85.8 & \cellcolor{lightyellow}85.8 & {\small\textcolor{babyblue}{$\uparrow$0.0\%}} &
    85.8 & \cellcolor{lightyellow}85.2 & {\small\textcolor{softgray}{$\downarrow$0.7\%}} &
    82.4 & \cellcolor{lightyellow}82.9 & {\small\textcolor{babyblue}{$\uparrow$0.6\%}} &
    86.0 & \cellcolor{lightyellow}86.4 & {\small\textcolor{babyblue}{$\uparrow$0.5\%}} &
    86.4 & \cellcolor{lightyellow}\textbf{87.2} & {\small\textcolor{babyblue}{$\uparrow$0.9\%}} \\
    & Adv. $\uparrow$ &
    79.3 & \cellcolor{lightyellow}80.0 & {\small\textcolor{babyblue}{$\uparrow$0.9\%}} & 
    \textbf{80.3} & \cellcolor{lightyellow}79.6 & {\small\textcolor{softgray}{$\downarrow$0.9\%}} &
    77.0 & \cellcolor{lightyellow}78.1 & {\small\textcolor{babyblue}{$\uparrow$1.4\%}} &
    79.5 & \cellcolor{lightyellow}79.9 & {\small\textcolor{babyblue}{$\uparrow$0.5\%}} &
    78.6 & \cellcolor{lightyellow}79.6 & {\small\textcolor{babyblue}{$\uparrow$1.3\%}}  \\
    
    \midrule
    \multirow{6}{*}{\rotatebox{90}{Shikra-7B}}
    & $C_s$ $\downarrow$ & 
    58.0 & \cellcolor{lightyellow}43.2 & {\small\textcolor{babyblue}{$\downarrow$25.5\%}} &
    34.8 & \cellcolor{lightyellow}28.8 & {\small\textcolor{babyblue}{$\downarrow$17.2\%}} &
    56.2 & \cellcolor{lightyellow}47.2 & {\small\textcolor{babyblue}{$\downarrow$16.0\%}} &
    32.4 & \cellcolor{lightyellow}22.2 & {\small\textcolor{babyblue}{$\downarrow$31.5\%}} &
    24.4 & \cellcolor{lightyellow}\textbf{20.6} & {\small\textcolor{babyblue}{$\downarrow$15.6\%}} \\
    & $C_i$ $\downarrow$ &
    15.6 & \cellcolor{lightyellow}11.7 & {\small\textcolor{babyblue}{$\downarrow$25.0\%}} &
    11.1 & \cellcolor{lightyellow}9.6 & {\small\textcolor{babyblue}{$\downarrow$13.5\%}} &
    16.1 & \cellcolor{lightyellow}12.8 & {\small\textcolor{babyblue}{$\downarrow$20.5\%}} &
    7.8 & \cellcolor{lightyellow}\textbf{6.1} & {\small\textcolor{babyblue}{$\downarrow$21.8\%}} &
    7.6 & \cellcolor{lightyellow}6.8 & {\small\textcolor{babyblue}{$\downarrow$10.5\%}} \\
    & F1 $\uparrow$ &
    74.7 & \cellcolor{lightyellow}\textbf{76.9} & {\small\textcolor{babyblue}{$\uparrow$2.9\%}} & 
    74.2 & \cellcolor{lightyellow}74.2 & {\small\textcolor{babyblue}{$\uparrow$0.0\%}}  &
    74.4 & \cellcolor{lightyellow}75.2 & {\small\textcolor{babyblue}{$\uparrow$1.1\%}}  &
    76.7 & \cellcolor{lightyellow}75.1 & {\small\textcolor{softgray}{$\downarrow$2.1\%}}  &
    73.3 & \cellcolor{lightyellow}72.2 & {\small\textcolor{softgray}{$\downarrow$1.5\%}}  \\
    \cmidrule(lr){2-17}
    & Rand. $\uparrow$ & 
    83.2 & \cellcolor{lightyellow}85.1 & {\small\textcolor{babyblue}{$\uparrow$2.3\%}} &
    84.8 & \cellcolor{lightyellow}\textbf{85.4} & {\small\textcolor{babyblue}{$\uparrow$0.7\%}} &
    82.1 & \cellcolor{lightyellow}82.7 & {\small\textcolor{babyblue}{$\uparrow$0.7\%}} &
    83.9 & \cellcolor{lightyellow}84.0 & {\small\textcolor{babyblue}{$\uparrow$0.1\%}} &
    83.8 & \cellcolor{lightyellow}82.5 & {\small\textcolor{softgray}{$\downarrow$1.6\%}} \\
    & Pop. $\uparrow$ &
    82.3 & \cellcolor{lightyellow}82.6 & {\small\textcolor{babyblue}{$\uparrow$0.4\%}} &
    82.8 & \cellcolor{lightyellow}82.1 & {\small\textcolor{softgray}{$\downarrow$0.8\%}} &
    79.7 & \cellcolor{lightyellow}80.7 & {\small\textcolor{babyblue}{$\uparrow$1.3\%}} &
    \textbf{83.1} & \cellcolor{lightyellow}80.7 & {\small\textcolor{softgray}{$\downarrow$2.9\%}} &
    79.9 & \cellcolor{lightyellow}78.2 &  {\small\textcolor{softgray}{$\downarrow$2.1\%}} \\
    & Adv. $\uparrow$ &
    78.2 & \cellcolor{lightyellow}78.8 & {\small\textcolor{babyblue}{$\uparrow$0.8\%}} & 
    79.2 & \cellcolor{lightyellow}\textbf{79.7} & {\small\textcolor{babyblue}{$\uparrow$0.6\%}} &
    77.3 & \cellcolor{lightyellow}77.1 & {\small\textcolor{softgray}{$\downarrow$0.3\%}} &
    78.8 & \cellcolor{lightyellow}77.4 & {\small\textcolor{softgray}{$\downarrow$1.8\%}} &
    77.7 & \cellcolor{lightyellow}76.7 & {\small\textcolor{softgray}{$\downarrow$1.3\%}}  \\

    \midrule
    \multirow{6}{*}{\rotatebox{90}{MiniGPT-4}}
    & $C_s$ $\downarrow$ & 
    28.6 & \cellcolor{lightyellow}27.4 & {\small\textcolor{babyblue}{$\downarrow$4.2\%}} &
    23.8 & \cellcolor{lightyellow}22.6 & {\small\textcolor{babyblue}{$\downarrow$5.0\%}} &
    32.0 & \cellcolor{lightyellow}30.6 & {\small\textcolor{babyblue}{$\downarrow$4.4\%}} &
    19.6 & \cellcolor{lightyellow}\textbf{17.8} & {\small\textcolor{babyblue}{$\downarrow$9.2\%}} &
    21.6 & \cellcolor{lightyellow}20.8 & {\small\textcolor{babyblue}{$\downarrow$3.7\%}} \\
    & $C_i$ $\downarrow$ &
    8.5 & \cellcolor{lightyellow}8.3 & {\small\textcolor{babyblue}{$\downarrow$2.4\%}} &
    8.8 & \cellcolor{lightyellow}8.5 & {\small\textcolor{babyblue}{$\downarrow$3.4\%}} &
    9.7 & \cellcolor{lightyellow}9.1 & {\small\textcolor{babyblue}{$\downarrow$6.2\%}} &
    6.2 & \cellcolor{lightyellow}\textbf{6.0} & {\small\textcolor{babyblue}{$\downarrow$3.2\%}} &
    7.5 & \cellcolor{lightyellow}7.0 & {\small\textcolor{babyblue}{$\downarrow$6.7\%}} \\
    & F1 $\uparrow$ &
    71.5 & \cellcolor{lightyellow}71.3 & {\small\textcolor{softgray}{$\downarrow$0.3\%}} & 
    69.8 & \cellcolor{lightyellow}70.0 & {\small\textcolor{babyblue}{$\uparrow$0.3\%}} &
    70.2 & \cellcolor{lightyellow}71.3 & {\small\textcolor{babyblue}{$\uparrow$1.7\%}} &
    \textbf{71.7} & \cellcolor{lightyellow}\textbf{71.7} & {\small\textcolor{babyblue}{$\uparrow$0.0\%}} &
    70.1 & \cellcolor{lightyellow}70.4 & {\small\textcolor{babyblue}{$\uparrow$0.4\%}}   \\
    \cmidrule(lr){2-17}
    & Rand. $\uparrow$ & 
    \textbf{82.8} & \cellcolor{lightyellow}82.5 & {\small\textcolor{softgray}{$\downarrow$0.4\%}} &
    74.2 & \cellcolor{lightyellow}74.4 & {\small\textcolor{babyblue}{$\uparrow$0.3\%}} &
    59.2 & \cellcolor{lightyellow}59.3 & {\small\textcolor{babyblue}{$\uparrow$0.2\%}} &
    82.1 & \cellcolor{lightyellow}82.0 & {\small\textcolor{softgray}{$\downarrow$0.1\%}} &
    77.4 & \cellcolor{lightyellow}77.8 & {\small\textcolor{babyblue}{$\uparrow$0.5\%}} \\
    & Pop. $\uparrow$ &
    75.1 & \cellcolor{lightyellow}74.6 & {\small\textcolor{softgray}{$\downarrow$0.7\%}} &
    71.3 & \cellcolor{lightyellow}71.8 & {\small\textcolor{babyblue}{$\uparrow$0.7\%}} &
    54.9 & \cellcolor{lightyellow}55.0 & {\small\textcolor{babyblue}{$\uparrow$0.2\%}} &
    \textbf{75.8} & \cellcolor{lightyellow}75.2 & {\small\textcolor{softgray}{$\downarrow$0.8\%}} &
    68.4 & \cellcolor{lightyellow}68.6 & {\small\textcolor{babyblue}{$\uparrow$0.3\%}} \\
    & Adv. $\uparrow$ &
    71.8  & \cellcolor{lightyellow}71.2 & {\small\textcolor{softgray}{$\downarrow$0.8\%}} & 
    69.7 & \cellcolor{lightyellow}69.4 & {\small\textcolor{softgray}{$\downarrow$0.4\%}} &
    53.8 & \cellcolor{lightyellow}54.2 & {\small\textcolor{babyblue}{$\uparrow$1.1\%}} &
    \textbf{72.1} & \cellcolor{lightyellow}71.6 & {\small\textcolor{softgray}{$\downarrow$0.7\%}} &
    65.2 & \cellcolor{lightyellow}65.3 & {\small\textcolor{babyblue}{$\uparrow$0.2\%}}  \\
    \bottomrule
  \end{tabular}
  }
\end{table}

\begin{table}[!t]
  \centering
  \caption{\textbf{Quantitative results on AMBER benchmark for LLaVA-1.5-7B.} We evaluate object hallucination using the AMBER benchmark under various mitigation methods, including combinations with our approach. AMBER measures hallucination in generative (Gen.) and discriminative (Disc.) settings, with its score offering a comprehensive assessment across both. The maximum token length is set to 512 for generative task. $\Delta$\% denotes the relative difference in performance. }
  \label{tab:llava7b_amber}
  \renewcommand{\arraystretch}{1.2}
  \setlength{\tabcolsep}{5pt}
  \resizebox{\textwidth}{!}{
  \begin{tabular}{cl|ccc|ccc|ccc|ccc|ccc}
    \toprule
     & \multirow{2}{*}{Method} & \multicolumn{3}{c}{Greedy} & \multicolumn{3}{c}{OPERA} & \multicolumn{3}{c}{VCD} & \multicolumn{3}{c}{PAI} & \multicolumn{3}{c}{Devils} \\
    \cmidrule(lr){3-5} \cmidrule(lr){6-8} \cmidrule(lr){9-11} \cmidrule(lr){12-14} \cmidrule(lr){15-17}
    & & Orig. & +Ours & $\Delta$\% & Orig. & +Ours& $\Delta$\% & Orig. & +Ours& $\Delta$\% & Orig. & +Ours& $\Delta$\% & Orig. & +Ours& $\Delta$\% \\
    \midrule
    \multirow{3}{*}{\rotatebox{90}{Gen.}} &
    CHAIR $\downarrow$ &
    6.7 & \cellcolor{lightyellow}5.1 & {\small\textcolor{babyblue}{$\downarrow$23.9\%}} &
    7.4 & \cellcolor{lightyellow}5.8 & {\small\textcolor{babyblue}{$\downarrow$21.6\%}} &
    8.5 & \cellcolor{lightyellow}6.1 & {\small\textcolor{babyblue}{$\downarrow$28.2\%}} &
    5.1 & \cellcolor{lightyellow}4.7 & {\small\textcolor{babyblue}{$\downarrow$7.8\%}} &
    4.1 & \cellcolor{lightyellow}\textbf{3.9} & {\small\textcolor{babyblue}{$\downarrow$4.9\%}} \\
    & Hal $\downarrow$ &
    30.2 & \cellcolor{lightyellow}24.2 & {\small\textcolor{babyblue}{$\downarrow$19.9\%}} &
    33.0 & \cellcolor{lightyellow}23.3 & {\small\textcolor{babyblue}{$\downarrow$29.4\%}} &
    38.4 & \cellcolor{lightyellow}28.6 & {\small\textcolor{babyblue}{$\downarrow$25.5\%}} &
    25.1 & \cellcolor{lightyellow}22.5 & {\small\textcolor{babyblue}{$\downarrow$10.4\%}} &
    21.0 & \cellcolor{lightyellow}\textbf{20.9} & {\small\textcolor{babyblue}{$\downarrow$0.5\%}} \\
    & Cog $\downarrow$ &
    3.8 & \cellcolor{lightyellow}2.3 & {\small\textcolor{babyblue}{$\downarrow$39.5\%}} &
    3.7 & \cellcolor{lightyellow}2.1 & {\small\textcolor{babyblue}{$\downarrow$43.2\%}} &
    4.4 & \cellcolor{lightyellow}2.3 & {\small\textcolor{babyblue}{$\downarrow$47.7\%}} &
    1.9 & \cellcolor{lightyellow}1.9 & {\small\textcolor{babyblue}{$\downarrow$0.0\%}} &
    \textbf{1.4} & \cellcolor{lightyellow}1.5 & {\small\textcolor{softgray}{$\uparrow$7.1\%}} \\
    \cmidrule(lr){1-17}
    \multirow{3}{*}{\rotatebox{90}{Disc.}} & 
     Pre. $\uparrow$ & 
    100.0 & \cellcolor{lightyellow}100.0 & {\small\textcolor{babyblue}{$\uparrow$0.0\%}} &
    100.0 & \cellcolor{lightyellow}100.0 & {\small\textcolor{babyblue}{$\uparrow$0.0\%}} &
    100.0 & \cellcolor{lightyellow}100.0 & {\small\textcolor{babyblue}{$\uparrow$0.0\%}} &
    100.0 & \cellcolor{lightyellow}100.0 & {\small\textcolor{babyblue}{$\uparrow$0.0\%}} &
    100.0 & \cellcolor{lightyellow}100.0 & {\small\textcolor{babyblue}{$\uparrow$0.0\%}} \\
    & Rec. $\uparrow$ & 
    71.2 & \cellcolor{lightyellow}78.0 & {\small\textcolor{babyblue}{$\uparrow$9.6\%}} &
    74.9 & \cellcolor{lightyellow}\textbf{81.0} & {\small\textcolor{babyblue}{$\uparrow$7.5\%}} &
    67.3 & \cellcolor{lightyellow}75.7 & {\small\textcolor{babyblue}{$\uparrow$12.5\%}} &
    71.9 & \cellcolor{lightyellow}74.1 & {\small\textcolor{babyblue}{$\uparrow$3.1\%}} &
    72.5 & \cellcolor{lightyellow}75.2 & {\small\textcolor{babyblue}{$\uparrow$3.7\%}} \\
    & F1 $\uparrow$ & 
    83.2 & \cellcolor{lightyellow}87.6 & {\small\textcolor{babyblue}{$\uparrow$5.3\%}} &
    85.6 & \cellcolor{lightyellow}\textbf{89.5} & {\small\textcolor{babyblue}{$\uparrow$4.6\%}} &
    80.4 & \cellcolor{lightyellow}86.2 & {\small\textcolor{babyblue}{$\uparrow$7.2\%}} &
    83.6 & \cellcolor{lightyellow}85.1 & {\small\textcolor{babyblue}{$\uparrow$1.8\%}} &
    84.1 & \cellcolor{lightyellow}85.8 & {\small\textcolor{babyblue}{$\uparrow$2.0\%}} \\
    \midrule
    \multicolumn{2}{c|}{AMBER$\uparrow$} & 
    88.2 & \cellcolor{lightyellow}91.2 & {\small\textcolor{babyblue}{$\uparrow$3.4\%}} &
    89.1 & \cellcolor{lightyellow}\textbf{91.8} & {\small\textcolor{babyblue}{$\uparrow$3.0\%}} &
    86.0 & \cellcolor{lightyellow}90.1 & {\small\textcolor{babyblue}{$\uparrow$4.8\%}} &
    89.2 & \cellcolor{lightyellow}90.2 & {\small\textcolor{babyblue}{$\uparrow$1.1\%}} &
    90.0 & \cellcolor{lightyellow}91.0 & {\small\textcolor{babyblue}{$\uparrow$1.1\%}} \\
    \bottomrule
  \end{tabular}
  \vspace{-1em}
  }
\end{table}

\subsection{Experimental results}
\paragraph{Quantitative Results.}
We evaluate the effectiveness of our method in mitigating object hallucinations in multiple LVLM using CHAIR~\cite{rohrbach2018object} and POPE~\cite{li2023evaluating} benchmarks. As shown in Table~\ref{tab:main_quanti}, our method consistently reduces hallucination rates ($C_s$, $C_i$) across LLaVA-1.5-7B, Shikra-7B, and MiniGPT-4, while preserving caption quality (F1). For example, on LLaVA-1.5-7B, $C_s$ drops from 47.4 to 29.2 and $C_i$ from 12.2 to 9.3. Although the improvement on MiniGPT-4 is smaller, this is likely due to its Q-Former module between the vision encoder and LLM, which limits the effect of our method modifying the vision encoder. In POPE, our method yields comparable or slightly improved performance across all models, indicating robustness under discriminative evaluation settings. Furthermore, it integrates well with existing mitigation methods such as OPERA, VCD, PAI, and Devils, providing additional gains without compromising caption quality. We also present results on newer models (DeepSeek-VL~\cite{lu2024deepseek}, Qwen2.5-VL~\cite{bai2025qwen2}), and larger models (LLaVA-1.5-13B) are provided in Appendix Sec.~\ref{append_sec:quanti}.

We further evaluate our method on the AMBER benchmark~\cite{wang2023llm} using LLaVA-1.5-7B across five strategies as depicted in Table~\ref{tab:llava7b_amber}. Our approach substantially reduces object hallucinations in both generative and discriminative tasks, achieving up to a 28.2\% reduction in CHAIR and a 7.2\% improvement in F1, resulting in consistently higher AMBER scores across all settings.

\paragraph{Qualitative results.}
\begin{figure}[!t]
    \centering
    \includegraphics[width=\linewidth]{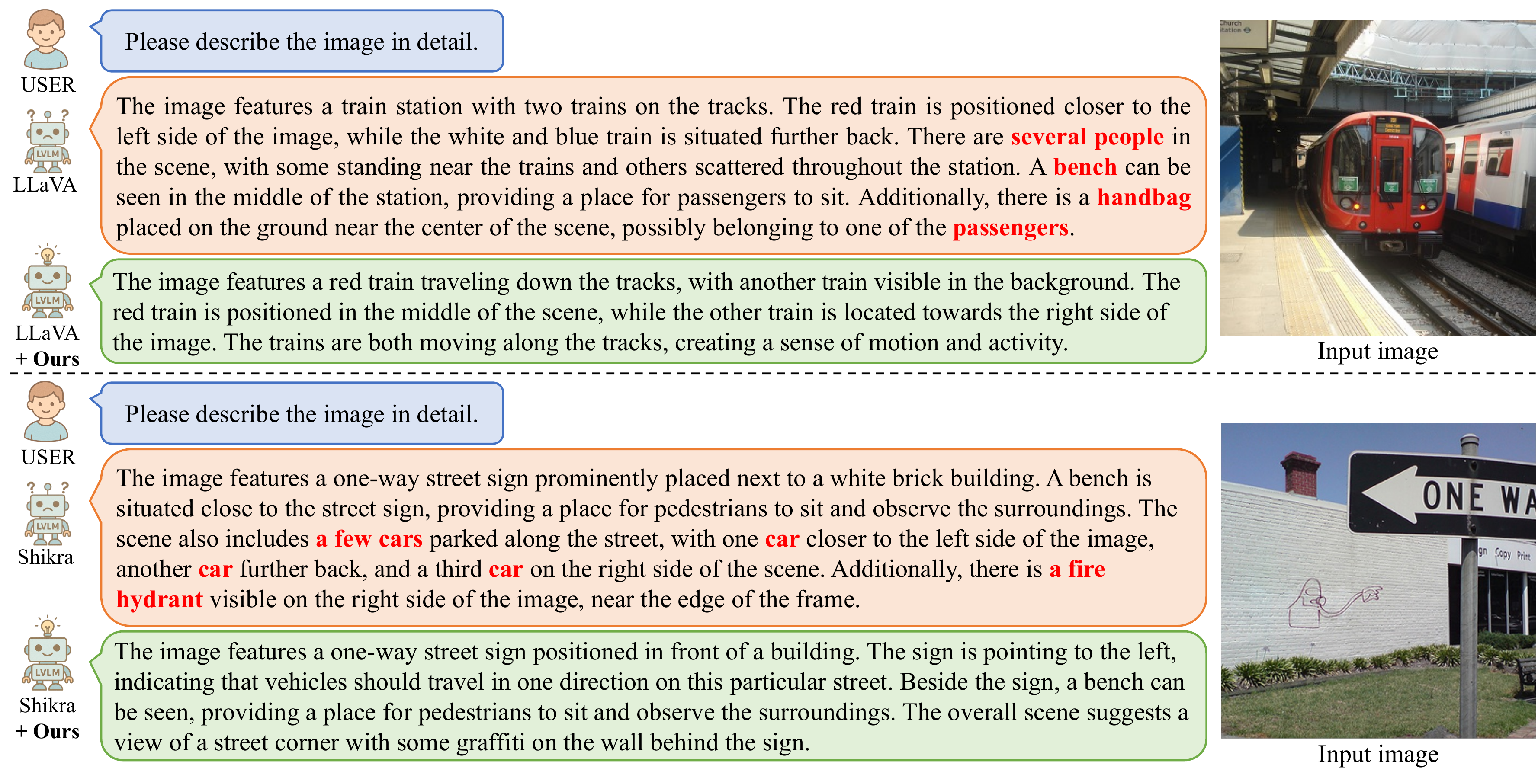}
    \caption{\textbf{Qualitative results of our method on LLaVA-1.5-7B and Shikra-7B.} Greedy decoding leads to object hallucinations by describing non-existent objects in the image (e.g., `\textit{several people}', `\textit{bench}', `\textit{handbag}', `\textit{passengers}' in LLaVA; `\textit{a few cars}', ` \textit{car}', `\textit{a fire hydrant}' in Shikra). In contrast, our method, which modifies only the vision encoder, substantially reduces such hallucinations.}
    \label{fig:main_quali}
    \vspace{-1em}
\end{figure}
We provide qualitative examples to demonstrate the effectiveness of our method. As shown in Fig.~\ref{fig:main_quali}, greedy decoding with vanilla LVLMs leads to object hallucinations, generating descriptions that mention non-existent objects such as \textit{several people}, \textit{bench}, \textit{car}, or \textit{a fire hydrant}. In contrast, our method substantially reduces such hallucinations in the generated outputs. Notably, in the case of Shikra integrated with our method, the model is able to correctly identify previously overlooked objects like \textit{graffiti}, reflecting improved visual grounding and descriptiveness. We provide further qualitative results for various combinations of models and methods in the Appendix Sec.~\ref{append_sec:quali}.

\subsection{Ablation Study and Analysis}\label{subsec:ablations}
To assess the impact of each component on reducing object hallucination, we perform ablation studies on the LLaVA-1.5-7B~\cite{liu2024improved} model. We examine two key factors in the vision encoder: (1) uncertain visual token estimation and (2) a training-free masking strategy. Each experiment isolates one variable to ensure fair comparison. Limitations of our method are discussed in Appendix~\ref{append_limitation}.

\begin{table}[!t]
  \begin{minipage}[t]{0.48\textwidth}
    \centering
    \small
    \caption{\textbf{Impact of vision encoder layers on generating the uncertainty mask $M$.} Using early layers of vision encoder (1--10) to compute $M$ yields the most effective object hallucination mitigation performance.}
    \label{tab:early_layers}
    \begin{tabular}{lccc}
      \toprule
      Mask Source Layer & $C_s$$\downarrow$ & $C_i$  $\downarrow$ & F1 $\uparrow$ \\
      \midrule
      Greedy & 47.4 & 12.2 & 77.9 \\
      \textbf{Layers 1--10} & \cellcolor{lightyellow} \textbf{29.2} & \cellcolor{lightyellow} \textbf{9.3} & \cellcolor{lightyellow} \textbf{78.2} \\
      Layers 11--20 & 44.2 & 12.7 & 77.4 \\
      Layers 21--22 & 41.8 & 12.1 & 77.9 \\
      \bottomrule
    \end{tabular}
  \end{minipage}
  \hfill
  \begin{minipage}[t]{0.50\textwidth}
    \centering
    \small
    \caption{\textbf{Effect of applying the uncertainty mask $M$ to different layers in the vision encoder.} Applying the mask at middle layers of vision encoder (13--17) results in the most effective performance.}
    \label{tab:masking_layers}
    \begin{tabular}{lccc}
      \toprule
      Masking Layer Range & $C_s$ $\downarrow$ & $C_i$ $\downarrow$ & F1 $\uparrow$ \\
      \midrule
      Greedy & 47.4 & 12.2 & 77.9 \\
      Layers 1--8 & 45.0 & 12.6 & 77.9 \\
      Layers 8--12 & 55.8 & 15.5 & 75.7 \\
      \textbf{Layers 13--17} & \cellcolor{lightyellow} \textbf{29.2} & \cellcolor{lightyellow} \textbf{9.3} & \cellcolor{lightyellow} \textbf{78.2} \\
      Layers 18--22 & 45.8 & 13.0 & 77.7 \\
      \bottomrule
    \end{tabular}
  \end{minipage}
\end{table}
\begin{table}[!t]
  \centering
  \small
  \caption{\textbf{Comparison of masking strategies for uncertain visual tokens.} We compare our attention-level masking method with alternatives applied at different stages of the vision encoder (VE). S.M. denotes soft masking, which attenuates uncertain tokens by a small factor (e.g., 0.1 or 0.2).}
  \label{tab:masking_strategies}
  \begin{tabular}{lcccccc}
    \toprule
    \textbf{Strategy} & Greedy & Input of VE & Output of VE & MLP Layer & S.M. (0.1 / 0.2) & \textbf{Ours} \\
    \midrule
    $C_s$ $\downarrow$ & 47.4 & 47.4 & 34.4 & 51.0 & 35.0 / 40.0 & \cellcolor{lightyellow} \textbf{29.2} \\
    $C_i$ $\downarrow$ & 12.2 & 12.5 & 10.0 & 13.5 & 10.4 / 11.5 & \cellcolor{lightyellow} \textbf{9.3} \\
    F1  $\uparrow$    & 77.9 & 77.5 & 74.7 & 77.9 & \textbf{78.3} / 78.1 & \cellcolor{lightyellow} 78.2 \\
    \bottomrule
  \end{tabular}
\end{table}

\paragraph{Uncertainty estimation of visual tokens from early layers of vision encoder.}
We examine which layers of vision encoder are most effective for generating the binary uncertainty mask $M$ using PGD-based adversarial attacks. As shown in Table~\ref{tab:early_layers}, extracting uncertainty from early layers (1 to 10) leads to the largest reduction in hallucinations ($C_s$, $C_i$) and the highest F1 score, outperforming intermediate or deeper layers. This result aligns with Sec.\ref{subsubsec:adv_emp} and Fig.\ref{fig:layerwise_diff}, where early layers exhibit smaller adversarial feature shifts, making them more suitable for uncertainty estimation.

\paragraph{Masking uncertain visual tokens in intermediate layers of vision encoder.}
We investigate the effect of applying the binary uncertainty mask $M$ to different layers of self-attention process within the vision encoder. As shown in Table~\ref{tab:masking_layers}, masking at intermediate layers (13 to 17) yields the best performance, significantly reducing hallucination ($C_s$, $C_i$) and achieving the highest F1 score. In contrast, masking in earlier layers shows limited benefit, and deeper layers provide minimal gains. 

\paragraph{Comparative analysis of masking strategies for uncertain visual tokens.}
We compare several masking strategies using the binary uncertainty mask $M$, including masking at the input image, the output of the vision encoder, the MLP layer before the residual connection in the transformer block, and soft masking applied to the self-attention that attenuates uncertain visual tokens by a small factor. As shown in Table~\ref{tab:masking_strategies}, our method, which applies hard masking within the self-attention mechanism using $M$, achieves the best hallucination scores while maintaining a competitive F1 score. 

\section{Conclusion}
We present a simple yet effective approach for mitigating object hallucination in Large Vision-Language Models (LVLMs) by identifying uncertain visual tokens within the vision encoder and reducing their influence through masking these tokens in their self-attention layers. Our theoretical and empirical analyses show that adversarial perturbations efficiently approximate an upper bound of epistemic uncertainty, which we confirm to be strongly correlated with object hallucination in LVLMs. Extensive experiments demonstrate that our approach consistently reduces object hallucination across diverse models and integrates seamlessly with other prior arts to improve performance.

\section*{Acknowledgement}

This work was supported in part by Institute of Information \& communications Technology Planning \& Evaluation (IITP) grants funded by the Korea government(MSIT) [NO.RS-2021-II211343, Artificial Intelligence Graduate School Program (Seoul National University)], (No.RS-2025-02314125, Effective Human-Machine Teaming With Multimodal Hazy Oracle Models) and by the National Research Foundation of Korea(NRF) grant funded by the Korea government(MSIT) (No. RS-2025-02263628). Also, the authors acknowledged the financial support from the BK21 FOUR program of the Education and Research Program for Future ICT Pioneers, Seoul National University.

{
    \small
    \bibliographystyle{plain}
    \bibliography{main}

\begin{thebibliography}{10}

\bibitem{agrawal2019nocaps}
Harsh Agrawal, Karan Desai, Yufei Wang, Xinlei Chen, Rishabh Jain, Mark Johnson, Dhruv Batra, Devi Parikh, Stefan Lee, and Peter Anderson.
\newblock Nocaps: Novel object captioning at scale.
\newblock In {\em ICCV}, pages 8948--8957, 2019.

\bibitem{an2024agla}
Wenbin An, Feng Tian, Sicong Leng, Jiahao Nie, Haonan Lin, QianYing Wang, Guang Dai, Ping Chen, and Shijian Lu.
\newblock Agla: Mitigating object hallucinations in large vision-language models with assembly of global and local attention.
\newblock {\em arXiv preprint arXiv:2406.12718}, 2024.

\bibitem{qwen}
Jinze Bai, Shuai Bai, Yunfei Chu, Zeyu Cui, Kai Dang, Xiaodong Deng, Yang Fan, Wenbin Ge, Yu~Han, Fei Huang, Binyuan Hui, Luo Ji, Mei Li, Junyang Lin, Runji Lin, Dayiheng Liu, Gao Liu, Chengqiang Lu, Keming Lu, Jianxin Ma, Rui Men, Xingzhang Ren, Xuancheng Ren, Chuanqi Tan, Sinan Tan, Jianhong Tu, Peng Wang, Shijie Wang, Wei Wang, Shengguang Wu, Benfeng Xu, Jin Xu, An~Yang, Hao Yang, Jian Yang, Shusheng Yang, Yang Yao, Bowen Yu, Hongyi Yuan, Zheng Yuan, Jianwei Zhang, Xingxuan Zhang, Yichang Zhang, Zhenru Zhang, Chang Zhou, Jingren Zhou, Xiaohuan Zhou, and Tianhang Zhu.
\newblock Qwen technical report.
\newblock {\em arXiv preprint arXiv:2309.16609}, 2023.

\bibitem{bai2025qwen2}
Shuai Bai, Keqin Chen, Xuejing Liu, Jialin Wang, Wenbin Ge, Sibo Song, Kai Dang, Peng Wang, Shijie Wang, Jun Tang, et~al.
\newblock Qwen2. 5-vl technical report.
\newblock {\em arXiv preprint arXiv:2502.13923}, 2025.

\bibitem{bai2024hallucination}
Zechen Bai, Pichao Wang, Tianjun Xiao, Tong He, Zongbo Han, Zheng Zhang, and Mike~Zheng Shou.
\newblock Hallucination of multimodal large language models: A survey.
\newblock {\em arXiv preprint arXiv:2404.18930}, 2024.

\bibitem{cai2023ensemble}
Zikui Cai, Yaoteng Tan, and M~Salman Asif.
\newblock Ensemble-based blackbox attacks on dense prediction.
\newblock In {\em CVPR}, pages 4045--4055, 2023.

\bibitem{carlini2023aligned}
Nicholas Carlini, Milad Nasr, Christopher~A Choquette-Choo, Matthew Jagielski, Irena Gao, Pang Wei~W Koh, Daphne Ippolito, Florian Tramer, and Ludwig Schmidt.
\newblock Are aligned neural networks adversarially aligned?
\newblock {\em NeurIPS}, 36:61478--61500, 2023.

\bibitem{che2025eazy}
Liwei Che, Tony~Qingze Liu, Jing Jia, Weiyi Qin, Ruixiang Tang, and Vladimir Pavlovic.
\newblock Eazy: Eliminating hallucinations in lvlms by zeroing out hallucinatory image tokens.
\newblock {\em arXiv preprint arXiv:2503.07772}, 2025.

\bibitem{chen2023shikra}
Keqin Chen, Zhao Zhang, Weili Zeng, Richong Zhang, Feng Zhu, and Rui Zhao.
\newblock Shikra: Unleashing multimodal llm's referential dialogue magic.
\newblock {\em arXiv preprint arXiv:2306.15195}, 2023.

\bibitem{chen2024pali}
Xi~Chen, Xiao Wang, Soravit Changpinyo, AJ~Piergiovanni, Piotr Padlewski, Daniel Salz, Sebastian Goodman, Adam Grycner, Basil Mustafa, Lucas Beyer, et~al.
\newblock Pali: A jointly-scaled multilingual language-image model.
\newblock In {\em ICLR}, 2023.

\bibitem{vicuna2023}
Wei-Lin Chiang, Zhuohan Li, Zi~Lin, Ying Sheng, Zhanghao Wu, Hao Zhang, Lianmin Zheng, Siyuan Zhuang, Yonghao Zhuang, Joseph~E. Gonzalez, Ion Stoica, and Eric~P. Xing.
\newblock Vicuna: An open-source chatbot impressing gpt-4 with 90\%* chatgpt quality, March 2023.

\bibitem{chuang2024dola}
Yung-Sung Chuang, Yujia Xie, Hongyin Luo, Yoon Kim, James~R. Glass, and Pengcheng He.
\newblock Dola: Decoding by contrasting layers improves factuality in large language models.
\newblock In {\em ICLR}, 2024.

\bibitem{conover1999practical}
William~Jay Conover.
\newblock {\em Practical nonparametric statistics}.
\newblock john wiley \& sons, 1999.

\bibitem{dai2023instructblip}
Wenliang Dai, Junnan Li, Dongxu Li, Anthony Tiong, Junqi Zhao, Weisheng Wang, Boyang Li, Pascale Fung, and Steven Hoi.
\newblock Instruct{BLIP}: Towards general-purpose vision-language models with instruction tuning.
\newblock In {\em NeurIPS}, 2023.

\bibitem{dosovitskiy2020vit}
Alexey Dosovitskiy, Lucas Beyer, Alexander Kolesnikov, Dirk Weissenborn, Xiaohua Zhai, Thomas Unterthiner, Mostafa Dehghani, Matthias Minderer, Georg Heigold, Sylvain Gelly, Jakob Uszkoreit, and Neil Houlsby.
\newblock An image is worth 16x16 words: Transformers for image recognition at scale.
\newblock {\em ICLR}, 2021.

\bibitem{duan2025truthprint}
Jinhao Duan, Fei Kong, Hao Cheng, James Diffenderfer, Bhavya Kailkhura, Lichao Sun, Xiaofeng Zhu, Xiaoshuang Shi, and Kaidi Xu.
\newblock Truthprint: Mitigating lvlm object hallucination via latent truthful-guided pre-intervention.
\newblock {\em arXiv preprint arXiv:2503.10602}, 2025.

\bibitem{fang2023eva}
Yuxin Fang, Wen Wang, Binhui Xie, Quan Sun, Ledell Wu, Xinggang Wang, Tiejun Huang, Xinlong Wang, and Yue Cao.
\newblock Eva: Exploring the limits of masked visual representation learning at scale.
\newblock In {\em CVPR}, pages 19358--19369, 2023.

\bibitem{goodfellow2015explaining}
Ian~J Goodfellow, Jonathon Shlens, and Christian Szegedy.
\newblock Explaining and harnessing adversarial examples.
\newblock {\em ICLR}, 2015.

\bibitem{gunjal2024detecting}
Anisha Gunjal, Jihan Yin, and Erhan Bas.
\newblock Detecting and preventing hallucinations in large vision language models.
\newblock In {\em AAAI}, volume~38, pages 18135--18143, 2024.

\bibitem{huang2024opera}
Qidong Huang, Xiaoyi Dong, Pan Zhang, Bin Wang, Conghui He, Jiaqi Wang, Dahua Lin, Weiming Zhang, and Nenghai Yu.
\newblock Opera: Alleviating hallucination in multi-modal large language models via over-trust penalty and retrospection-allocation.
\newblock In {\em CVPR}, pages 13418--13427, 2024.

\bibitem{huo2024self}
Fushuo Huo, Wenchao Xu, Zhong Zhang, Haozhao Wang, Zhicheng Chen, and Peilin Zhao.
\newblock Self-introspective decoding: Alleviating hallucinations for large vision-language models.
\newblock {\em arXiv preprint arXiv:2408.02032}, 2024.

\bibitem{ilharco_gabriel_2021_5143773}
Gabriel Ilharco, Mitchell Wortsman, Ross Wightman, Cade Gordon, Nicholas Carlini, Rohan Taori, Achal Dave, Vaishaal Shankar, Hongseok Namkoong, John Miller, Hannaneh Hajishirzi, Ali Farhadi, and Ludwig Schmidt.
\newblock Openclip, July 2021.
\newblock If you use this software, please cite it as below.

\bibitem{jiang2024hallucination}
Chaoya Jiang, Haiyang Xu, Mengfan Dong, Jiaxing Chen, Wei Ye, Ming Yan, Qinghao Ye, Ji~Zhang, Fei Huang, and Shikun Zhang.
\newblock Hallucination augmented contrastive learning for multimodal large language model.
\newblock In {\em CVPR}, pages 27036--27046, 2024.

\bibitem{jiang2024devils}
Zhangqi Jiang, Junkai Chen, Beier Zhu, Tingjin Luo, Yankun Shen, and Xu~Yang.
\newblock Devils in middle layers of large vision-language models: Interpreting, detecting and mitigating object hallucinations via attention lens.
\newblock {\em arXiv preprint arXiv:2411.16724}, 2024.

\bibitem{kang2025see}
Seil Kang, Jinyeong Kim, Junhyeok Kim, and Seong~Jae Hwang.
\newblock See what you are told: Visual attention sink in large multimodal models.
\newblock In {\em ICLR}, 2025.

\bibitem{laves2020calibration}
Max-Heinrich Laves, Sontje Ihler, Karl-Philipp Kortmann, and Tobias Ortmaier.
\newblock Calibration of model uncertainty for dropout variational inference.
\newblock {\em arXiv preprint arXiv:2006.11584}, 2020.

\bibitem{leng2024mitigating}
Sicong Leng, Hang Zhang, Guanzheng Chen, Xin Li, Shijian Lu, Chunyan Miao, and Lidong Bing.
\newblock Mitigating object hallucinations in large vision-language models through visual contrastive decoding.
\newblock In {\em CVPR}, pages 13872--13882, 2024.

\bibitem{li2024llava}
Bo~Li, Yuanhan Zhang, Dong Guo, Renrui Zhang, Feng Li, Hao Zhang, Kaichen Zhang, Peiyuan Zhang, Yanwei Li, Ziwei Liu, et~al.
\newblock Llava-onevision: Easy visual task transfer.
\newblock {\em arXiv preprint arXiv:2408.03326}, 2024.

\bibitem{li2025mitigating}
Jiaming Li, Jiacheng Zhang, Zequn Jie, Lin Ma, and Guanbin Li.
\newblock Mitigating hallucination for large vision language model by inter-modality correlation calibration decoding.
\newblock {\em arXiv preprint arXiv:2501.01926}, 2025.

\bibitem{li2023blip}
Junnan Li, Dongxu Li, Silvio Savarese, and Steven Hoi.
\newblock Blip-2: Bootstrapping language-image pre-training with frozen image encoders and large language models.
\newblock In {\em ICML}, pages 19730--19742. PMLR, 2023.

\bibitem{li2023evaluating}
Yifan Li, Yifan Du, Kun Zhou, Jinpeng Wang, Xin Zhao, and Ji-Rong Wen.
\newblock Evaluating object hallucination in large vision-language models.
\newblock In {\em EMNLP}, 2023.

\bibitem{liang2021parallel}
Siyuan Liang, Baoyuan Wu, Yanbo Fan, Xingxing Wei, and Xiaochun Cao.
\newblock Parallel rectangle flip attack: A query-based black-box attack against object detection.
\newblock In {\em ICCV}, pages 7677--7687. IEEE Computer Society, 2021.

\bibitem{lin2014microsoft}
Tsung-Yi Lin, Michael Maire, Serge Belongie, James Hays, Pietro Perona, Deva Ramanan, Piotr Doll{\'a}r, and C~Lawrence Zitnick.
\newblock Microsoft coco: Common objects in context.
\newblock In {\em ECCV}, pages 740--755. Springer, 2014.

\bibitem{liu2024survey}
Hanchao Liu, Wenyuan Xue, Yifei Chen, Dapeng Chen, Xiutian Zhao, Ke~Wang, Liping Hou, Rongjun Li, and Wei Peng.
\newblock A survey on hallucination in large vision-language models.
\newblock {\em arXiv preprint arXiv:2402.00253}, 2024.

\bibitem{liu2024improved}
Haotian Liu, Chunyuan Li, Yuheng Li, and Yong~Jae Lee.
\newblock Improved baselines with visual instruction tuning.
\newblock In {\em CVPR}, pages 26296--26306, 2024.

\bibitem{liu2023visual}
Haotian Liu, Chunyuan Li, Qingyang Wu, and Yong~Jae Lee.
\newblock Visual instruction tuning.
\newblock {\em NeurIPS}, 36:34892--34916, 2023.

\bibitem{liu2024paying}
Shi Liu, Kecheng Zheng, and Wei Chen.
\newblock Paying more attention to image: A training-free method for alleviating hallucination in lvlms.
\newblock In {\em ECCV}, pages 125--140. Springer, 2024.

\bibitem{lu2024deepseek}
Haoyu Lu, Wen Liu, Bo~Zhang, Bingxuan Wang, Kai Dong, Bo~Liu, Jingxiang Sun, Tongzheng Ren, Zhuoshu Li, Hao Yang, et~al.
\newblock Deepseek-vl: towards real-world vision-language understanding.
\newblock {\em arXiv preprint arXiv:2403.05525}, 2024.

\bibitem{lyu2024alleviating}
Xinyu Lyu, Beitao Chen, Lianli Gao, Hengtao Shen, and Jingkuan Song.
\newblock Alleviating hallucinations in large vision-language models through hallucination-induced optimization.
\newblock {\em NeurIPS}, 37:122811--122832, 2024.

\bibitem{madry2018towards}
Aleksander Madry, Aleksandar Makelov, Ludwig Schmidt, Dimitris Tsipras, and Adrian Vladu.
\newblock Towards deep learning models resistant to adversarial attacks.
\newblock In {\em ICLR}, 2018.

\bibitem{madry2018deep}
Aleksander Madry, Aleksandar Makelov, Ludwig Schmidt, Dimitris Tsipras, and Adrian Vladu.
\newblock Towards deep learning models resistant to adversarial attacks.
\newblock In {\em ICLR}, 2018.

\bibitem{mao2025through}
Shunqi Mao, Chaoyi Zhang, and Weidong Cai.
\newblock Through the magnifying glass: Adaptive perception magnification for hallucination-free vlm decoding.
\newblock {\em arXiv preprint arXiv:2503.10183}, 2025.

\bibitem{marino2019ok}
Kenneth Marino, Mohammad Rastegari, Ali Farhadi, and Roozbeh Mottaghi.
\newblock Ok-vqa: A visual question answering benchmark requiring external knowledge.
\newblock In {\em CVPR}, pages 3195--3204, 2019.

\bibitem{mukhoti2018evaluating}
Jishnu Mukhoti and Yarin Gal.
\newblock Evaluating bayesian deep learning methods for semantic segmentation.
\newblock {\em arXiv preprint arXiv:1811.12709}, 2018.

\bibitem{qi2024visual}
Xiangyu Qi, Kaixuan Huang, Ashwinee Panda, Peter Henderson, Mengdi Wang, and Prateek Mittal.
\newblock Visual adversarial examples jailbreak aligned large language models.
\newblock In {\em AAAI}, volume~38, pages 21527--21536, 2024.

\bibitem{radford2021learning}
Alec Radford, Jong~Wook Kim, Chris Hallacy, Aditya Ramesh, Gabriel Goh, Sandhini Agarwal, Girish Sastry, Amanda Askell, Pamela Mishkin, Jack Clark, et~al.
\newblock Learning transferable visual models from natural language supervision.
\newblock In {\em ICML}, pages 8748--8763. PMLR, 2021.

\bibitem{rohrbach2018object}
Anna Rohrbach, Lisa~Anne Hendricks, Kaylee Burns, Trevor Darrell, and Kate Saenko.
\newblock Object hallucination in image captioning.
\newblock In {\em EMNLP}, pages 4035--4045, 2018.

\bibitem{rovai2013social}
Alfred~P Rovai, Jason~D Baker, and Michael~K Ponton.
\newblock {\em Social science research design and statistics: A practitioner's guide to research methods and IBM SPSS}.
\newblock Watertree Press LLC, 2013.

\bibitem{schlarmann2023adversarial}
Christian Schlarmann and Matthias Hein.
\newblock On the adversarial robustness of multi-modal foundation models.
\newblock In {\em ICCV}, pages 3677--3685, 2023.

\bibitem{shayegani2024jailbreak}
Erfan Shayegani, Yue Dong, and Nael Abu-Ghazaleh.
\newblock Jailbreak in pieces: Compositional adversarial attacks on multi-modal language models.
\newblock In {\em ICLR}, 2024.

\bibitem{sun2023eva}
Quan Sun, Yuxin Fang, Ledell Wu, Xinlong Wang, and Yue Cao.
\newblock Eva-clip: Improved training techniques for clip at scale.
\newblock {\em arXiv preprint arXiv:2303.15389}, 2023.

\bibitem{sutskever2014sequence}
Ilya Sutskever, Oriol Vinyals, and Quoc~V Le.
\newblock Sequence to sequence learning with neural networks.
\newblock {\em Advances in neural information processing systems}, 27, 2014.

\bibitem{szegedy2013intriguing}
Christian Szegedy, Wojciech Zaremba, Ilya Sutskever, Joan Bruna, Dumitru Erhan, Ian Goodfellow, and Rob Fergus.
\newblock Intriguing properties of neural networks.
\newblock {\em arXiv preprint arXiv:1312.6199}, 2013.

\bibitem{touvron2023llama}
Hugo Touvron, Louis Martin, Kevin Stone, Peter Albert, Amjad Almahairi, Yasmine Babaei, Nikolay Bashlykov, Soumya Batra, Prajjwal Bhargava, Shruti Bhosale, et~al.
\newblock Llama 2: Open foundation and fine-tuned chat models.
\newblock {\em arXiv preprint arXiv:2307.09288}, 2023.

\bibitem{wang2024mllm}
Chenxi Wang, Xiang Chen, Ningyu Zhang, Bozhong Tian, Haoming Xu, Shumin Deng, and Huajun Chen.
\newblock Mllm can see? dynamic correction decoding for hallucination mitigation.
\newblock {\em arXiv preprint arXiv:2410.11779}, 2024.

\bibitem{wang2023llm}
Junyang Wang, Yuhang Wang, Guohai Xu, Jing Zhang, Yukai Gu, Haitao Jia, Ming Yan, Ji~Zhang, and Jitao Sang.
\newblock An llm-free multi-dimensional benchmark for mllms hallucination evaluation.
\newblock {\em CoRR}, 2023.

\bibitem{wang2023evaluation}
Junyang Wang, Yiyang Zhou, Guohai Xu, Pengcheng Shi, Chenlin Zhao, Haiyang Xu, Qinghao Ye, Ming Yan, Ji~Zhang, Jihua Zhu, et~al.
\newblock Evaluation and analysis of hallucination in large vision-language models.
\newblock {\em arXiv preprint arXiv:2308.15126}, 2023.

\bibitem{wang2023image}
Wenhui Wang, Hangbo Bao, Li~Dong, Johan Bjorck, Zhiliang Peng, Qiang Liu, Kriti Aggarwal, Owais~Khan Mohammed, Saksham Singhal, Subhojit Som, et~al.
\newblock Image as a foreign language: Beit pretraining for vision and vision-language tasks.
\newblock In {\em CVPR}, pages 19175--19186, 2023.

\bibitem{wang2024break}
Yubo Wang, Chaohu Liu, Yanqiu Qu, Haoyu Cao, Deqiang Jiang, and Linli Xu.
\newblock Break the visual perception: Adversarial attacks targeting encoded visual tokens of large vision-language models.
\newblock In {\em ACM MM}, pages 1072--1081, 2024.

\bibitem{xie2025tarac}
Chunzhao Xie, Tongxuan Liu, Lei Jiang, Yuting Zeng, Yunheng Shen, Weizhe Huang, Jing Li, Xiaohua Xu, et~al.
\newblock Tarac: Mitigating hallucination in lvlms via temporal attention real-time accumulative connection.
\newblock {\em arXiv preprint arXiv:2504.04099}, 2025.

\bibitem{xie2017adversarial}
Cihang Xie, Jianyu Wang, Zhishuai Zhang, Yuyin Zhou, Lingxi Xie, and Alan Yuille.
\newblock Adversarial examples for semantic segmentation and object detection.
\newblock In {\em ICCV}, pages 1369--1378, 2017.

\bibitem{qwen2}
An~Yang, Baosong Yang, Binyuan Hui, Bo~Zheng, Bowen Yu, Chang Zhou, Chengpeng Li, Chengyuan Li, Dayiheng Liu, Fei Huang, Guanting Dong, Haoran Wei, Huan Lin, Jialong Tang, Jialin Wang, Jian Yang, Jianhong Tu, Jianwei Zhang, Jianxin Ma, Jin Xu, Jingren Zhou, Jinze Bai, Jinzheng He, Junyang Lin, Kai Dang, Keming Lu, Keqin Chen, Kexin Yang, Mei Li, Mingfeng Xue, Na~Ni, Pei Zhang, Peng Wang, Ru~Peng, Rui Men, Ruize Gao, Runji Lin, Shijie Wang, Shuai Bai, Sinan Tan, Tianhang Zhu, Tianhao Li, Tianyu Liu, Wenbin Ge, Xiaodong Deng, Xiaohuan Zhou, Xingzhang Ren, Xinyu Zhang, Xipin Wei, Xuancheng Ren, Yang Fan, Yang Yao, Yichang Zhang, Yu~Wan, Yunfei Chu, Yuqiong Liu, Zeyu Cui, Zhenru Zhang, and Zhihao Fan.
\newblock Qwen2 technical report.
\newblock {\em arXiv preprint arXiv:2407.10671}, 2024.

\bibitem{ye2023mplug}
Qinghao Ye, Haiyang Xu, Guohai Xu, Jiabo Ye, Ming Yan, Yiyang Zhou, Junyang Wang, Anwen Hu, Pengcheng Shi, Yaya Shi, et~al.
\newblock mplug-owl: Modularization empowers large language models with multimodality.
\newblock {\em arXiv preprint arXiv:2304.14178}, 2023.

\bibitem{yue2024less}
Zihao Yue, Liang Zhang, and Qin Jin.
\newblock Less is more: Mitigating multimodal hallucination from an eos decision perspective.
\newblock In {\em Proceedings of the 62nd Annual Meeting of the Association for Computational Linguistics (Volume 1: Long Papers)}, pages 11766--11781, 2024.

\bibitem{zhai2023sigmoid}
Xiaohua Zhai, Basil Mustafa, Alexander Kolesnikov, and Lucas Beyer.
\newblock Sigmoid loss for language image pre-training.
\newblock In {\em ICCV}, pages 11975--11986, 2023.

\bibitem{zhang2025mllms}
Jiarui Zhang, Mahyar Khayatkhoei, Prateek Chhikara, and Filip Ilievski.
\newblock Mllms know where to look: Training-free perception of small visual details with multimodal llms.
\newblock {\em arXiv preprint arXiv:2502.17422}, 2025.

\bibitem{zhao2023evaluating}
Yunqing Zhao, Tianyu Pang, Chao Du, Xiao Yang, Chongxuan Li, Ngai-Man~Man Cheung, and Min Lin.
\newblock On evaluating adversarial robustness of large vision-language models.
\newblock {\em NeurIPS}, 36:54111--54138, 2023.

\bibitem{zhou2024analyzing}
Yiyang Zhou, Chenhang Cui, Jaehong Yoon, Linjun Zhang, Zhun Deng, Chelsea Finn, Mohit Bansal, and Huaxiu Yao.
\newblock Analyzing and mitigating object hallucination in large vision-language models.
\newblock In {\em ICLR}, 2024.

\bibitem{zhu2024minigpt}
Deyao Zhu, Jun Chen, Xiaoqian Shen, Xiang Li, and Mohamed Elhoseiny.
\newblock Minigpt-4: Enhancing vision-language understanding with advanced large language models.
\newblock In {\em ICLR}, 2024.

\bibitem{zhu2024ibd}
Lanyun Zhu, Deyi Ji, Tianrun Chen, Peng Xu, Jieping Ye, and Jun Liu.
\newblock Ibd: Alleviating hallucinations in large vision-language models via image-biased decoding.
\newblock {\em arXiv preprint arXiv:2402.18476}, 2024.

\end{thebibliography}
}


\clearpage
\newpage
\section*{NeurIPS Paper Checklist}

\begin{enumerate}

\item {\bf Claims}
    \item[] Question: Do the main claims made in the abstract and introduction accurately reflect the paper's contributions and scope?
    \item[] Answer: \answerYes{} 
    \item[] Justification: We introduced our approach to mitigate object hallucination in title, abstract, and introduction. Also, we summarized our contributions explicitly in Sec.~\ref{intro}.

\item {\bf Limitations}
    \item[] Question: Does the paper discuss the limitations of the work performed by the authors?
    \item[] Answer: \answerYes{} 
    \item[] Justification: We discussed limitation of our work in Appendix Sec~\ref{append_limitation}.

\item {\bf Theory assumptions and proofs}
    \item[] Question: For each theoretical result, does the paper provide the full set of assumptions and a complete (and correct) proof?
    \item[] Answer: \answerYes{} 
    \item[] Justification: We provide the complete set of assumptions and full proofs for Lemma~\ref{lem:local_gaussianity} and Theorem~\ref{thm:entropy}, with appropriate references to the appendix.

    \item {\bf Experimental result reproducibility}
    \item[] Question: Does the paper fully disclose all the information needed to reproduce the main experimental results of the paper to the extent that it affects the main claims and/or conclusions of the paper (regardless of whether the code and data are provided or not)?
    \item[] Answer: \answerYes{} 
    \item[] Justification: We provide code for reproducibility and detailed implementation details in Appendix Sec.~\ref{append_sec:imple}, benchmarks and baseline models in Appendix Sec.~\ref{append_sec:exp}, along with the corresponding GitHub link.

\item {\bf Open access to data and code}
    \item[] Question: Does the paper provide open access to the data and code, with sufficient instructions to faithfully reproduce the main experimental results, as described in supplemental material?
    \item[] Answer: \answerYes{} 
    \item[] Justification: We provide our code implementation in supplementary materials.

\item {\bf Experimental setting/details}
    \item[] Question: Does the paper specify all the training and test details (e.g., data splits, hyperparameters, how they were chosen, type of optimizer, etc.) necessary to understand the results?
    \item[] Answer: \answerYes{} 
    \item[] Justification: We provide experimental setting and details in Appendix Sec.~\ref{append_sec:imple} and Sec.~\ref{append_sec:exp}.

\item {\bf Experiment statistical significance}
    \item[] Question: Does the paper report error bars suitably and correctly defined or other appropriate information about the statistical significance of the experiments?
    \item[] Answer: \answerYes{} 
    \item[] Justification: We provide error bars on our experimental results in Fig.~\ref{fig:layerwise_diff}, and report our results' statistical significance of Fig.~\ref{fig:variance_vs_metric} in Sec.~\ref{subsubsec:adv_emp} and Sec.~\ref{subsubsec:does}.

\item {\bf Experiments compute resources}
    \item[] Question: For each experiment, does the paper provide sufficient information on the computer resources (type of compute workers, memory, time of execution) needed to reproduce the experiments?
    \item[] Answer: \answerYes{} 
    \item[] Justification: We specify in Appendix Sec.~\ref{append_sec:imple} that all main experiments were conducted using an NVIDIA A100 GPU with 80GB of memory.
    
\item {\bf Code of ethics}
    \item[] Question: Does the research conducted in the paper conform, in every respect, with the NeurIPS Code of Ethics \url{https://neurips.cc/public/EthicsGuidelines}?
    \item[] Answer: \answerYes{} 
    \item[] Justification: We comply with the NeurIPS Code of Ethics.

\item {\bf Broader impacts}
    \item[] Question: Does the paper discuss both potential positive societal impacts and negative societal impacts of the work performed?
    \item[] Answer: \answerYes{} 
    \item[] Justification: We discuss both the potential positive and negative societal impacts of our work in Appendix Sec.~\ref{append_broader}.
    
\item {\bf Safeguards}
    \item[] Question: Does the paper describe safeguards that have been put in place for responsible release of data or models that have a high risk for misuse (e.g., pretrained language models, image generators, or scraped datasets)?
    \item[] Answer: \answerNA{} 
    \item[] Justification: Our paper does not involve releasing any models or datasets that pose a high risk of misuse.

\item {\bf Licenses for existing assets}
    \item[] Question: Are the creators or original owners of assets (e.g., code, data, models), used in the paper, properly credited and are the license and terms of use explicitly mentioned and properly respected?
    \item[] Answer: \answerYes{} 
    \item[] Justification: We properly cite all utilized code, benchmark datasets, and models, and provide the corresponding GitHub links in Appendix Sec.~\ref{append_sec:exp}.

\item {\bf New assets}
    \item[] Question: Are new assets introduced in the paper well documented and is the documentation provided alongside the assets?
    \item[] Answer: \answerYes{} 
    \item[] Justification: We will provide a README file alongside the released code in the supplementary materials, which includes usage instructions, details of the benchmark datasets, and descriptions of the models used in our experiments.

\item {\bf Crowdsourcing and research with human subjects}
    \item[] Question: For crowdsourcing experiments and research with human subjects, does the paper include the full text of instructions given to participants and screenshots, if applicable, as well as details about compensation (if any)? 
    \item[] Answer: \answerNA{} 
    \item[] Justification: Our paper does not involve crowdsourcing nor research with human subjects.

\item {\bf Institutional review board (IRB) approvals or equivalent for research with human subjects}
    \item[] Question: Does the paper describe potential risks incurred by study participants, whether such risks were disclosed to the subjects, and whether Institutional Review Board (IRB) approvals (or an equivalent approval/review based on the requirements of your country or institution) were obtained?
    \item[] Answer: \answerNA{} 
    \item[] Justification: Our paper does not involve crowdsourcing nor research with human subjects.

\item {\bf Declaration of LLM usage}
    \item[] Question: Does the paper describe the usage of LLMs if it is an important, original, or non-standard component of the core methods in this research? Note that if the LLM is used only for writing, editing, or formatting purposes and does not impact the core methodology, scientific rigorousness, or originality of the research, declaration is not required.
    \item[] Answer: \answerNA{}{}
    \item[] Justification: Our core method development in this research does not incorporate large language models as any essential, novel, or non-standard components.

\end{enumerate}

\clearpage
\appendix

\renewcommand{\thefigure}{A\arabic{figure}}
\renewcommand{\thetable}{A\arabic{table}}
\renewcommand{\theequation}{A\arabic{equation}}
\setcounter{section}{0}
\setcounter{figure}{0}
\setcounter{table}{0}
\setcounter{theorem}{0}
\setcounter{equation}{0}

\section{Proofs}\label{append_sec:proofs}
\subsection{Proof of Lemma~\ref{lem:local_gaussianity}}\label{append_proof:local_gaussianity}
\renewcommand{\thetheorem}{3.1}

\begin{lemma}[Approximate local Gaussianity under small perturbation]\label{lem:local_gaussianity}
Let $f = \{f_t\}_{t=1}^L$ be a smooth $L$-layer neural network parameterized by $\theta$. For an input $x \in \mathbb{R}^{N \times 3}$, define the hidden state at layer $t$ as $z^{(t)} = f_t \circ \cdots \circ f_1(x)$. For a perturbed input $x+\epsilon$, with $\|\epsilon\|_\infty \le k$ for sufficiently small $k>0$, define the perturbed hidden state as $Z^{(t)} = f_t \circ \cdots \circ f_1(x+\epsilon)$. 
Then, under the assumption that the perturbation is small and $f \in C^2$, $Z^{(t)}$ can be locally approximated by a Gaussian centered at $z^{(t)}$, with a third-order remainder in the log-density.
\end{lemma}

\begin{proof}
Let $f = \{f_t\}_{t=1}^L$ be a smooth $L$-layer neural network parameterized by $\theta$, and let $z^{(t)} := f_t \circ \cdots \circ f_1(x)$ denote the hidden state at layer $t$ for a clean input $x \in \mathbb{R}^{N \times 3}$. For a perturbed input $x+\epsilon$, with $\|\epsilon\|_\infty \le k$ for small $k>0$, define the perturbed hidden state as $Z^{(t)} := f_t \circ \cdots \circ f_1(x+\epsilon)$.

For the clean and perturbed inputs, define
\begin{equation}
y^* := f(x;\theta) = f^{(t)}(z^{(t)};\theta^{(t)}), \quad 
y := f(x+\epsilon;\theta) = f^{(t)}(z^{(t)} + \epsilon';\theta^{(t)}),
\end{equation}
where $f^{(t)} = f_L \circ \cdots \circ f_{t+1}$, $\theta^{(t)}$ are its parameters, and $\epsilon'$ is the residual vector at layer $t$ induced by the input perturbation $\epsilon$. The perturbation $\epsilon$ is chosen to maximize the adversarial objective $C \|y-y^*\|_2^2$, or equivalently minimize $\exp(-C\|y-y^*\|_2^2)$, under $\|\epsilon\|_\infty \le k$.

Motivated by this, we approximate the conditional distribution of hidden states near $z^{(t)}$ using a local energy-based form,
\begin{equation}
p_\theta(z \mid y^*) \;\propto\; \exp\!\big(-C \,\| f^{(t)}(z;\theta^{(t)}) - f^{(t)}(z^{(t)};\theta^{(t)}) \|_2^2\big),
\end{equation}
for $z$ in a neighborhood of $z^{(t)}$. 
Since $f$ is twice continuously differentiable, the conditional log-density admits a second-order Taylor expansion around $z^{(t)}$:
\begin{equation}
\begin{split}
\log p_\theta(z \mid y^*) 
= \log p_\theta(z^{(t)} \mid y^*) 
+ (z - z^{(t)})^\top \nabla_z \log p_\theta(z \mid y^*)\big|_{z=z^{(t)}} \\
+ \tfrac{1}{2}(z - z^{(t)})^\top H^{(t)} (z - z^{(t)}) + R(z),
\end{split}
\end{equation}
where $H^{(t)} := \nabla_z^2 \log p_\theta(z \mid y^*)|_{z=z^{(t)}}$ is the Hessian and $R(z) = \mathcal{O}(\|z-z^{(t)}\|^3)$.

The first-order term vanishes as follows:
\begin{equation}
\nabla_{z} \log p_\theta(z \mid y^*)\big|_{z=z^{(t)}} 
= -2C \cdot J_{f^{(t)}}(z;\theta^{(t)})^\top 
   \Big(f^{(t)}(z;\theta^{(t)}) - f^{(t)}(z^{(t)};\theta^{(t)})\Big)\big|_{z=z^{(t)}}  = 0.
\end{equation}

Therefore,
\begin{equation}
\log p_\theta(z \mid y^*) 
= \log p_\theta(z^{(t)} \mid y^*) 
+ \tfrac{1}{2}(z - z^{(t)})^\top H^{(t)} (z - z^{(t)}) + R(z).
\end{equation}

The quadratic term coincides with the log-density of a Gaussian centered at $z^{(t)}$ with covariance $(-H^{(t)})^{-1}$, while the remainder $R(z)$ is of order $\mathcal{O}(\|z-z^{(t)}\|^3)$. 

Therefore, the perturbed hidden state $Z^{(t)}$ under small input perturbations can be locally approximated by a Gaussian centered at $z^{(t)}$, with approximation error of third order in the log-density.
\end{proof}

\subsection{Proof of Theorem~\ref{thm:entropy}}\label{append_proof:thm}
\renewcommand{\thetheorem}{3.2}
\begin{theorem}[Upper bound of differential entropy increases as hidden state deviation increases under adversarial attack]\label{thm:entropy}
Let \( x \) be an input image, and let \( \epsilon \) be a small adversarial perturbation. Define the perturbed input as \( X := x + \epsilon \). Let \( f = \{f_t\}_{t=1}^L \) be a smooth \( L \)-block transformer that processes a sequence of \( N \) input tokens. Let \( z^{(t)} := f_t \circ \cdots \circ f_1(x) \in \mathbb{R}^{N \times d} \) and \( Z^{(t)} := f_t \circ \cdots \circ f_1(X) \in \mathbb{R}^{N \times d} \) be the hidden states at layer \( t \) for the clean and perturbed inputs, respectively. Denote the \( i \)-th token representation at layer \( t \) as \( z_i^{(t)} \in \mathbb{R}^{d} \) and \( Z_i^{(t)} \in \mathbb{R}^{d} \).
If \( Z_i^{(t)} \) changes smoothly with small \( \epsilon \), then the upper bound of the differential entropy of \( Z_i^{(t)} \) increases as \( \mathbb{E}_{\epsilon}[\|Z_i^{(t)} - z_i^{(t)}\|_2^2] \) increases.
\end{theorem}

\begin{proof}
Let $x$ be an input image and $\epsilon$ a small perturbation satisfying $\|\epsilon\|_\infty \leq k$, where $k$ is sufficiently small for a first-order Taylor expansion. Define
\begin{equation}
    z^{(t)}_i := f^{(t)}_i(x), \quad Z^{(t)}_i := f^{(t)}_i(x + \epsilon),
\end{equation}
where $f^{(t)}_i$ denotes the hidden state of token $i$ at layer $t$, and $f = f_t \circ \cdots \circ f_1$ is assumed to be twice continuously differentiable.

By the multivariate Taylor expansion of $f^{(t)}_i(x + \epsilon)$ around $x$, we have
\begin{equation}
    Z^{(t)}_i = z^{(t)}_i + J^{(t)}_i \epsilon + R^{(t)}_i(\epsilon),
\end{equation}
where $J^{(t)}_i := \left.\frac{\partial z^{(t)}_i}{\partial x}\right|_{x} \in \mathbb{R}^{d \times D}$ is the Jacobian matrix, and $\|R^{(t)}_i(\epsilon)\| = \mathcal{O}(\|\epsilon\|^2)$.

With the assumption of the perturbation upper bound $k$, the remainder $R^{(t)}_i(\epsilon)$ is negligible compared to the linear term. Under this assumption, we define the deviation:

\begin{equation}
    \Delta Z^{(t)}_i := Z^{(t)}_i - z^{(t)}_i = J^{(t)}_i \epsilon.
\end{equation}

Let $\Sigma_\epsilon := \mathbb{E}[\epsilon \epsilon^\top]$. Then the covariance of $\Delta Z^{(t)}_i$ is
\begin{equation}
    \Sigma_{\Delta Z^{(t)}_i} := \operatorname{Cov}[\Delta Z^{(t)}_i] 
    = J^{(t)}_i \Sigma_\epsilon (J^{(t)}_i)^\top.
\end{equation}

By the local Gaussianity assumption (Lemma~\ref{lem:local_gaussianity}), 
\( Z_i^{(t)} \) can be approximated as a multivariate Gaussian. 
Hence, by the entropy formula for multivariate Gaussians, the differential entropy is
\begin{equation}
    h(Z^{(t)}_i) = \frac{1}{2} \log \left( (2\pi e)^d \cdot \det(\Sigma_{\Delta Z^{(t)}_i}) \right).
\end{equation}

Applying the AM–GM inequality to the eigenvalues of $\Sigma_{\Delta Z^{(t)}_i}$, we obtain
\begin{equation}
    \det(\Sigma_{\Delta Z^{(t)}_i})^{1/d} \leq \frac{1}{d} \operatorname{tr}(\Sigma_{\Delta Z^{(t)}_i})
    = \frac{1}{d} \, \mathbb{E}[\|\Delta Z^{(t)}_i\|_2^2].
\end{equation}

Thus, the entropy is bounded as:
\begin{equation}
    h(Z^{(t)}_i) \leq \frac{d}{2} \log\!\left( \frac{1}{d} \, \mathbb{E}[\|\Delta Z^{(t)}_i\|_2^2] \right) + C,
\end{equation}
where $C = \tfrac{d}{2} \log(2\pi e)$ is a constant.

Hence, the upper bound of the entropy increases as $\mathbb{E}[\|\Delta Z^{(t)}_i\|_2^2]$ increases, which completes the proof.
\end{proof}

\subsection{On practicality of the proved upper bound}
Assuming that the deviation of hidden states follows a Gaussian distribution, the differential entropy of each token is proportional to the determinant of the covariance matrix $\Sigma_{\Delta Z}$. However, our empirical analysis reveals that this covariance matrix is highly low-rank. By decomposing the covariance matrix obtained from 2048 adversarial attacks on the visual tokens of 100 images with LLaVA-1.5-7B~\cite{li2024llava} using PCA, we found that the top 8 components ($8/1024=0.8\%$ of the total dimension) account for 94.2\% ($\pm 0.4\%$) of the total variance, with most eigenvalues close to zero. Under such conditions, computing $\det(\Sigma_{\Delta Z})$ for entropy estimation becomes numerically unstable, as values underflow to zero, making direct entropy comparison infeasible. In contrast, using $\text{tr}(\Sigma_{\Delta Z})$ provides a numerically stable alternative that is theoretically well-grounded under anisotropy and preserves token-wise uncertainty ordering. This trace-based measure also aligns with the qualitative uncertainty maps in Fig.~\ref{fig:mc_vs_ours}, further supporting its practical validity.

\section{Code}
To support reproducibility, we include the implementation of our method in the supplementary material. Detailed instructions for running the code and setting up the environment are provided in the accompanying \texttt{README.md} file.

\section{Implementation Details}\label{append_sec:imple}
As our method is designed to work in conjunction with various LVLMs and existing mitigation methods such as OPERA, VCD, PAI and Devils, we set the value of $\sigma_{\text{th}}$ individually for each combination, as shown in Table~\ref{append_tab:sigma}. The selected $\sigma_{\text{th}}$ values are used consistently to evaluate hallucination performance throughout all experiments in the main paper. As described in Section~\ref{subsec:baselines_imple_details}, PGD-based adversarial attacks are performed with $k = 3$ and 200 iterations. For uncertainty estimation, masks $M$ are extracted from layers $\mathcal{S} = \{1, \dots, 10\}$ of the vision encoder. The masking operation is applied within the self-attention mechanism of the vision encoder, targeting layers 13–17 for LLaVA-1.5 and Shikra, and layers 9–16 for MiniGPT-4. All experiments in the main paper were conducted on an NVIDIA A100 GPU with 80GB of memory.

\begin{table}[h]
    \caption{\textbf{Values of $\sigma_{\text{th}}$ for each model and method combination.} We determine $\sigma_{\text{th}}$ individually for each combination and use the selected value consistently across all evaluations to ensure fair and robust comparisons.}
    \label{append_tab:sigma}
    \centering
    \begin{tabular}{l|ccccc}
    \toprule
    Model     & Greedy & OPERA & VCD & PAI & Devils \\
    \midrule
        LLaVA-1.5-7B & 1.1 & 1.1 & 1.0 & 1.8 & 1.9 \\
        LLaVA-1.5-13B & 1.2 & 1.2 & 1.1 & 1.6 & 1.6 \\
        Shikra-7B & 1.0 & 1.0 & 1.0 & 1.5 & 1.9 \\
        MiniGPT-4 & 0.0 & 2.0 & 0.0 & 1.1 & -0.1 \\
    \bottomrule
    \end{tabular}
\end{table}
\section{Experimental Details}\label{append_sec:exp}
\subsection{Benchmarks}\label{append_sec:benchmark}
\paragraph{CHAIR.} To evaluate the robustness of image captioning models against object hallucination, we adopt the CHAIR~\cite{rohrbach2018object} metric (Caption Hallucination Assessment with Image Relevance). This benchmark quantifies hallucination by comparing generated captions with ground truth object annotations and sentence descriptions in the MSCOCO dataset. Two variants, CHAIR$_i$ and CHAIR$_s$, measure hallucination at the object and sentence levels, respectively, as shown in Eq.~\ref{eq:chair}.

This metric enables a systematic comparison of hallucination severity across models and offers insights into the alignment between visual input and generated language beyond standard evaluation metrics. We use the prompt ``Please describe this image in detail.''.

\paragraph{POPE.} To obtain a more reliable and instruction-agnostic assessment of object hallucination in large vision-language models (LVLMs), we adopt the POPE (Polling-based Object Probing Evaluation) framework~\cite{li2023evaluating}. Unlike traditional caption-based metrics that are sensitive to prompt phrasing and rely on manual parsing, POPE probes a model’s visual grounding through binary yes/no questions about object presence. This enables stable and scalable evaluation across both annotated and unannotated datasets. POPE constructs evaluation sets using three sampling strategies: Random, Popular, and Adversarial. Each strategy targets a different source of hallucination, allowing us to test whether models tend to hallucinate arbitrary objects, frequently occurring objects, or objects that often co-occur with those actually present in the image. We use the prompt ``Is there a/an [object] in the image?''.

\paragraph{AMBER.} To evaluate object hallucination comprehensively in large vision-language models (LVLMs), we adopt the AMBER benchmark~\cite{wang2023llm}. AMBER assesses hallucinations across both generative and discriminative tasks, focusing on three primary types: existence, attribute, and relation. In the generative setting, it employs metrics such as CHAIR, Hal, and Cog to measure hallucination frequency, object coverage, and cognitive tendencies. For discriminative tasks, standard binary classification metrics are used, and the AMBER Score integrates CHAIR from the generative setting with the F1 score from the discriminative setting. Notably, we focus exclusively on `existence' subset to assess object hallucination, which involves generating descriptions of objects that are not present in the input image. We use the prompt ``Describe this image.'' for generative task and ``Is there a [object] in this image?'' for discriminative task.

\subsection{Base models}\label{append_sec:models}
\paragraph{LLaVA-1.5.} In our experiments, we employed LLaVA-1.5~\cite{liu2024improved}, a versatile multimodal model developed for visual instruction tuning. LLaVA-1.5 builds upon the original LLaVA~\cite{liu2023visual} architecture by integrating a two-layer MLP as a vision-language connector, leveraging the \texttt{CLIP-ViT-L-336px}~\cite{radford2021learning} vision encoder, and incorporating academic task-oriented VQA data with response formatting prompts. These modifications significantly enhance the model’s capability for both visual reasoning and instruction following, while retaining strong data efficiency. LLaVA-1.5 achieves competitive performance across a broad set of multimodal benchmarks using only publicly available data and modest computational resources. To investigate the robustness of our method across different model scales, we conducted experiments using both the 7B and 13B versions of LLaVA-1.5. This enabled us to evaluate whether our approach maintains performance consistency under varying model capacities.
For the experiments, we utilized the official implementation~\footnote{\url{https://github.com/haotian-liu/LLaVA}} along with the provided code and model weights.

\paragraph{Shikra.} In our experiments, we adopt the Shikra-7B~\cite{chen2023shikra} model, a LVLM specifically designed for referential dialogue. Shikra-7B integrates a \texttt{CLIP-ViT-L/14}~\cite{radford2021learning} vision encoder with a Vicuna-7B language model via a simple alignment layer, allowing end-to-end processing without the need for additional vocabularies, position encoders, detection modules, or external plug-ins. A key feature of Shikra is its ability to represent spatial information directly in natural language using numerical coordinates, allowing it to handle both inputs and outputs involving region references seamlessly. This architecture supports a broad range of vision-language tasks, including Visual Question Answering (VQA), image captioning, referring expression comprehension (REC), and PointQA, all within a unified framework and without task-specific fine-tuning. Its strong performance across both conventional and location-sensitive tasks makes it a compelling choice for measuring object hallucination.
For the experiments, we utilized the official implementation~\footnote{\url{https://github.com/shikras/shikra}} along with the provided code and model weights.

\paragraph{MiniGPT-4.} In our experiments, we employed MiniGPT-4~\cite{zhu2024minigpt} as a vision-language model to evaluate effectiveness of our method. MiniGPT-4 combines a frozen vision encoder from BLIP-2~\cite{li2023blip} (\texttt{EVA-CLIP-ViT-G/14}~\cite{sun2023eva} with Q-Former) and a large frozen language model, Vicuna, using a single trainable linear projection layer to align visual features with the input space of the language model. The model is pre-trained on approximately 5 million image-text pairs to establish initial multimodal capabilities. To address issues such as repetitive or fragmented outputs observed after pretraining, a second stage fine-tuning is applied using a curated set of 3,500 detailed image-description pairs, formatted with a conversational prompt template. This two-stage training strategy improves the fluency and relevance of the model’s responses, enabling it to handle a variety of vision-language tasks more effectively.
When applying our methodology to MiniGPT-4, we conducted the adversarial attack on the features prior to their input into the Q-Former. For the experiments, we utilized the official implementation~\footnote{\url{https://github.com/Vision-CAIR/MiniGPT-4}} along with the provided code and model weights.

\subsection{Baselines}\label{append_sec:baselines}
\paragraph{Greedy.}
Greedy decoding is one of the most basic decoding strategies for generative language models, where the token with the highest prediction probability is selected at each step. This approach is fast and straightforward to implement. Among various decoding strategies for LVLMs, we adopt the na\"ive and fundamental greedy decoding method as one of our baselines to evaluate the object hallucination mitigation performance of our method.

\paragraph{OPERA.}
The authors of OPERA~\cite{huang2024opera} identify that object hallucination in LVLMs is closely linked to specific knowledge aggregation patterns within the model’s self-attention matrix. It defines tokens that induce such attention patterns as summary tokens and mitigates hallucination by detecting excessive attention toward these tokens and preventing their influence on next-token prediction. Specifically, OPERA extracts a local window from the self-attention map, quantifies the degree of aggregation via column-wise multiplication, and applies a logit penalty during beam search to suppress over-confident candidates. While effective, OPERA relies on beam search, which introduces significant additional computational cost. For comparison and integration with our method, we used the official implementation~\footnote{\url{https://github.com/shikiw/OPERA}} provided by the authors.

\paragraph{Visual Constrastive Decoding.}
The authors of Visual Contrastive Decoding (VCD)~\cite{qi2024visual} attribute object hallucination to statistical biases, such as object cooccurrence frequencies in training data, and language priors inherent to large language models. By injecting Gaussian noise into the input image, the LVLM's reliance on visual information is reduced, causing it to lean more heavily on these language priors. To counteract this, VCD introduces both the original image $v$ and a distorted version $v'$ as input, computes their respective output probability distributions, and then extrapolates a contrastive probability distribution that suppresses language-driven biases. For comparison and integration with our method, we use the official implementation~\footnote{\url{https://github.com/DAMO-NLP-SG/VCD}}. When applying our method to VCD, we performed uncertain token suppression only on the original image $v$.

\paragraph{PAI.}
The authors of Paying more Attention to Image (PAI)~\cite{liu2024paying} argue that object hallucination arises when visual information is ignored and propose a training-free method to enhance the influence of images during inference. Specifically, they manipulate the self-attention matrix to amplify attention toward visual tokens and selectively strengthen particular attention heads to guide the model toward more trustworthy directions. To avoid excessive attention toward the beginning-of-sentence (BOS) token, they introduce a layer prior that excludes shallow layers from modulation. Additionally, they compare outputs with and without the input image to attenuate language model biases. Since PAI does not modify the vision encoder, our method can be additionally applied. For comparison, we utilized the official implementation~\footnote{\url{https://github.com/LALBJ/PAI}}.

\paragraph{Devils in the middle layers.}
In Devils in the Middle Layers (Devils)~\cite{jiang2024devils}, the authors find that in large vision-language models (LVLMs), visual information is strongly processed in the middle layers of the language model. They observe that inactive attention can induce hallucinations, and that during such instances, attention heads tend to focus inconsistently on unrelated objects. To address this, the authors propose integrating information across attention heads during inference to encourage focus on more consistent visual regions. They achieve this by reweighting the attention scores to emphasize coherent areas. Since this is an intervention on the LLM component, their methodology is applicable in our setting as well. To implement it, we adopted their official codebase~\footnote{\url{https://github.com/ZhangqiJiang07/middle_layers_indicating_hallucinations}}.

\section{Additional Analysis}
\subsection{Monte Carlo vs. Adversarial attack}\label{append_sec:mc_vs_ours}
\begin{table}
    \caption{\textbf{Runtime comparison between MC dropout and our method using PGD-based adversarial attack.} When comparing the mean runtime, our method is $\times 5.1$ faster. The symbol $\pm$ denotes the $1\sigma$ interval.}
    \label{append_tab:mc_ours_efficiency}
    \centering
    \begin{tabular}{c|cc}
    \toprule
        Method & MC dropout & Adversarial attack (Ours)  \\
    \midrule
        Time (s) & 12.4 ($\pm 0.12$) & 2.43 ($\pm 0.08$)  \\
    \bottomrule
    \end{tabular}
\end{table}

\begin{figure}
    \centering
    \includegraphics[width=\linewidth]{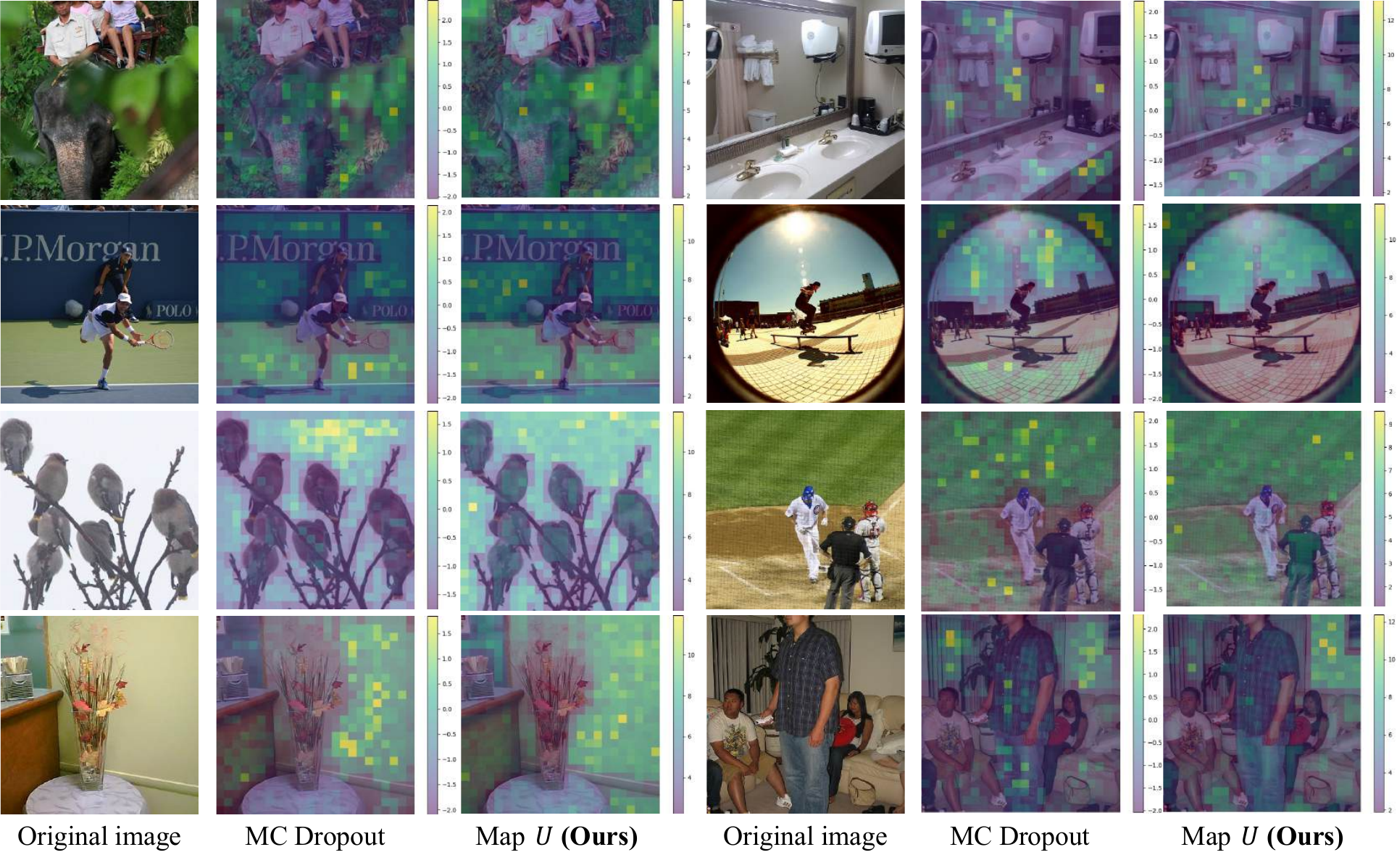}
    \caption{\textbf{Additional visual comparison of estimated uncertainty from MC dropout and our method.} Our uncertainty map $U$ identifies uncertain regions similar to the uncertainty map obtained via MC dropout. MC dropout was applied to the self-attention and MLP residuals of the LLaVA-1.5 vision encoder with a dropout rate of $p=0.5$. Layers $i=5$ to $j=8$ were used, and the variance of each token was estimated over 1,000 forward passes. For the adversarial attack, we applied 100 iterations of PGD with $k=3$. The MC-based uncertainty values were log-scaled for visualization clarity.}
    \label{append_fig:mc_vs_ours}
\end{figure}

In the main paper, we verify the similarity between the uncertainty map $U$ obtained via adversarial attacks and the one derived from the Monte Carlo (MC) dropout using pre-trained vision encoder. To further confirm this similarity, we provide an additional qualitative comparison in Fig.~\ref{append_fig:mc_vs_ours}. Although our method tends to slightly overestimate the uncertainty, it consistently identifies high-uncertainty regions that closely align with those highlighted by MC dropout.
To assess the computational efficiency of our approach, we compare the runtime of uncertainty estimation using Monte Carlo dropout and our adversarial-based method. Specifically, we apply both techniques to the vision encoder from LLaVA-1.5-7B. The adversarial attack is performed 100 times with $k=3$ top perturbations, while the Monte Carlo dropout requires 1,000 forward passes, both executed on a single NVIDIA RTX 4090 GPU. The results, presented in Table~\ref{append_tab:mc_ours_efficiency}, demonstrate that our method enables significantly more efficient extraction of uncertainty masks, highlighting its practical advantage in identifying visually uncertain tokens.

\subsection{Effect of adversarial attacks on LVLM outputs}
We conducted PGD-based adversarial attacks on the vision encoder to identify the uncertain visual tokens. To evaluate whether such attacks effectively influence the output of LVLMs, we applied adversarial perturbations with varying magnitudes of $\epsilon$ and performed both quantitative and qualitative analyses.

\begin{figure}[!t]
    \centering
    \includegraphics[width=\linewidth]{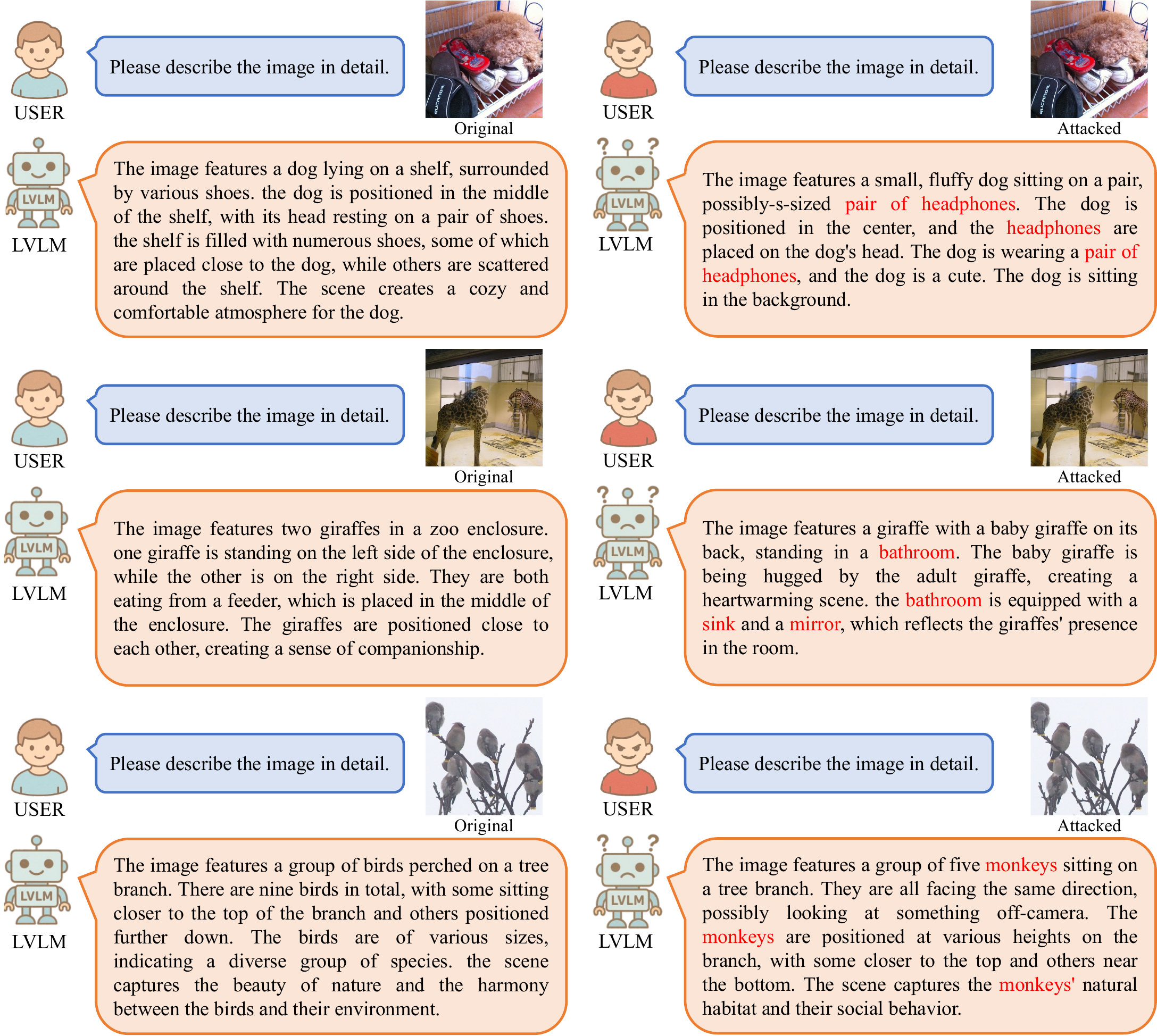}
    \caption{\textbf{Qualitative comparison of LVLM outputs when using the original image versus the adversarially attacked image as input.} When conditioned on the attacked image (right column), the model generates descriptions involving non-existent objects or scenes, indicating a more severe object hallucination effect compared to the original image (left column). This experiment was conducted on the LLaVA-1.5-7B model using the CHAIR dataset, with $k = 3$ and 200 PGD iterations.}
    \label{append_fig:attack_hallucination}
\end{figure}

As shown in Fig.~\ref{append_fig:attack_hallucination}, the responses generated from the attacked images often exhibited hallucinations or failed to produce correct answers. As demonstrated in Table~\ref{append_tab:attack_result}, we also observe that higher attack intensities lead to increased severity of hallucinations. These experimental results highlight that the visual features extracted by the vision encoder play a crucial role in LVLMs' performance of downstream task, emphasizing that enhancing visual perception is critical for reducing hallucination and improving overall reliability.

\begin{table}[!t]
    \caption{\textbf{Object hallucination benchmark results under varying attack strengths ($\|\epsilon\|_{\infty}$)}. To investigate the effect of adversarial perturbations on the image encoder, we applied PGD attacks of different magnitudes for 200 iterations to LLaVA-1.5-7B and evaluated performance using the CHAIR benchmark. Adversarial attacks on the image encoder increase the likelihood of hallucinated outputs, with the severity of hallucination correlating positively with the attack strength.}
    \label{append_tab:attack_result}
    \centering
    \begin{tabular}{c|ccccc}
        \toprule
        $\|\epsilon\|_{\infty}$ & CHAIR$_s\downarrow$ & CHAIR$_i\downarrow$ & Recall$\uparrow$ & Precision$\uparrow$ & F1$\uparrow$  \\
        \midrule
        0 & 47.4 & 12.2 & 78.9 & 76.9 & 77.9 \\
        1 & 53.0 & 16.2 & 76.9 & 72.9 & 74.8 \\
        3 & 64.0 & 25.5 & 63.0 & 62.4 & 62.7 \\
        5 & 65.6 & 25.9 & 55.9 & 60.1 & 57.9 \\
        7 & 61.6 & 26.6 & 50.5 & 59.6 & 54.7 \\
        \bottomrule
    \end{tabular}
\end{table}

\subsection{Consistency and robustness of uncertainty masks from adversarial attacks}
We identify uncertain visual tokens by applying PGD-based adversarial attacks to the features of the vision encoder. In our implementation, the attack is initialized from the original image without added noise. To evaluate the consistency and robustness of the resulting uncertainty masks $M$, we also perform attacks with different initial noise seeds, generating diverse adversarial perturbations. From each perturbed image, we extract a mask and compute the mean Intersection over Union (mIoU) between the masks $M$ generated from different seeds.

\begin{figure}[!t]
    \centering
    \includegraphics[width=\linewidth]{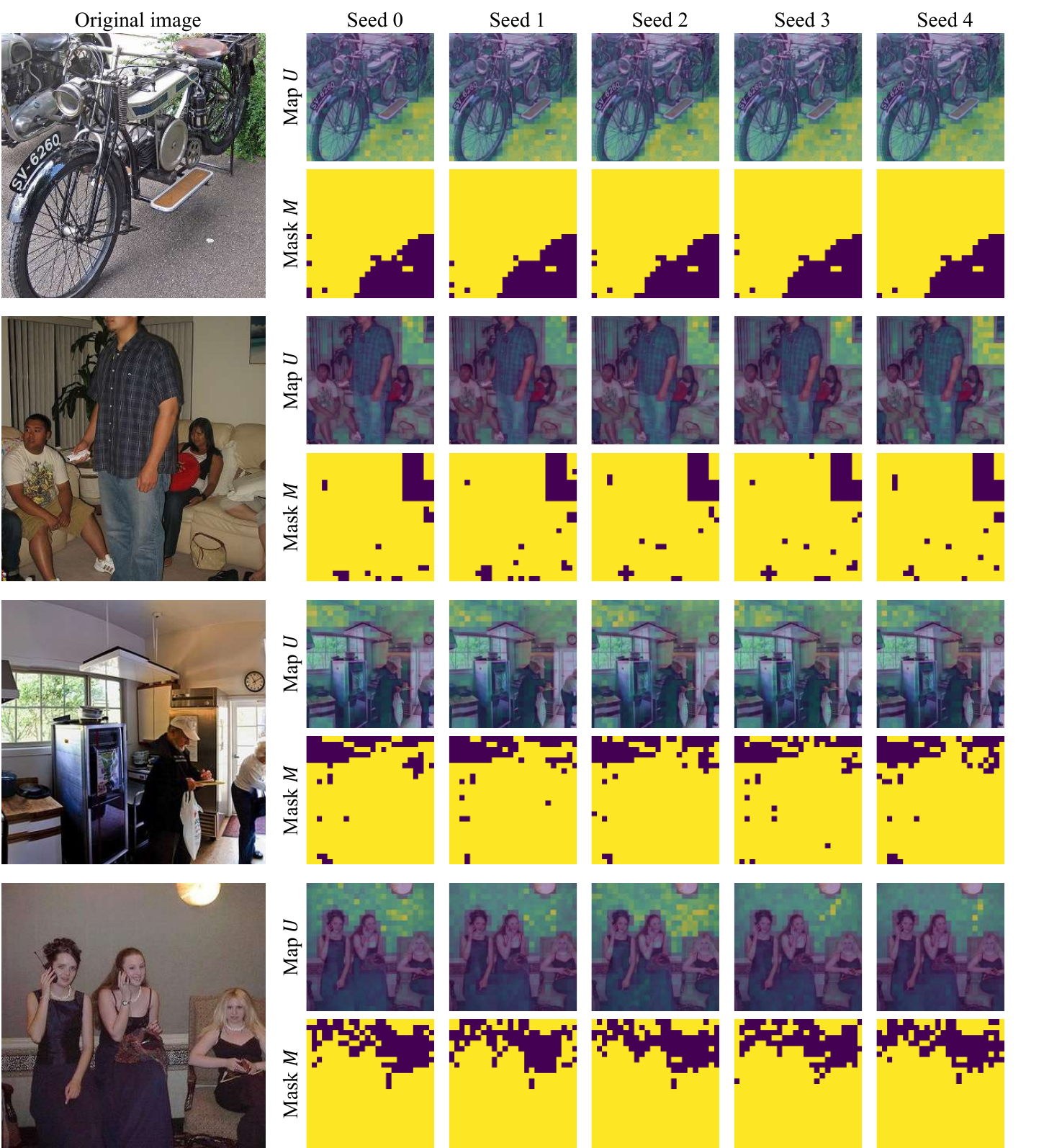}
    \caption{\textbf{Uncertainty maps $U$ and masks $M$ generated from different initial noises using PGD-based adversarial attacks.} We qualitatively demonstrate the consistency of the uncertainty maps $U$ and corresponding masks $M$ ($\sigma_{\text{th}}=1.1$), generated using PGD-based adversarial attacks with five different random seeds for initializing noise. Despite variations in the initial noise, the resulting uncertainty maps $U$ and masks $M$ remain highly similar, highlighting the robustness and stability of the attack-based uncertainty estimation in LLaVA-1.5-7B vision encoder.}
    \label{append_fig:seed_mask}
\end{figure}

As shown in Table~\ref{append_tab:mask_consist}, the uncertainty masks $M$ remain highly consistent across different initializations. Qualitative examples in Fig.~\ref{append_fig:seed_mask} further demonstrate that the uncertainty maps $U$ and masks $M$ maintain stable and coherent structures. These results confirm the reliability of our method in consistently identifying uncertain tokens under varying adversarial conditions.

\begin{table}[!t]
    \caption{\textbf{Mask consistency measured by mean Intersection over Union (mIoU).} We applied adversarial attacks to the LLaVA-1.5-7B image encoder on 500 images across five different seeds and measured the mIoU to verify mask consistency. The results indicate that the masks obtained through adversarial attack are robust and consistent. The threshold $\sigma_{\text{th}}$ was set to $1.1$.}
    \label{append_tab:mask_consist}
    \centering
    \resizebox{\textwidth}{!}{
    \begin{tabular}{c|ccccc}
    \toprule
    Seed pair & (0, 1) & (0, 2) & (0, 3) & (0, 4) & (1, 2) \\
    \midrule
    mIoU  & 0.899 ($\pm 0.034$) & 0.898 ($\pm 0.035$) & 0.898 ($\pm 0.036$) & 0.899 ($\pm 0.036$) & 0.899($\pm 0.035$)  \\
    \bottomrule
    \end{tabular}
    }
    \resizebox{\textwidth}{!}{
    \begin{tabular}{c|ccccc}
    \toprule
    Seed pair & (1, 3) & (1, 4) & (2, 3) & (2, 4) & (3, 4) \\
    \midrule
    mIoU  & 0.898 ($\pm 0.036$) & 0.898 ($\pm 0.035$) & 0.897 ($\pm 0.036$) & 0.897 ($\pm 0.036$) & 0.897 ($\pm 0.036$)  \\
    \bottomrule
    \end{tabular}
    }
\end{table}

\section{Additional Ablation Studies}
\paragraph{Masking Threshold Hyperparameter $\sigma_{\text{th}}$.}
To construct the binary uncertainty mask $M$, we introduce a threshold hyperparameter $\sigma_{\text{th}}$. Its optimal value depends on the characteristics of each model and method combination, and is determined through grid search. Table~\ref{append_tab:sigma_ablation} presents an ablation study conducted on the LLaVA-1.5-7B model using six different threshold values. Considering the trade-offs among $C_s$, $C_i$, and F1 score, we select $\sigma_{\text{th}} = 1.1$ as it yields the best overall performance. Based on this analysis, we apply the optimal $\sigma_{\text{th}}$ for each configuration in our experiments.

\begin{table}[]
    \caption{\textbf{Ablation study of the thresholding parameter $\sigma_{\text{th}}$ for generating the uncertainty mask $M$.} We use LLaVA-1.5-7B with greedy decoding and evaluate hallucination performance while varying the threshold $\sigma_{\text{th}}$.}
    \label{append_tab:sigma_ablation}
    \centering
    \begin{tabular}{l|ccccccc}
      \toprule
      $\sigma_{\text{th}}$ & Greedy & 0.8 & 0.9 & 1.0 & 1.1 & 1.2 & 1.3  \\
      \midrule
      $C_s$$\downarrow$ &47.4 & 27.0 & 27.0 & 30.0 & \cellcolor{lightyellow}29.2 & 33.6 & 36.4 \\
      $C_i$$\downarrow$ &12.2 & 8.4 & 8.2 & 9.0 & \cellcolor{lightyellow}9.3 & 9.7 & 10.3 \\
      F1$\uparrow$     &77.9 & 76.7 & 77.7 & 77.6 & \cellcolor{lightyellow}78.2 & 78.0 & 78.5 \\
      \bottomrule
    \end{tabular}
\end{table}

\section{Additional Quantitative and Qualitative Results}
\begin{table}[]
    \caption{\textbf{Additional inference time introduced by each method compared to standard greedy decoding.} We performed text generation with request of image description with max 32 tokens. All experiments were conducted using LLaVA-1.5-7B on an NVIDIA A100 GPU. We report the mean and standard deviation over 30 samples. Although our method introduces some overhead due to backpropagation from PGD attacks, it remains comparable to or even faster than existing approaches.}
    \label{append_tab:inference_time}
    \centering
    \begin{tabular}{c|c}
    \toprule
        Method & Additional inference time (s) \\
    \midrule
        OPERA & 9.518$\pm 0.011$ \\
        VCD & 1.646$\pm 0.001$ \\
        PAI & 1.567$\pm 0.021$ \\
        Devils & 0.014$\pm 0.001$ \\
        Ours & \cellcolor{lightyellow}2.469$\pm 0.004$ \\
    \bottomrule
    \end{tabular}
\end{table}

\subsection{Computational Cost}
Our method identifies uncertain tokens via PGD-based adversarial attacks implemented through backpropagation, which naturally introduces additional computational overhead compared to standard greedy decoding. To quantify this cost, we measure the extra inference time and compare it with existing hallucination mitigation methods. As shown in Table~\ref{append_tab:inference_time}, while our method does incur some additional overhead, it offers comparable or even lower inference time than several baselines, achieving a favorable balance between performance and efficiency.

\subsection{Additional quantitative results}\label{append_sec:quanti}
\paragraph{Applicability of our method to larger model.} We assess the scalability and generalizability of our method using the larger LLaVA-1.5-13B model. As shown in Table~\ref{append_tab:llava13b_chair}, our method delivers substantial improvements over the greedy decoding baseline, reducing $C_s$ by 15.2 and $C_i$ by 2.9. It also integrates effectively with a variety of existing approaches, achieving the best performance when combined with Devils ($C_s = \textbf{20.4}$, $C_i = \textbf{6.0}$). These results demonstrate that our method generalizes well across model scales and enhances a wide range of existing hallucination mitigation strategies.
\begin{table}[!t]
  \centering
  \caption{\textbf{Quantitative results on CHAIR benchmark for LLaVA-1.5-13B.} We report object hallucination ($C_s$, $C_i$) for various mitigation methods and their combination with our method. The maximum token length is set to 512. $\Delta$\% denotes the relative improvement in performance.}
  \label{append_tab:llava13b_chair}
  \renewcommand{\arraystretch}{1.2}
  \setlength{\tabcolsep}{5pt}
  \resizebox{\textwidth}{!}{
  \begin{tabular}{l|ccc|ccc|ccc|ccc|ccc}
    \toprule
     \multirow{2}{*}{Method} & \multicolumn{3}{c}{Greedy} & \multicolumn{3}{c}{OPERA} & \multicolumn{3}{c}{VCD} & \multicolumn{3}{c}{PAI} & \multicolumn{3}{c}{Devils} \\
    \cmidrule(lr){2-4} \cmidrule(lr){5-7} \cmidrule(lr){8-10} \cmidrule(lr){11-13} \cmidrule(lr){14-16}
    & Orig. & +Ours & $\Delta$\% & Orig. & +Ours& $\Delta$\% & Orig. & +Ours& $\Delta$\% & Orig. & +Ours& $\Delta$\% & Orig. & +Ours& $\Delta$\% \\
    \midrule
    $C_s$ $\downarrow$ &
    45.4 & \cellcolor{lightyellow}30.2 & {\small\textcolor{babyblue}{$\uparrow$33.4\%}} &
    40.2 & \cellcolor{lightyellow}30.4 & {\small\textcolor{babyblue}{$\uparrow$24.4\%}} &
    49.0 & \cellcolor{lightyellow}35.4 & {\small\textcolor{babyblue}{$\uparrow$27.8\%}} &
    38.6 & \cellcolor{lightyellow}32.4 & {\small\textcolor{babyblue}{$\uparrow$16.1\%}} &
    28.2 & \cellcolor{lightyellow}\textbf{20.4} & {\small\textcolor{babyblue}{$\uparrow$26.2\%}} \\
    $C_i$ $\downarrow$ &
    11.2 & \cellcolor{lightyellow}8.3 & {\small\textcolor{babyblue}{$\uparrow$25.9\%}} &
    10.9 & \cellcolor{lightyellow}8.9 & {\small\textcolor{babyblue}{$\uparrow$18.3\%}} &
    13.4 & \cellcolor{lightyellow}10.3 & {\small\textcolor{babyblue}{$\uparrow$23.1\%}} &
    9.9 & \cellcolor{lightyellow}8.4 & {\small\textcolor{babyblue}{$\uparrow$15.2\%}} &
    8.7 & \cellcolor{lightyellow}\textbf{6.0} & {\small\textcolor{babyblue}{$\uparrow$31.0\%}} \\
    F1 $\uparrow$ &
    \textbf{79.1} & \cellcolor{lightyellow}78.9 & {\small\textcolor{softgray}{$\downarrow$0.2\%}} & 
    78.0 & \cellcolor{lightyellow}76.9 & {\small\textcolor{softgray}{$\downarrow$1.4\%}} &
    77.3 & \cellcolor{lightyellow}76.3 & {\small\textcolor{softgray}{$\downarrow$1.3\%}} &
    78.7 & \cellcolor{lightyellow}\textbf{79.1} & {\small\textcolor{babyblue}{$\uparrow$0.3\%}} &
    78.4 & \cellcolor{lightyellow}78.0 & {\small\textcolor{softgray}{$\downarrow$0.5\%}}  \\
    \bottomrule
  \end{tabular}
  }
\end{table}

\begin{table}[!t]
\centering
    \centering
    \caption{\textbf{Quantitative results of our method on state-of-the-art LVLMs.} We apply our approach to two SOTA models, DeepSeek-VL and Qwen2.5-VL, and compare performance against greedy decoding. For DeepSeek-VL we set $\sigma_{\text{th}}=1.0$, while for Qwen2.5-VL we use $\sigma_{\text{th}}=0.0$. These results demonstrate that our method is applicable to a wide range of LVLMs, including the most recent architectures.}
    \label{append_tab:sota_models}
    \begin{tabular}{l|cccccc}
      \toprule
      \multirow{2}{*}{Method} & \multicolumn{3}{c}{CHAIR} & \multicolumn{3}{c}{POPE} \\
      \cmidrule{2-7}
      & $C_s$$\downarrow$ & $C_i$  $\downarrow$ & F1 $\uparrow$ & Rand. & Pop. & Adv. \\
      \midrule
      DeepSeek-VL (Greedy) & 25.8 & 6.6 & \textbf{72.7} & 88.7 & 88.0 & 84.9\\
      +Ours &
      \cellcolor{lightyellow}\textbf{22.4} & \cellcolor{lightyellow}\textbf{5.5} & \cellcolor{lightyellow}72.6 & \cellcolor{lightyellow}\textbf{88.8} & \cellcolor{lightyellow}\textbf{88.0} & \cellcolor{lightyellow}\textbf{85.1} \\
      \midrule
      Qwen2.5-VL (Greedy) & 29.6 & 7.8 & 76.0 & 84.2 & 83.7 & 83.3\\
      +Ours &
      \cellcolor{lightyellow}\textbf{28.6} & \cellcolor{lightyellow}\textbf{7.0} & \cellcolor{lightyellow}\textbf{76.8} & \cellcolor{lightyellow}\textbf{84.3} & \cellcolor{lightyellow}\textbf{83.8} & \cellcolor{lightyellow}\textbf{83.4}
\\
      \bottomrule
    \end{tabular}
\end{table}

\paragraph{Applicability of our method to the state-of-the-art models.}
In the main paper, we conducted extensive experiments on LLaVA-1.5, Shikra, and MiniGPT, which are commonly used as target models in object hallucination mitigation studies and therefore served as our primary evaluation benchmarks. To further validate the applicability of our approach, we additionally evaluated state-of-the-art models such as  DeepSeek-VL~\cite{lu2024deepseek} and Qwen2.5-VL~\cite{bai2025qwen2}. These models not only demonstrate strong performance, but also involve joint fine-tuning of the vision encoder during vision-language alignment training, making them suitable indicators of the scalability of our method. The results presented in \cref{append_tab:sota_models} confirm that our approach effectively reduces object hallucination even in these latest models.

\begin{table}[!t]
\centering
    \centering
    \caption{\textbf{Additional quantitative results for an alternative adversarial attack on a Q-Former–based LVLM architecture.} MiniGPT-4 uses a Q-Former to effectively compress image tokens, which confers robustness to image-only perturbations. By jointly perturbing the Q-Former’s learnable query vectors together with the image, we enable a stronger attack and observe additional gains in attack effectiveness.}
    \label{append_tab:q_former_alternative}
    \begin{tabular}{l|ccc}
      \toprule
      Method & $C_s$$\downarrow$ & $C_i$  $\downarrow$ & F1 $\uparrow$ \\
      \midrule
      Greedy (MiniGPT-4) & 31.0 & 11.4 & 67.3 \\
      +Ours (Image only) & 29.0 & 10.6 & 67.5 \\
      +Ours (Image + Query) & \cellcolor{lightyellow}\textbf{27.0} & \cellcolor{lightyellow}\textbf{9.3} & \cellcolor{lightyellow}\textbf{68.1} \\
      \bottomrule
    \end{tabular}
\end{table}

\paragraph{Alternative attack methods on Q-Former design architecture.}
We observed that adversarial attacks applied solely to the image have limited effectiveness in Q-Former based architectures (\textit{e.g.}, MiniGPT-4). This appears to stem from the robustness introduced by the architectural design that relies on learnable queries. To validate this hypothesis, we additionally optimized the input queries during adversarial attacks to examine whether our approach provides further advantages. Unlike images, the query vectors are continuous, and thus we imposed a noise constraint on the query vector $q$ such that the perturbation scale matches that applied to the image.
\begin{equation}
    \|\epsilon_q\|_{\infty} = \frac{\|\epsilon\|_{\infty}}{255}\cdot\frac{(\max{(q)}-\min(q))}{2},
\end{equation}
where $\epsilon_q$ is the adversarial noise injected to query vectors $q$, $\epsilon$ is the noise added to the victim image.
The results are presented in~\cref{append_tab:q_former_alternative}, which report the outcomes of adversarial attacks jointly applied to both the image and the Q-Former queries. The evaluation on the CHAIR benchmark demonstrates that our method can achieve further performance improvements when combined with additional architectural considerations. However, for methodological consistency, the main paper focuses only on adversarial perturbations applied to the image.

\paragraph{Length of generated text.}
\cite{yue2024less} highlights that overly long outputs from LVLMs often lead to object hallucinations, as the generated content exceeds the model’s visual perception. As shown in Table~\ref{append_tab:length}, our method consistently and slightly reduces the length of image descriptions across various models and hallucination mitigation methods. However, in the case of MiniGPT-4, due to its Q-Former architecture, masking uncertain visual tokens within the vision encoder is less effective. As a result, the generated text length may occasionally remain unchanged or even slightly increase.

\begin{table}[!t]
  \centering
  \caption{\textbf{Average length of generated text with standard deviation.} We report the average length of generated texts across different models and hallucination mitigation methods, with and without our approach. Values are presented as mean~$\pm$~standard deviation. Our method slightly reduces output length, which has been linked to lower hallucination rates in LVLMs.}
  \label{append_tab:length}
  \resizebox{\textwidth}{!}{
  \begin{tabular}{l|cc|cc|cc|cc|cc}
    \toprule
    \multirow{2}{*}{Model} & \multicolumn{2}{c}{Greedy} & \multicolumn{2}{c}{OPERA} & \multicolumn{2}{c}{VCD} & \multicolumn{2}{c}{PAI} & \multicolumn{2}{c}{Devils} \\
    \cmidrule(lr){2-3} \cmidrule(lr){4-5} \cmidrule(lr){6-7} \cmidrule(lr){8-9} \cmidrule(lr){10-11}
    & Orig. & +Ours & Orig. & +Ours & Orig. & +Ours & Orig. & +Ours & Orig. & +Ours \\
    \midrule
    LLaVA-7B
    & 491$\pm104$
     & \cellcolor{lightyellow}426$\pm105$ & 473$\pm107$
     & \cellcolor{lightyellow}406$\pm118$ & 517$\pm114$
     & \cellcolor{lightyellow}420$\pm121$ & 514$\pm118$
     & \cellcolor{lightyellow}487$\pm120$ & 504$\pm206$
     & \cellcolor{lightyellow}448$\pm173$ \\
    LLaVA-13B
    & 495$\pm101$
     & \cellcolor{lightyellow}440$\pm 114$ & 452$\pm 136$
     & \cellcolor{lightyellow}402$\pm 142$ & 515$\pm 108$
     & \cellcolor{lightyellow}436$\pm 126$ & 510$\pm 122$
     & \cellcolor{lightyellow}468$\pm 115$ & 406$\pm 141$
     & \cellcolor{lightyellow}381$\pm 124$ \\
    Shikra-7B
    & 514$\pm 110$
     & \cellcolor{lightyellow}475$\pm 108$ & 370$\pm 120$
     & \cellcolor{lightyellow}354$\pm 109$ & 524$\pm 113$
     & \cellcolor{lightyellow}487$\pm 113$ & 493$\pm 195$
     & \cellcolor{lightyellow}427$\pm 213$ & 383$\pm 202$
     & \cellcolor{lightyellow}368$\pm 265$ \\
    MiniGPT-4
    & 408$\pm 206$
     & \cellcolor{lightyellow}418$\pm 202$ & 301$\pm 135$
     & \cellcolor{lightyellow}304$\pm 110$ & 404$\pm 167$
     & \cellcolor{lightyellow}404$\pm 172$ & 284$\pm 126$
     & \cellcolor{lightyellow}282$\pm 130$ & 415$\pm 444$
     & \cellcolor{lightyellow}391$\pm 389$ \\
    \bottomrule
  \end{tabular}
  }
\end{table}

\paragraph{Application of our method to other baselines.}
To validate the generalizability of our method for mitigating object hallucination in LLaVA-1.5-7B, we apply it to alternative decoding strategies, including beam search decoding~\cite{sutskever2014sequence}, DoLa~\cite{chuang2024dola} and VAR~\cite{kang2025see}, using the CHAIR dataset. As shown in Table~\ref{append_tab:other_baseline}, our method consistently reduces hallucination rates while maintaining or even improving the F1 score.

\begin{table}[!t]
\centering
\begin{minipage}[t]{0.48\linewidth}
    \caption{\textbf{Effectiveness of our method applied to different decoding baselines.} We evaluate our method on LLaVA-1.5-7B using various decoding strategies, including greedy decoding, beam search, DoLa and VAR. We set the $N_{beam}=5$. Across all settings, our method consistently reduces hallucination metrics ($C_s$, $C_i$) while maintaining or improving F1 score.}
    \label{append_tab:other_baseline}
    \centering
    \begin{tabular}{l|ccc}
      \toprule
      Method & $C_s$$\downarrow$ & $C_i$  $\downarrow$ & F1 $\uparrow$ \\
      \midrule
      Greedy & 47.4 & 12.2 & 77.9 \\
      +Ours & \cellcolor{lightyellow}29.2 & \cellcolor{lightyellow}9.3 & \cellcolor{lightyellow}78.2 \\
      \midrule
      Beam search & 47.2 & 12.7 & 77.8 \\
      +Ours & \cellcolor{lightyellow}28.2 & \cellcolor{lightyellow}8.6 & \cellcolor{lightyellow}78.5 \\
      \midrule
      DoLa & 46.0 & 12.2 & 78.5 \\
      +Ours & \cellcolor{lightyellow}30.4 & \cellcolor{lightyellow}9.5 & \cellcolor{lightyellow}78.2 \\
      \midrule
      VAR & 46.8 & 12.5 & 77.9 \\
      +Ours & \cellcolor{lightyellow}29.4 & \cellcolor{lightyellow}9.1 & \cellcolor{lightyellow}78.1 \\
      \bottomrule
    \end{tabular}
\end{minipage}
\hfill
\begin{minipage}[t]{0.48\linewidth}
    \caption{\textbf{Comparison of uncertainty estimation methods for generating mask $M$.} We evaluate the effectiveness of our adversarial attack-based uncertainty estimation method against MC dropout on LLaVA-1.5-7B using the CHAIR dataset.}
    \label{append_tab:mc_mask_result}
    \centering
    \begin{tabular}{l|ccc}
      \toprule
      Method & $C_s$$\downarrow$ & $C_i$  $\downarrow$ & F1 $\uparrow$ \\
      \midrule
      Greedy & 47.4 & 12.2 & 77.9  \\
      +Ours (w/Adv. attack) & 29.2 & 9.3 & 78.2  \\
      +Ours (w/MC dropout) & 32.6 & 10.5 & 77.8 \\
      \bottomrule
    \end{tabular}
\end{minipage}
\end{table}

\paragraph{Comparison of uncertainty estimation of visual token: Our Method vs. MC Dropout.}
Epistemic uncertainty of visual tokens introduced by a pre-trained vision encoder can be estimated using MC Dropout. However, this approach often requires intensive computation due to thousands of forward passes. As an efficient alternative, we propose a method that estimates uncertainty of visual tokens using PGD-based adversarial attacks.

We perform experiments on LLaVA-1.5-7B using the CHAIR dataset and compare the uncertainty masks $M$ for visual tokens, generated using Eq.\ref{eq:mask_gen}, between our method and MC Dropout. As shown in Table~\ref{append_tab:mc_mask_result}, our approach achieves comparable or better performance while being more computationally efficient. These results highlight that our PGD-based uncertainty estimation effectively captures the epistemic uncertainty of the pre-trained vision encoder and reliably identifies uncertain visual tokens.

Regarding the lower performance of MC dropout compared to our method, we conjecture that although MC dropout is widely used for uncertainty quantification, it remains only one estimation technique. In contrast, our approach provides a more conservative estimate of uncertainty through an upper bound, which we believe accounts for its superior performance.

\subsection{Additional qualitative results}\label{append_sec:quali}
\begin{figure}
    \centering
    \includegraphics[width=\linewidth]{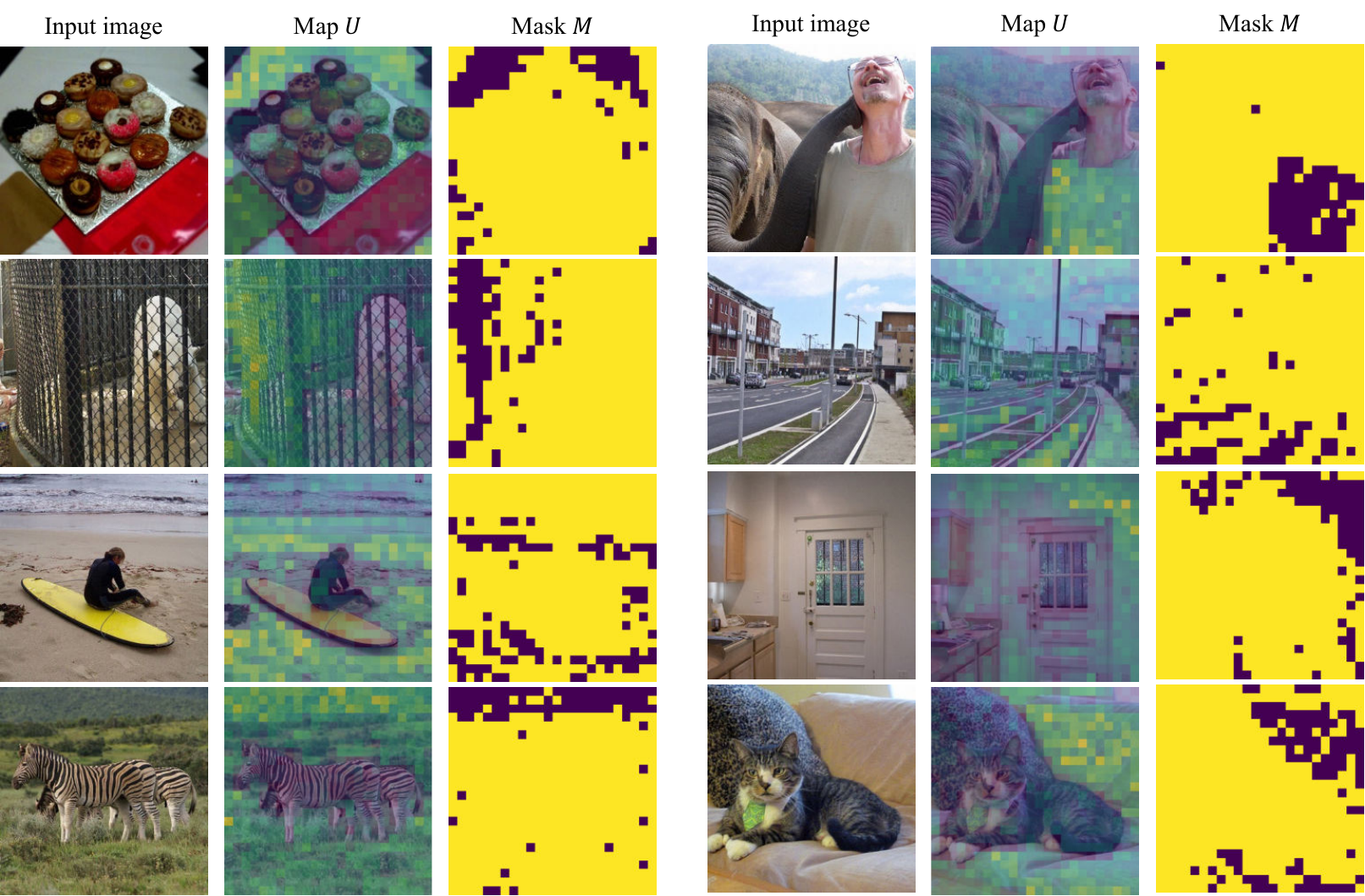}
    \caption{\textbf{Additional qualitative examples of uncertainty maps $U$ and masks $M$ using PGD-based adversarial attacks.} We visualize the uncertainty maps $U$ and the corresponding masks $M$ (with $\sigma_{\text{th}}=1.1$) generated from adversarially perturbed inputs across various samples. All examples were obtained using the LLaVA-1.5-7B vision encoder.}
    \label{append_fig:mask_example}
\end{figure}

\paragraph{Qualitative examples of binary uncertainty masks $M$.}~\label{appendix:Mask_addi_examples}
Fig.~\ref{append_fig:mask_example} presents additional examples of binary uncertainty masks $M$ generated for various input images under PGD-based adversarial attacks applied to the vision encoder of LLaVA-1.5-7B.

\begin{figure}
    \centering
    \includegraphics[width=\linewidth]{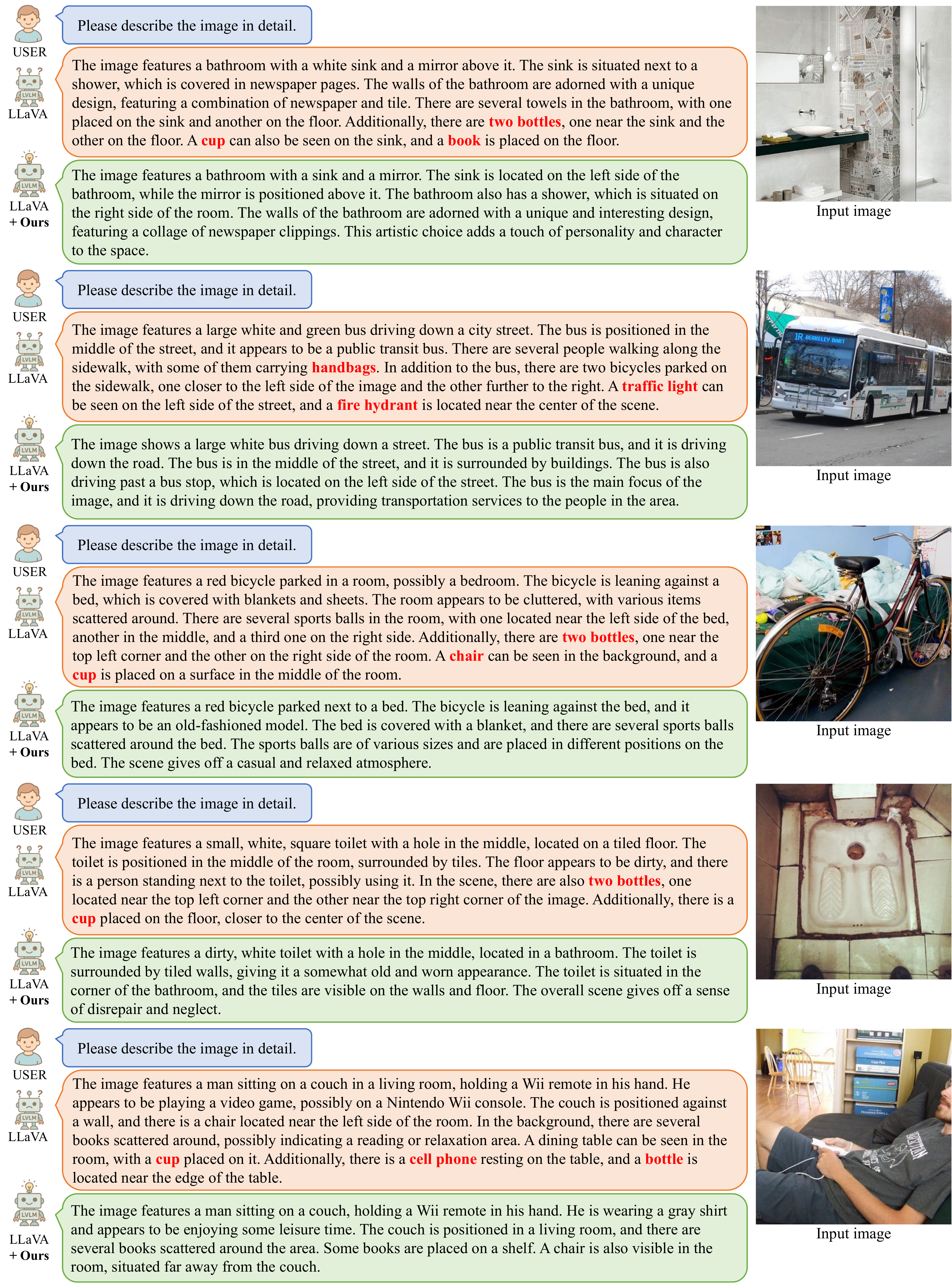}
    \caption{\textbf{Additional qualitative results of our method on LLaVA-1.5-7B with greedy decoding.} On the CHAIR dataset, our method effectively reduces object hallucinations, preventing the model from generating non-existent objects in the image. Hallucinated object words are highlighted in red.}
    \label{append_fig:quali_llava7b_greedy}
\end{figure}

\begin{figure}
    \centering
    \includegraphics[width=\linewidth]{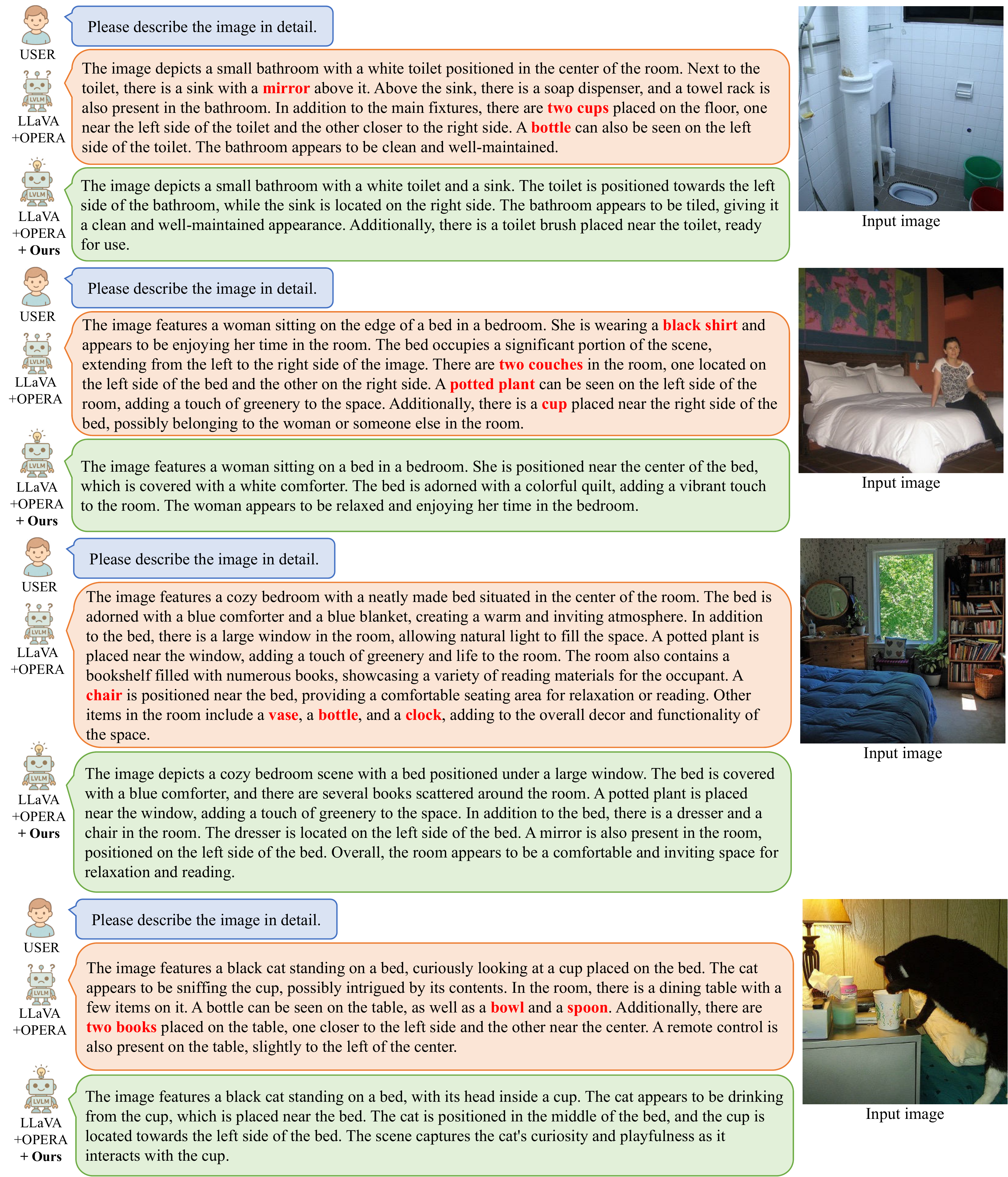}
    \caption{\textbf{Additional qualitative results of our method on LLaVA-1.5-7B with OPERA.} On the CHAIR dataset, our method effectively reduces object hallucinations, preventing the model from generating non-existent objects in the image. Hallucinated object words are highlighted in red.}
    \label{append_fig:quali_llava7b_opera}
\end{figure}

\begin{figure}
    \centering
    \includegraphics[width=\linewidth]{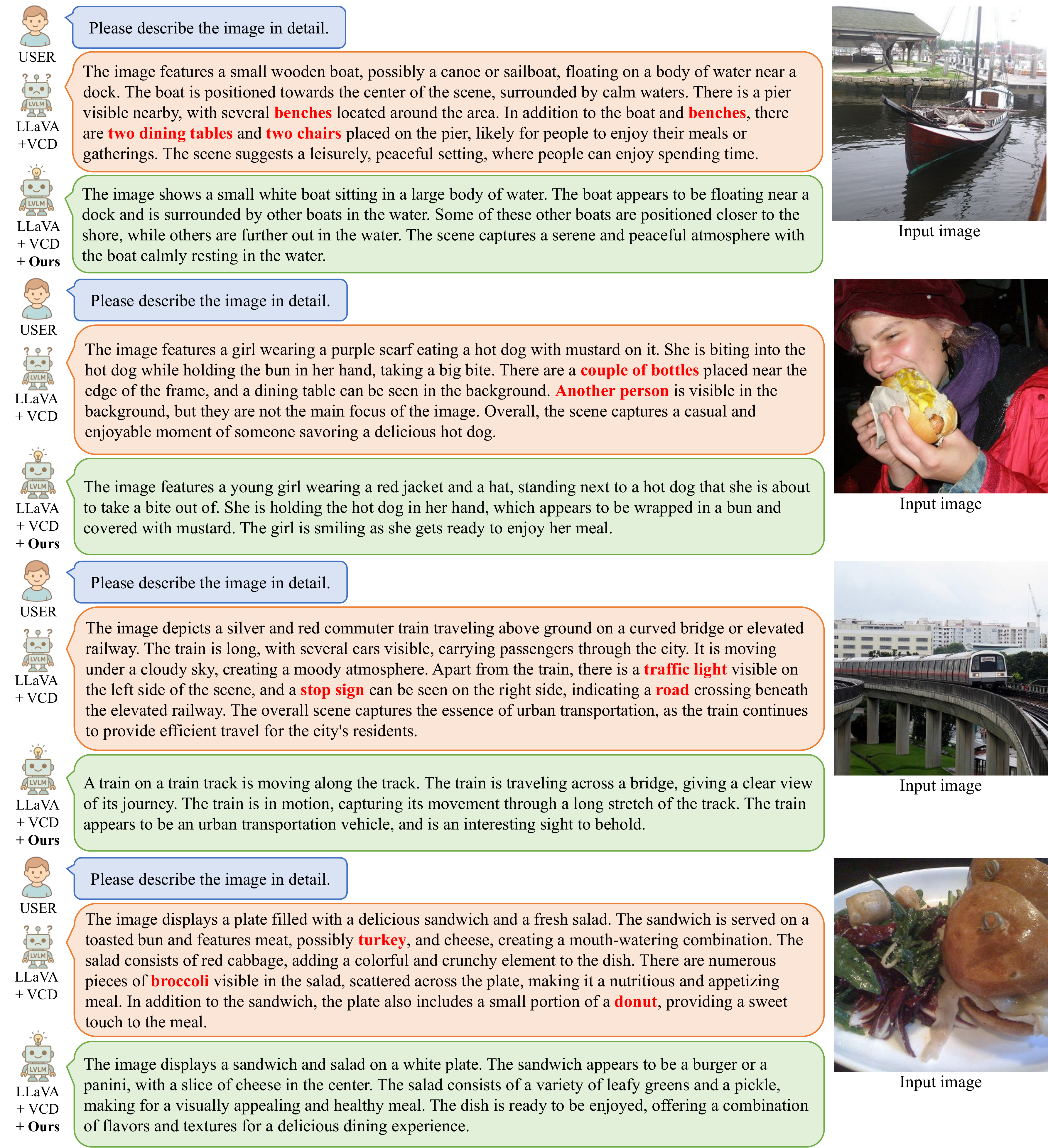}
    \caption{\textbf{Additional qualitative results of our method on LLaVA-1.5-7B with VCD.} On the CHAIR dataset, our method effectively reduces object hallucinations, preventing the model from generating non-existent objects in the image. Hallucinated object words are highlighted in red.}
    \label{append_fig:quali_llava7b_vcd}
\end{figure}

\begin{figure}
    \centering
    \includegraphics[width=\linewidth]{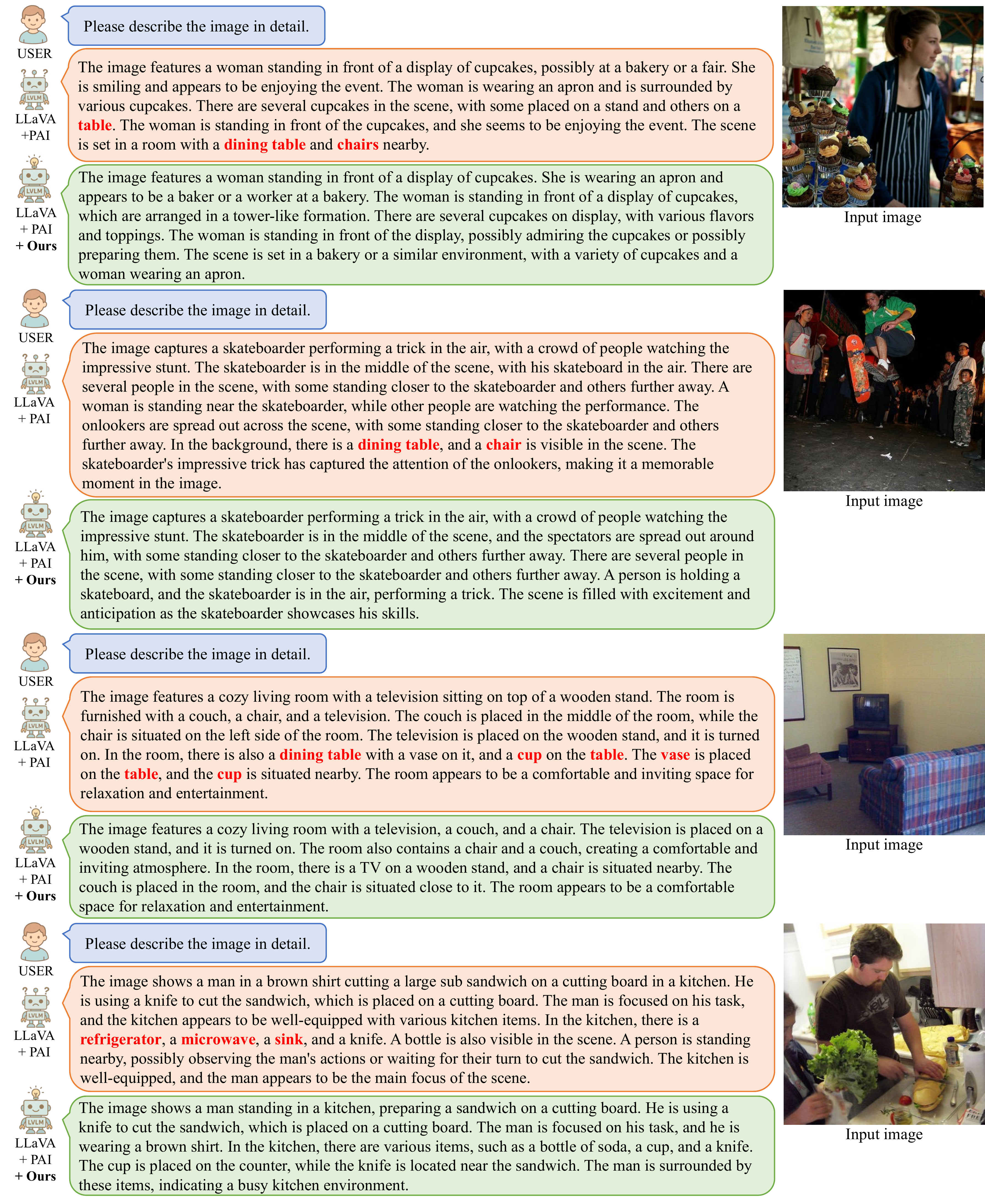}
    \caption{\textbf{Additional qualitative results of our method on LLaVA-1.5-7B with PAI.} On the CHAIR dataset, our method effectively reduces object hallucinations, preventing the model from generating non-existent objects in the image. Hallucinated object words are highlighted in red.}
    \label{append_fig:quali_llava7b_pai}
\end{figure}

\begin{figure}
    \centering
    \includegraphics[width=\linewidth]{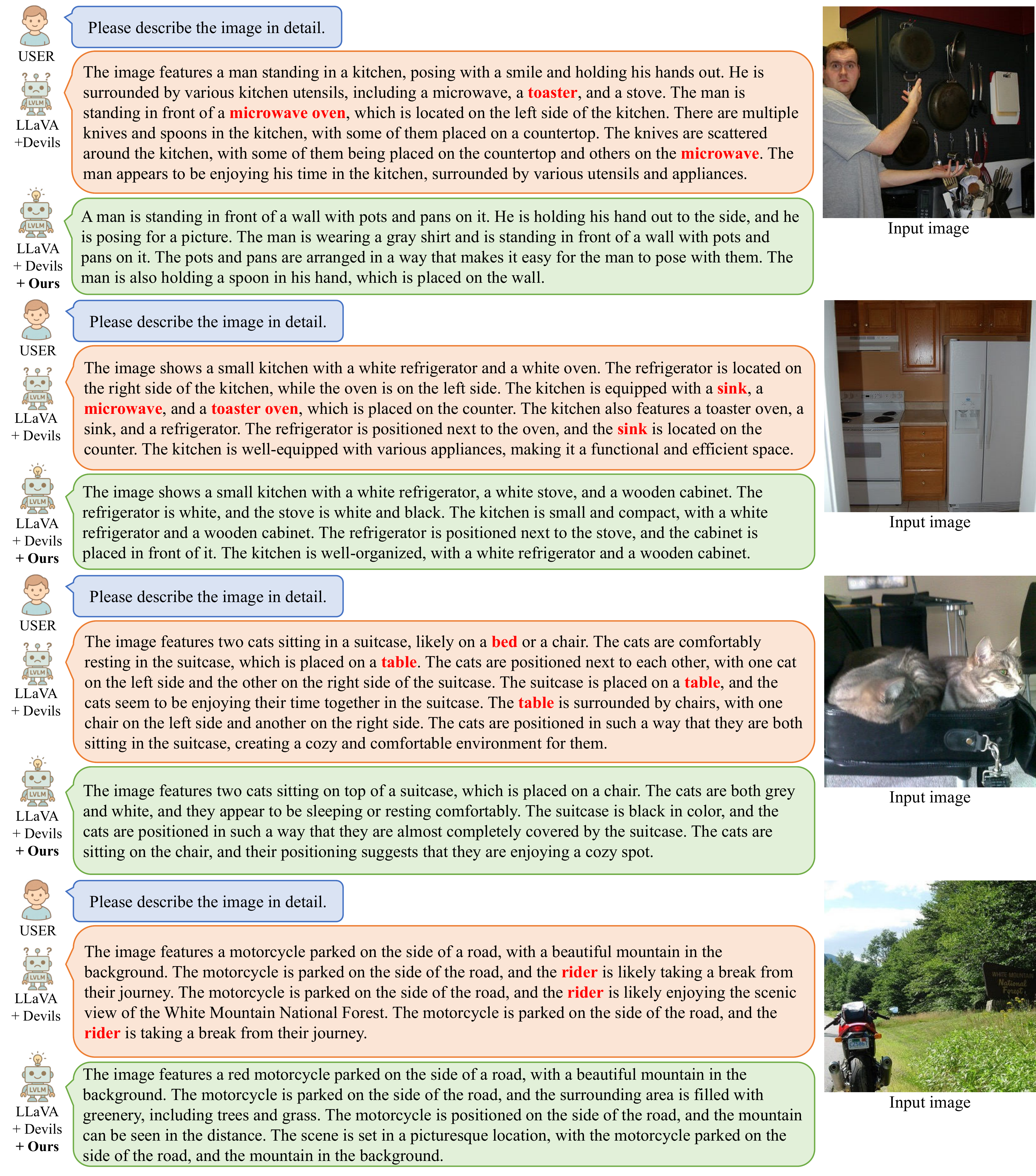}
    \caption{\textbf{Additional qualitative results of our method on LLaVA-1.5-7B with Devils.} On the CHAIR dataset, our method effectively reduces object hallucinations, preventing the model from generating non-existent objects in the image. Hallucinated object words are highlighted in red.}
    \label{append_fig:quali_llava7b_devis}
\end{figure}

\begin{figure}
    \centering
    \includegraphics[width=\linewidth]{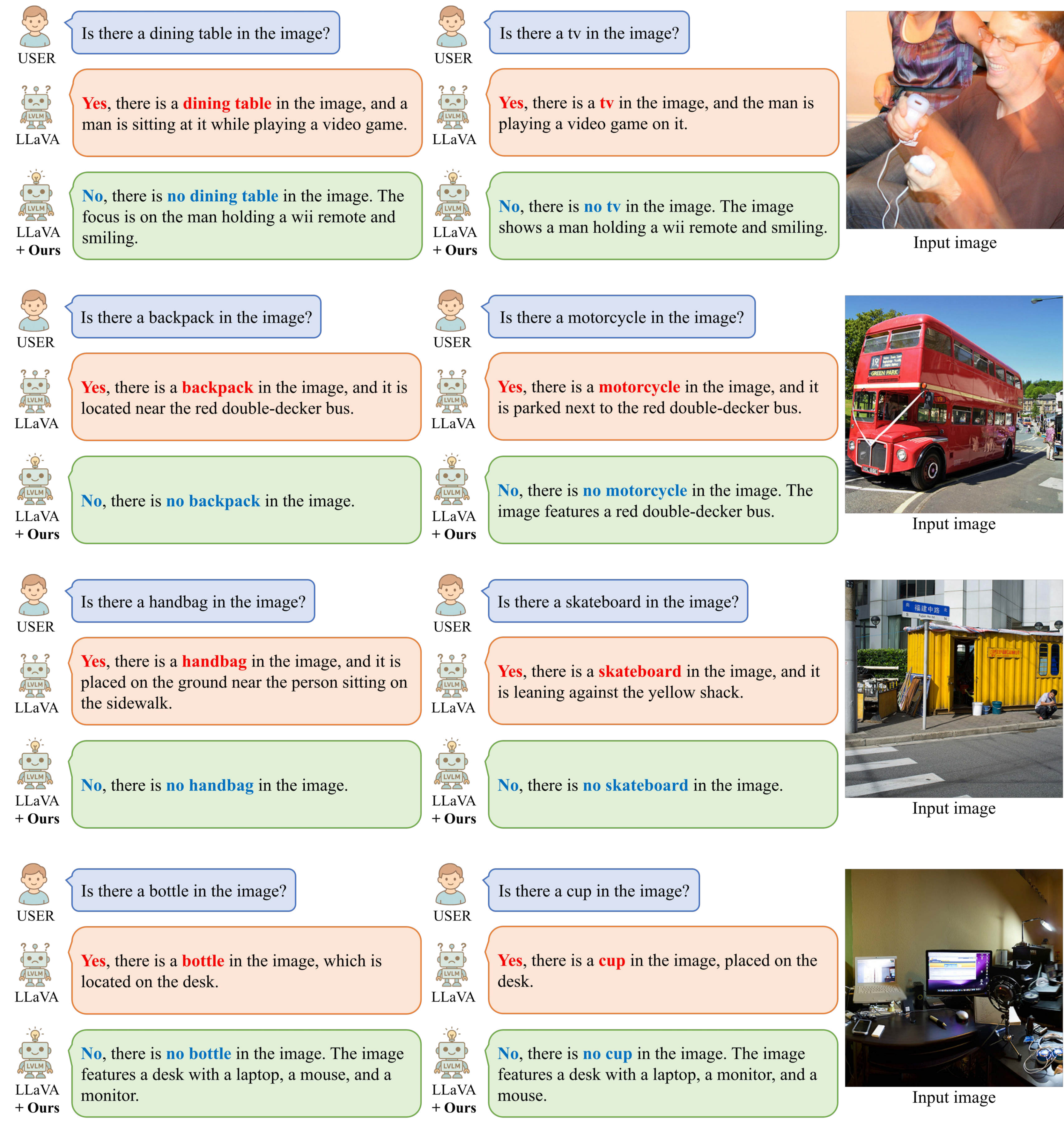}
    \caption{\textbf{Additional qualitative results of our method on LLaVA-1.5-7B with greedy decoding.} On the POPE dataset, our method correctly identifies objects present in the image. Correct and incorrect answers are highlighted in blue and red, respectively.}
    \label{append_fig:quali_llava7b_greedy_pope}
\end{figure}

\begin{figure}
    \centering
    \includegraphics[width=\linewidth]{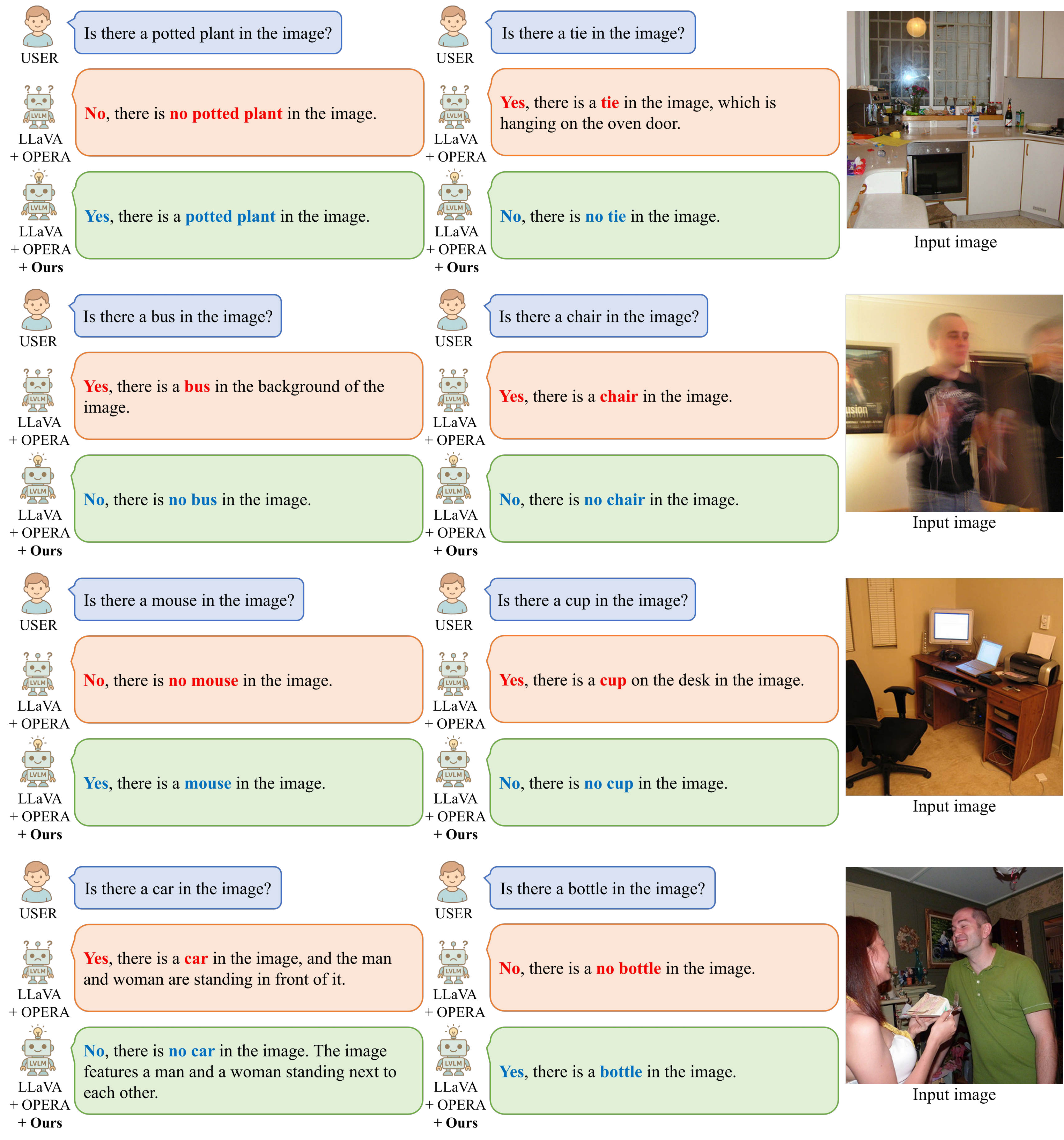}
    \caption{\textbf{Additional qualitative results of our method on LLaVA-1.5-7B with OPERA.} On the POPE dataset, our method correctly identifies objects present in the image. Correct and incorrect answers are highlighted in blue and red, respectively.}
    \label{append_fig:quali_llava7b_opera_pope}
\end{figure}

\begin{figure}
    \centering
    \includegraphics[width=\linewidth]{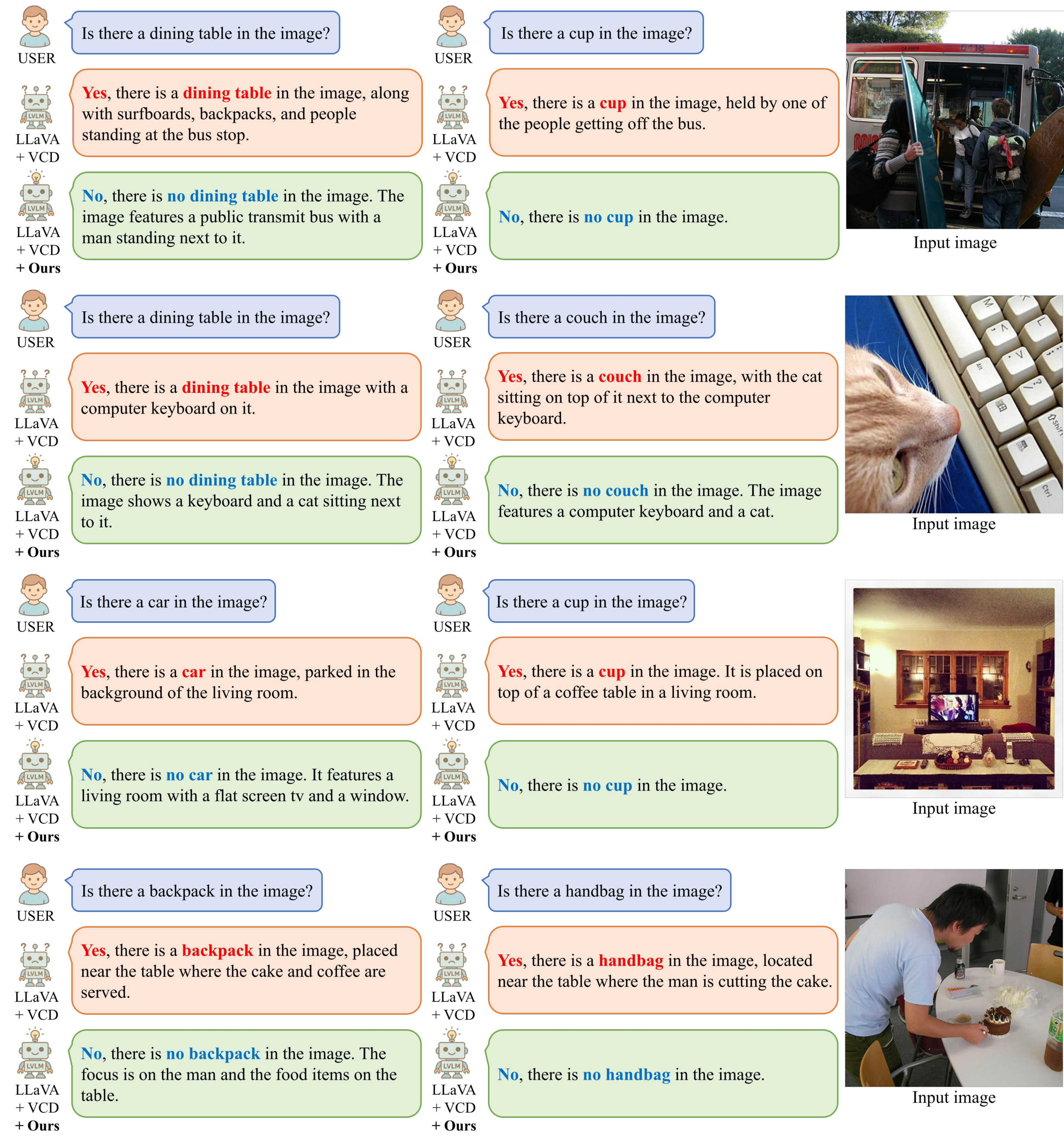}
    \caption{\textbf{Additional qualitative results of our method on LLaVA-1.5-7B with VCD.} On the POPE dataset, our method correctly identifies objects present in the image. Correct and incorrect answers are highlighted in blue and red, respectively.}
    \label{append_fig:quali_llava7b_vcd_pope}
\end{figure}

\begin{figure}
    \centering
    \includegraphics[width=\linewidth]{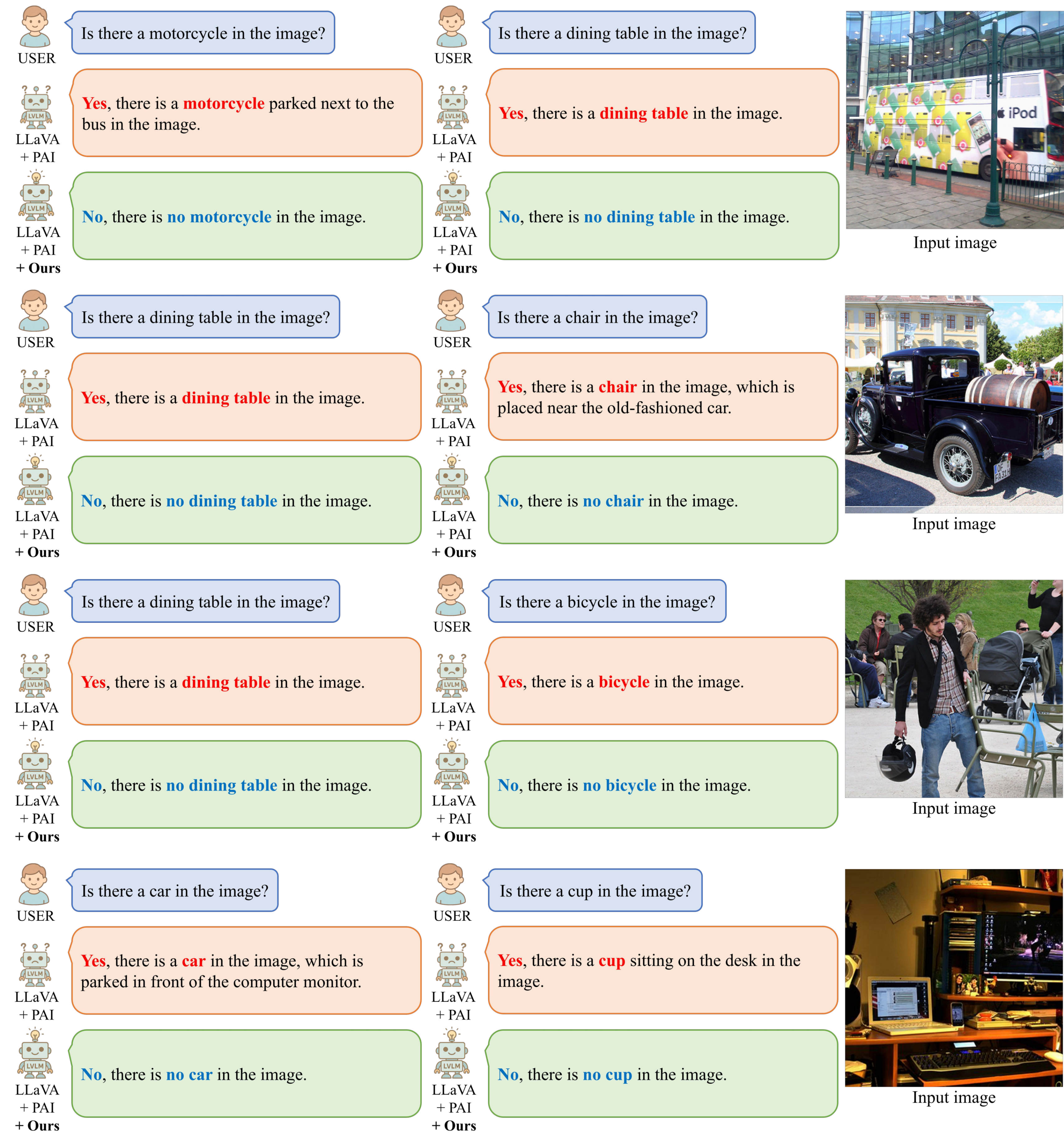}
    \caption{\textbf{Additional qualitative results of our method on LLaVA-1.5-7B with PAI.} On the POPE dataset, our method correctly identifies objects present in the image. Correct and incorrect answers are highlighted in blue and red, respectively.}
    \label{append_fig:quali_llava7b_pai_pope}
\end{figure}

\begin{figure}
    \centering
    \includegraphics[width=\linewidth]{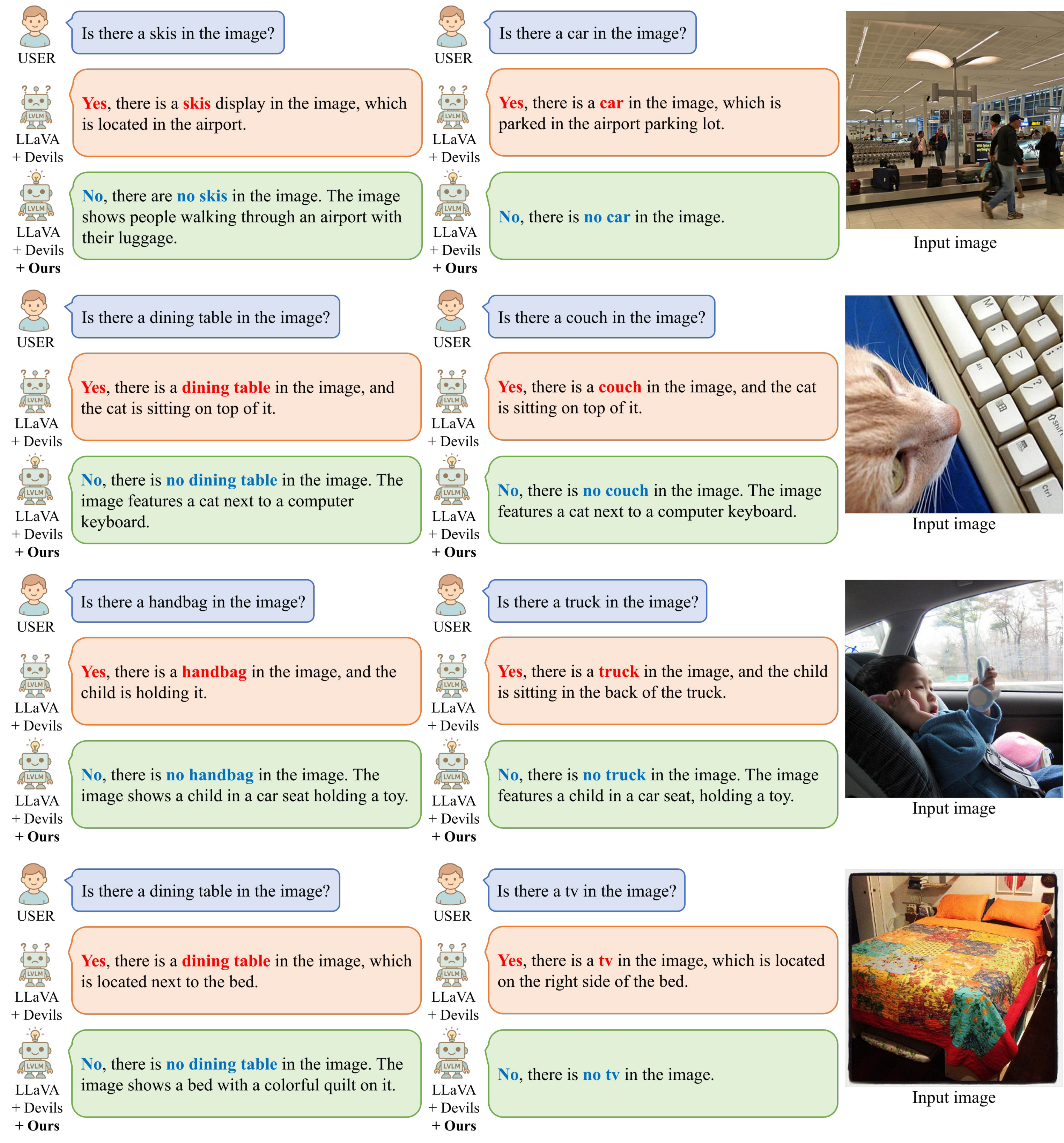}
    \caption{\textbf{Additional qualitative results of our method on LLaVA-1.5-7B with Devils.} On the POPE dataset, our method correctly identifies objects present in the image. Correct and incorrect answers are highlighted in blue and red, respectively.}
    \label{append_fig:quali_llava7b_devils_pope}
\end{figure}

\begin{figure}
    \centering
    \includegraphics[width=\linewidth]{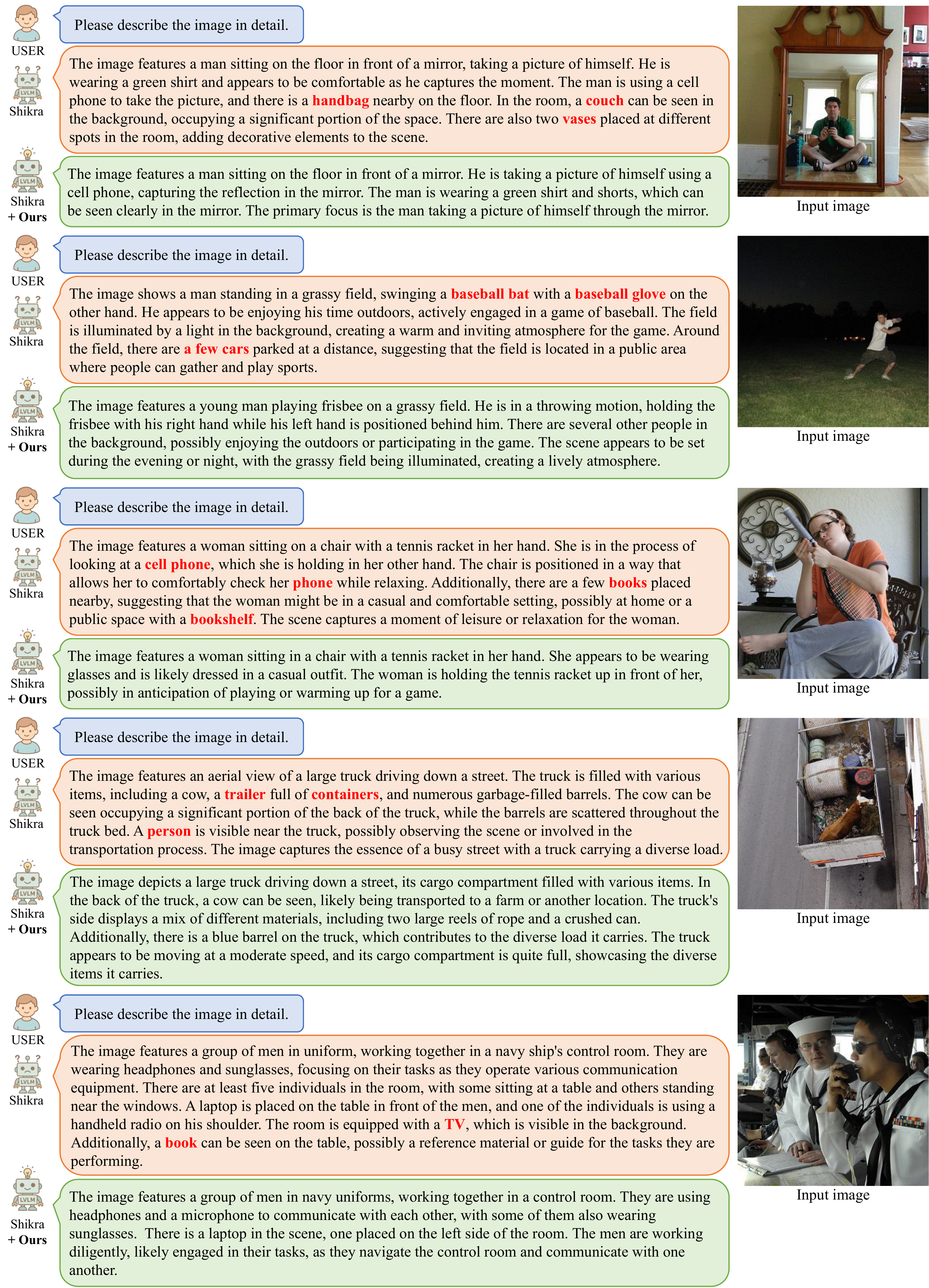}
    \caption{\textbf{Additional qualitative results of our method on Shikra-7B with greedy decoding.} On the CHAIR dataset, our method effectively reduces object hallucinations, preventing the model from generating non-existent objects in the image. Hallucinated object words are highlighted in red.}
    \label{append_fig:quali_shikra_greedy}
\end{figure}

\begin{figure}
    \centering
    \includegraphics[width=\linewidth]{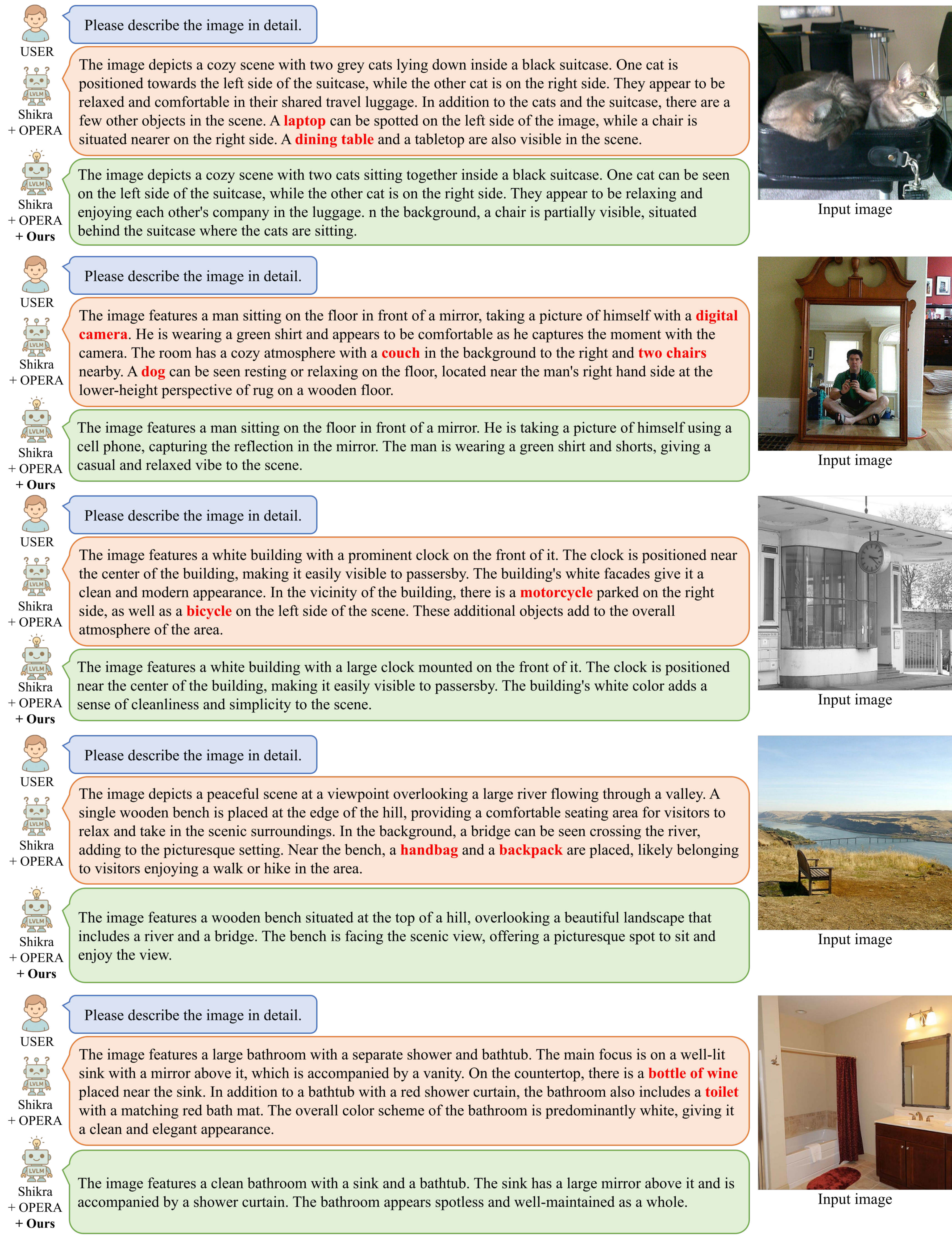}
    \caption{\textbf{Additional qualitative results of our method on Shikra-7B with OPERA.} On the CHAIR dataset, our method effectively reduces object hallucinations, preventing the model from generating non-existent objects in the image. Hallucinated object words are highlighted in red.}
    \label{append_fig:quali_shikra_opera}
\end{figure}

\begin{figure}
    \centering
    \includegraphics[width=\linewidth]{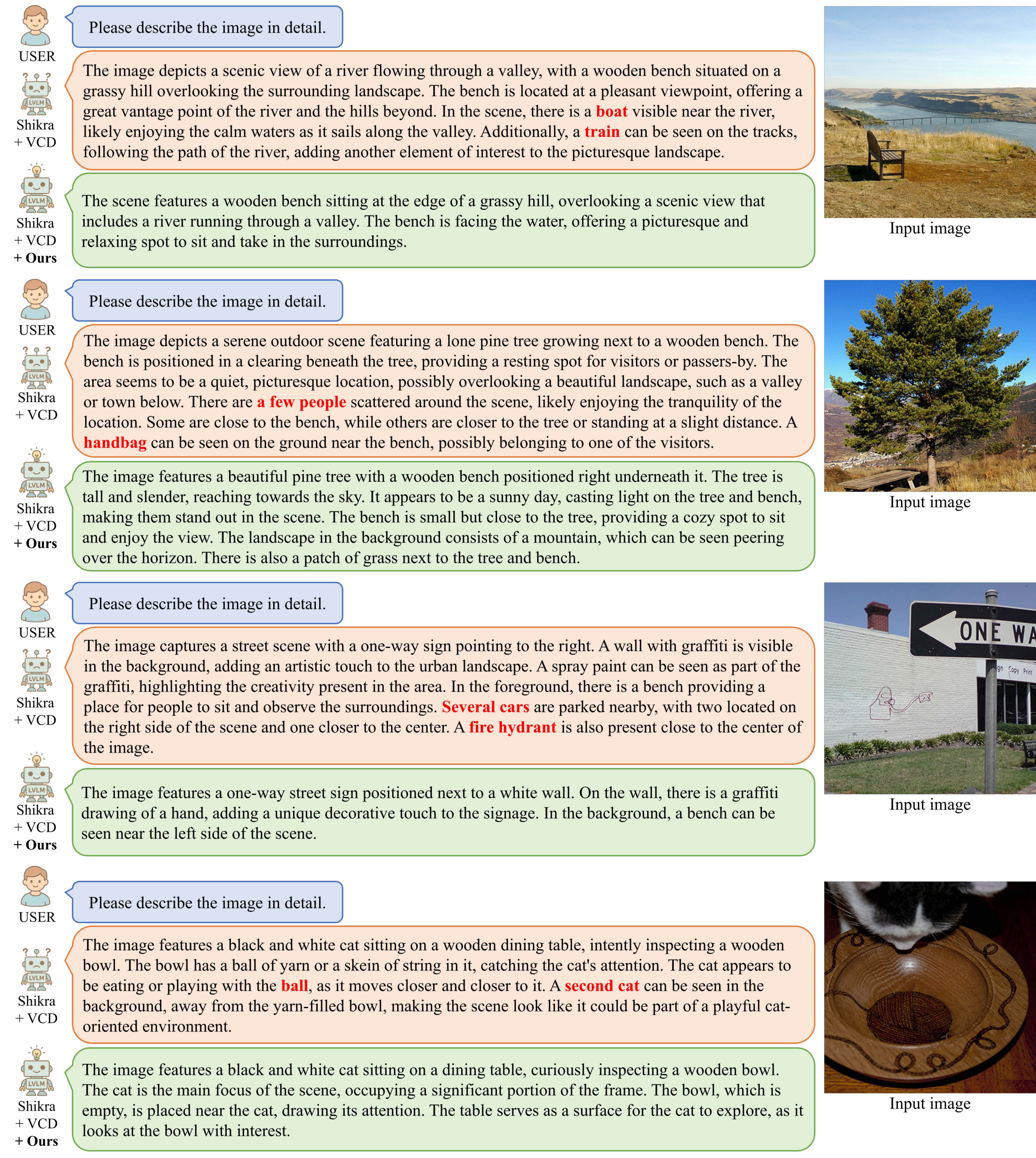}
    \caption{\textbf{Additional qualitative results of our method on Shikra-7B with VCD.} On the CHAIR dataset, our method effectively reduces object hallucinations, preventing the model from generating non-existent objects in the image. Hallucinated object words are highlighted in red.}
    \label{append_fig:quali_shikra_vcd}
\end{figure}

\begin{figure}
    \centering
    \includegraphics[width=\linewidth]{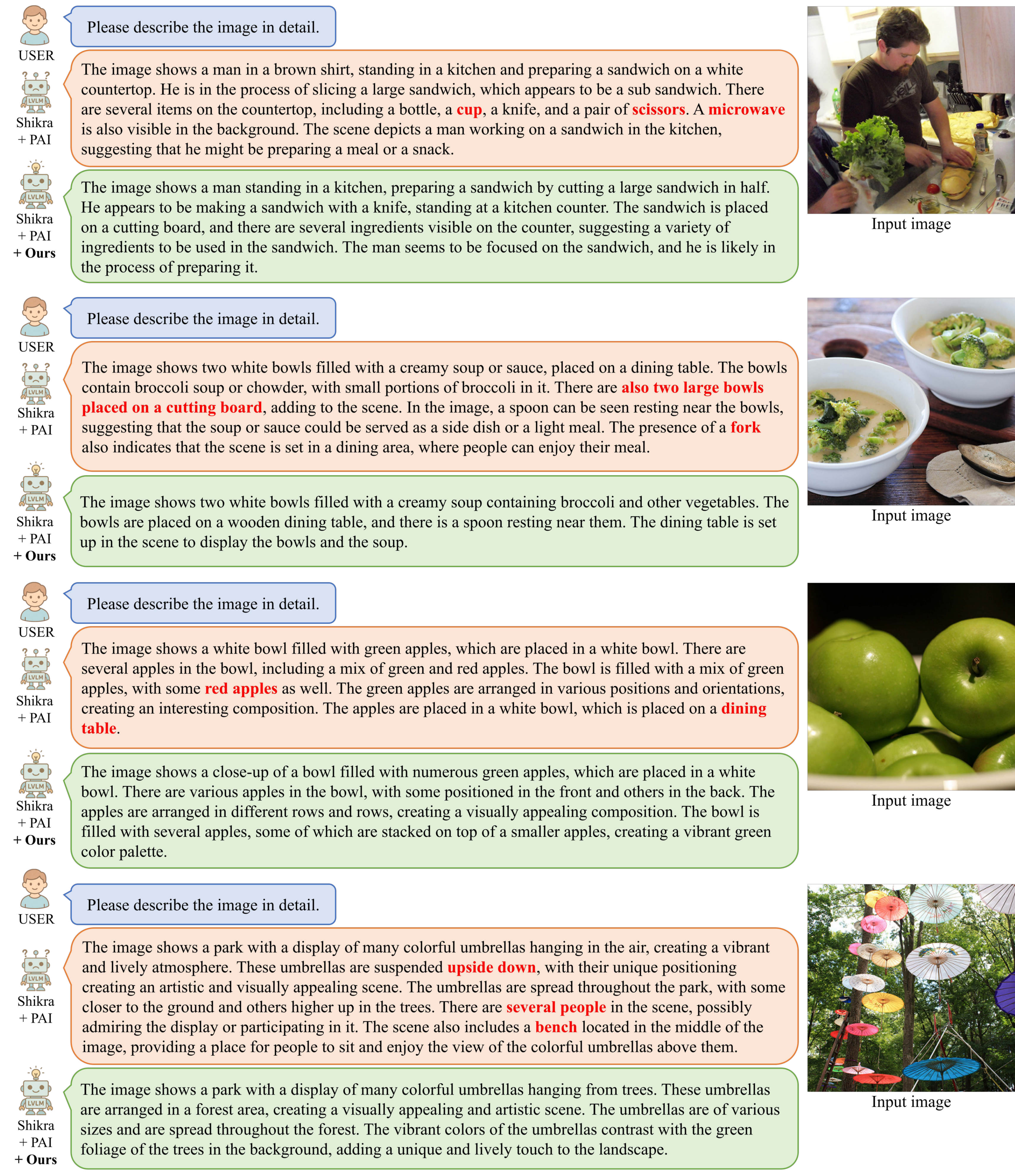}
    \caption{\textbf{Additional qualitative results of our method on Shikra-7B with PAI.} On the CHAIR dataset, our method effectively reduces object hallucinations, preventing the model from generating non-existent objects in the image. Hallucinated object words are highlighted in red.}
    \label{append_fig:quali_shikra_pai}
\end{figure}

\begin{figure}
    \centering
    \includegraphics[width=\linewidth]{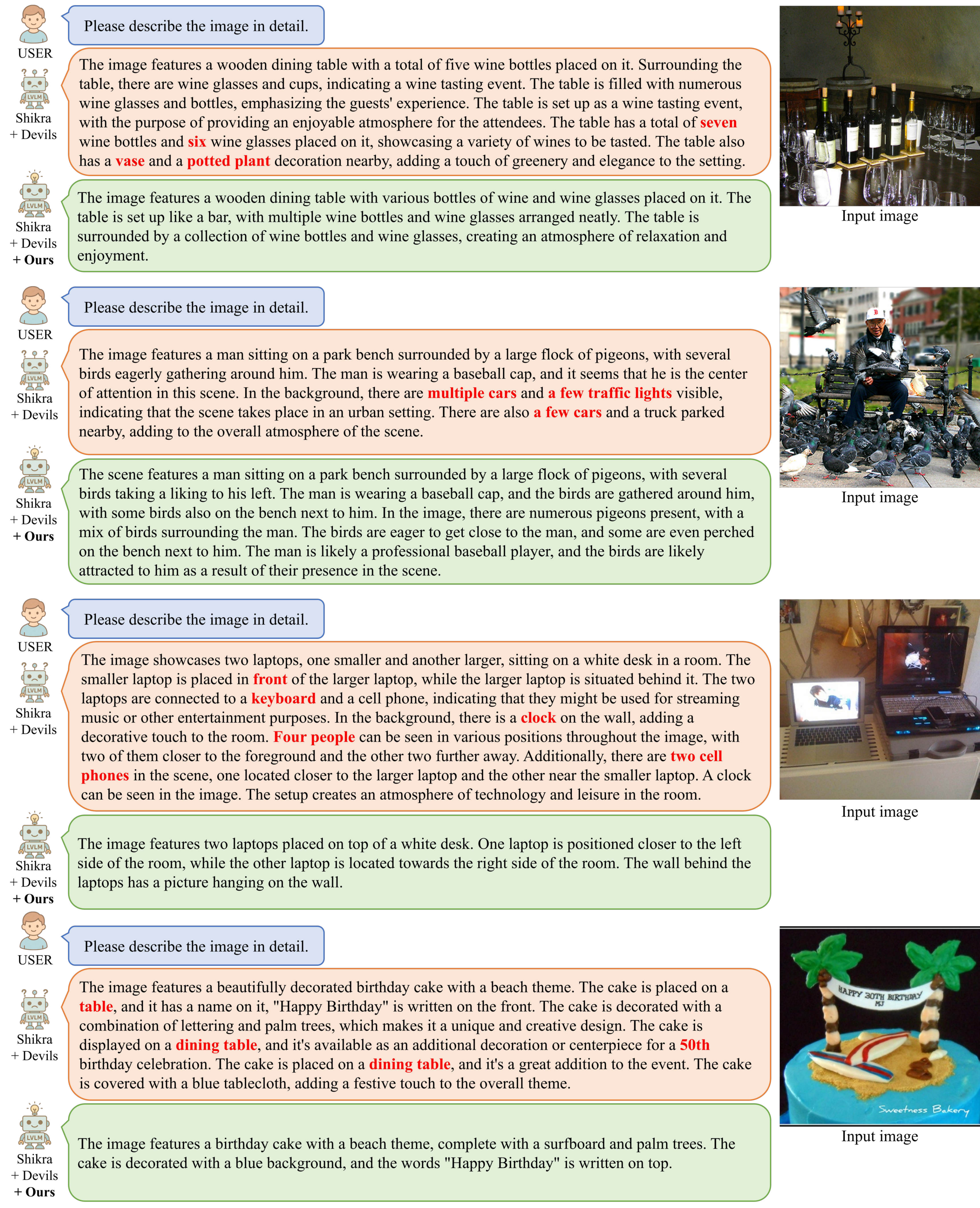}
    \caption{\textbf{Additional qualitative results of our method on Shikra-7B with Devils.} On the CHAIR dataset, our method effectively reduces object hallucinations, preventing the model from generating non-existent objects in the image. Hallucinated object words are highlighted in red.}
    \label{append_fig:quali_shikra_devils}
\end{figure}

\begin{figure}
    \centering
    \includegraphics[width=\linewidth]{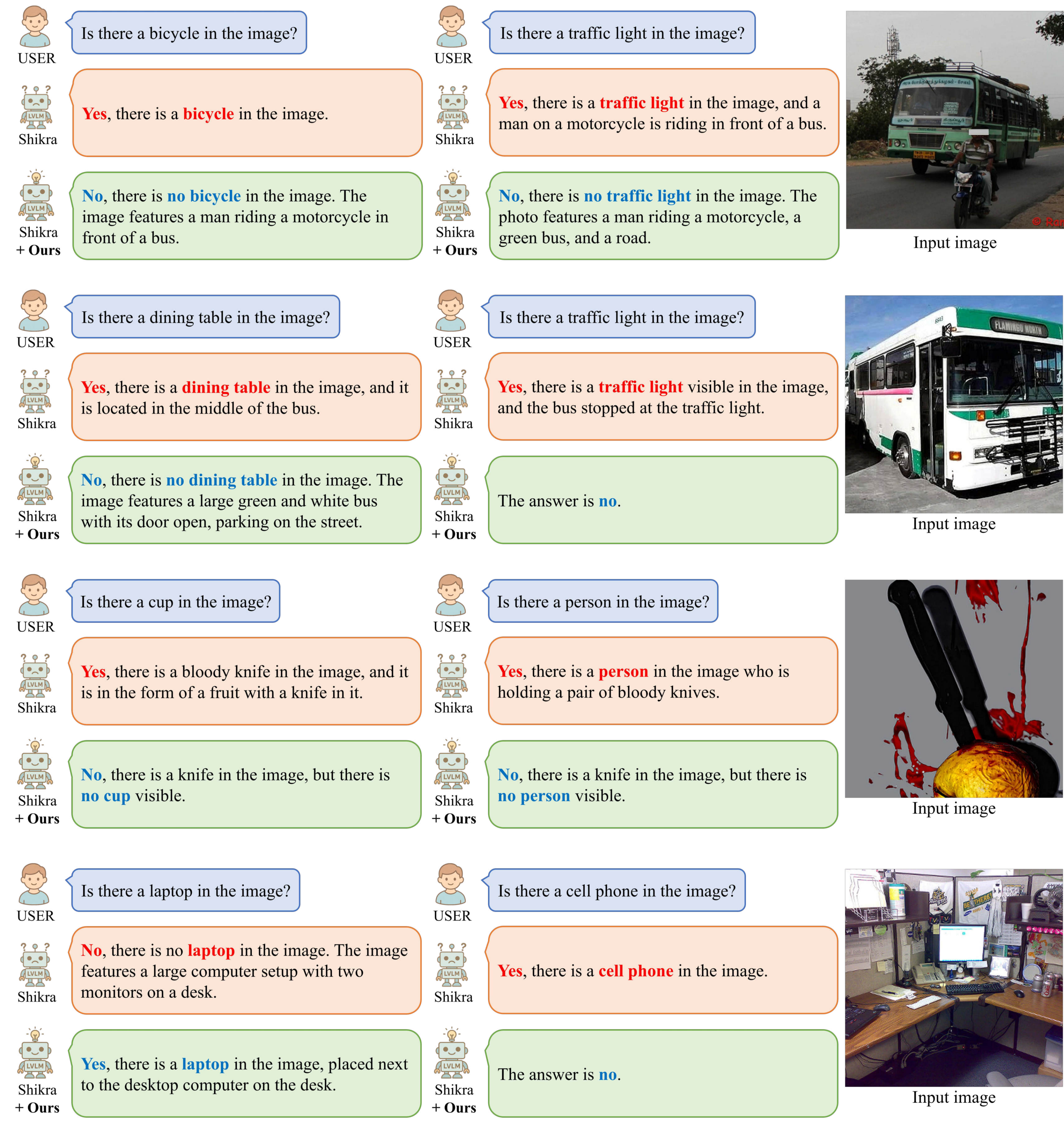}
    \caption{\textbf{Additional qualitative results of our method on Shikra-7B with greedy decoding.} On the POPE dataset, our method correctly identifies objects present in the image. Correct and incorrect answers are highlighted in blue and red, respectively.}
    \label{append_fig:quali_shikra_greedy_pope}
\end{figure}

\begin{figure}
    \centering
    \includegraphics[width=\linewidth]{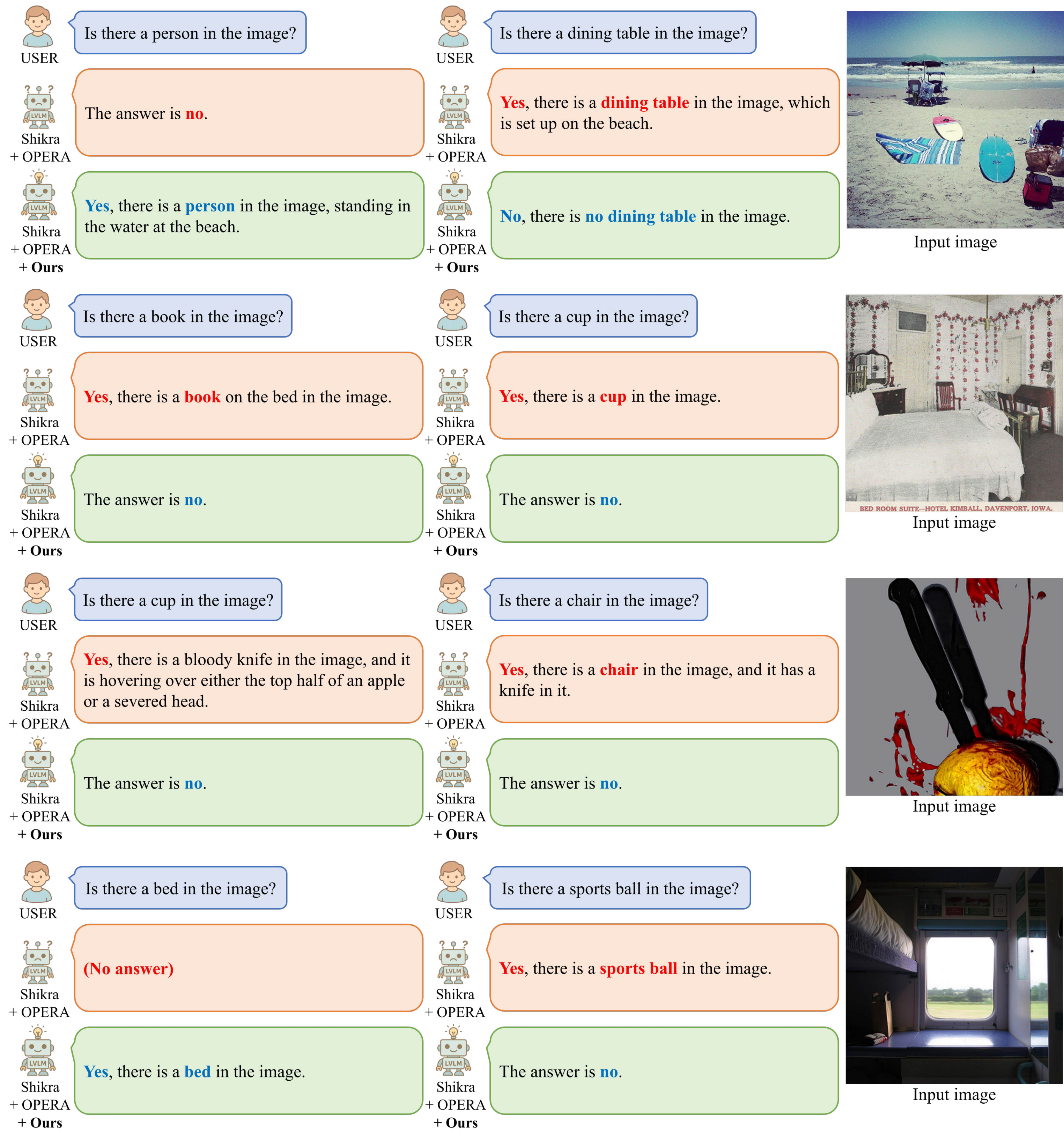}
    \caption{\textbf{Additional qualitative results of our method on Shikra-7B with OPERA.} On the POPE dataset, our method correctly identifies objects present in the image. Correct and incorrect answers are highlighted in blue and red, respectively.}
    \label{append_fig:quali_shikra_opera_pope}
\end{figure}

\begin{figure}
    \centering
    \includegraphics[width=\linewidth]{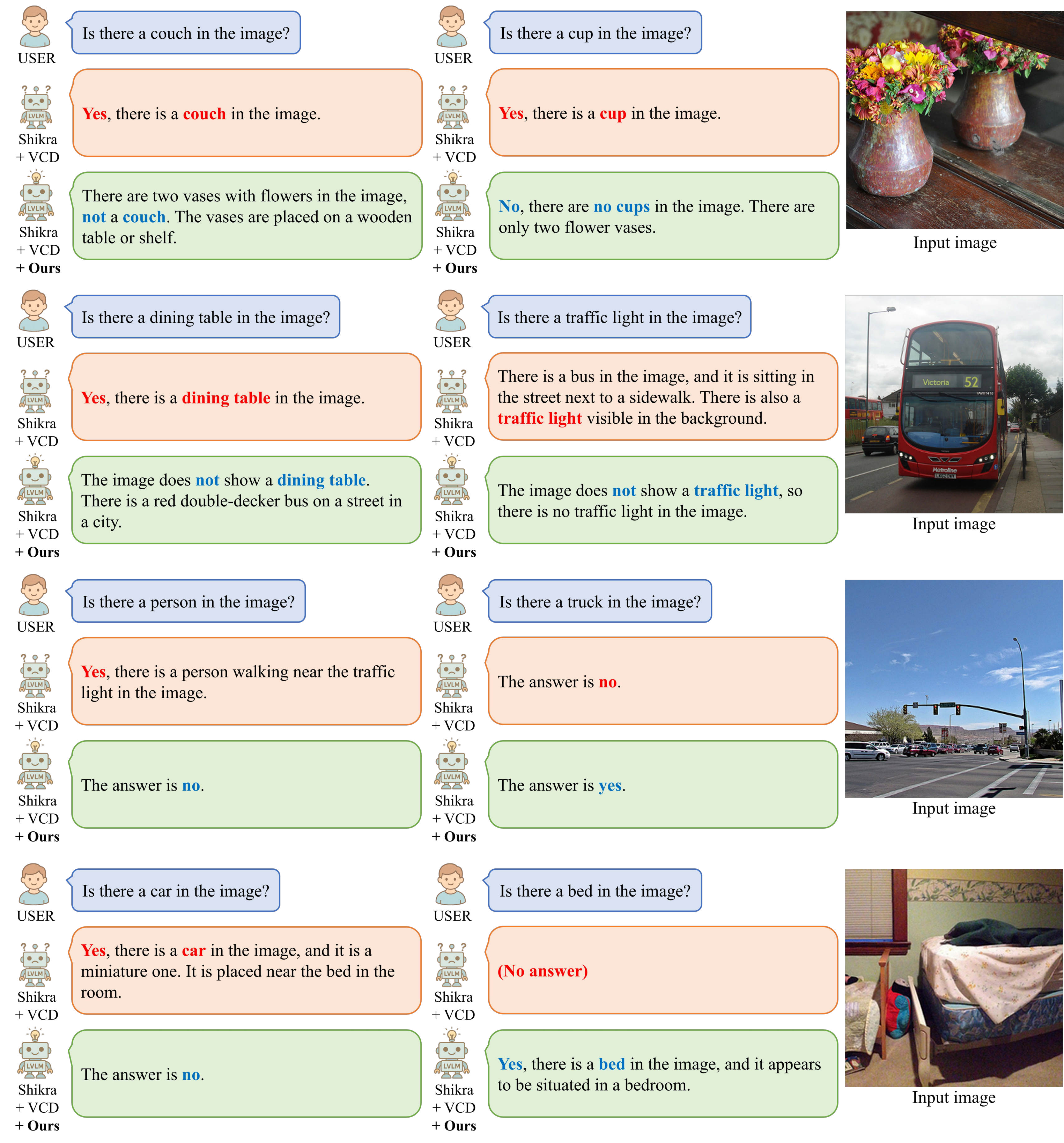}
    \caption{\textbf{Additional qualitative results of our method on Shikra-7B with VCD.} On the POPE dataset, our method correctly identifies objects present in the image. Correct and incorrect answers are highlighted in blue and red, respectively.}
    \label{append_fig:quali_shikra_vcd_pope}
\end{figure}

\begin{figure}
    \centering
    \includegraphics[width=\linewidth]{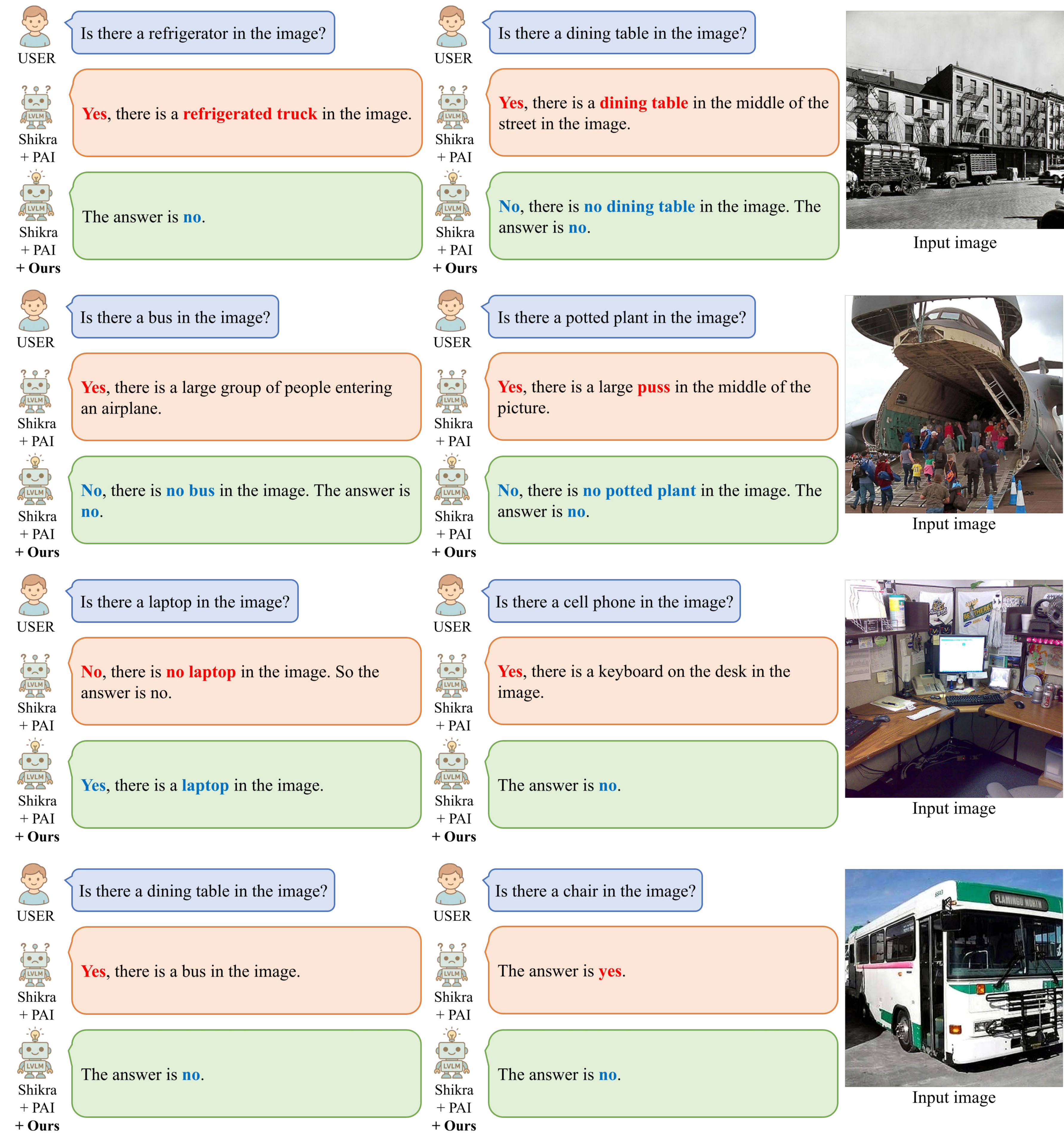}
    \caption{\textbf{Additional qualitative results of our method on Shikra-7B with PAI.} On the POPE dataset, our method correctly identifies objects present in the image. Correct and incorrect answers are highlighted in blue and red, respectively.}
    \label{append_fig:quali_shikra_pai_pope}
\end{figure}

\begin{figure}
    \centering
    \includegraphics[width=\linewidth]{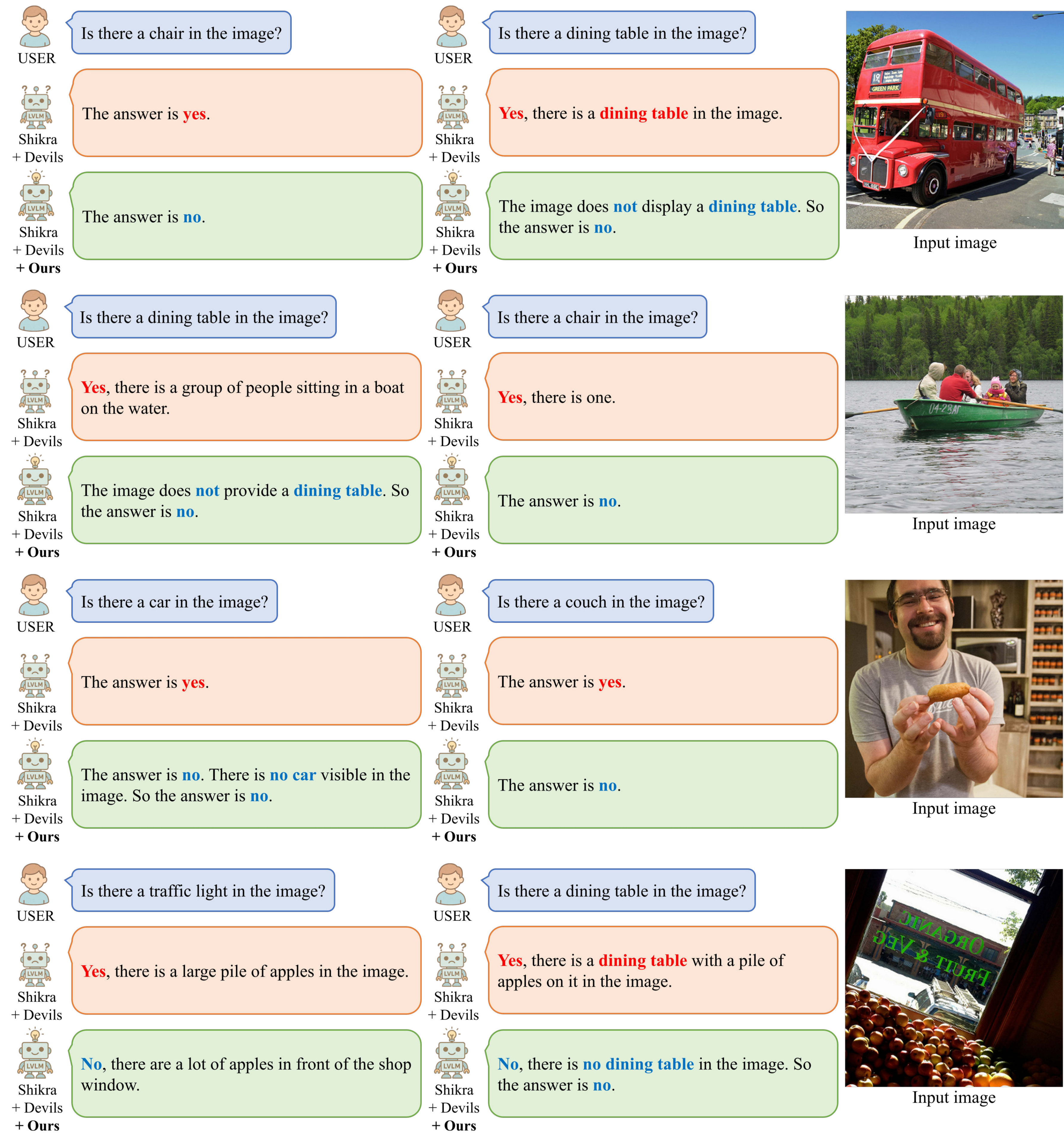}
    \caption{\textbf{Additional qualitative results of our method on Shikra-7B with Devils.} On the POPE dataset, our method correctly identifies objects present in the image. Correct and incorrect answers are highlighted in blue and red, respectively.}
    \label{append_fig:quali_shikra_devils_pope}
\end{figure}

\paragraph{Qualitative examples of our method on various LVLMs with different mitigation methods.}
We present additional qualitative examples of our method applied to different combinations of LVLMs (LLaVA-1.5-7B and Shikra-7B) and hallucination mitigation techniques, including greedy decoding, OPERA, VCD, PAI, and Devils. Our method integrates well with these approaches and effectively reduces object hallucinations by preventing the generation of non-existent objects. 
Fig.~\ref{append_fig:quali_llava7b_greedy}–\ref{append_fig:quali_shikra_devils_pope} illustrate qualitative examples on the CHAIR and POPE datasets using LLaVA-1.5-7B and Shikra-7B across various hallucination mitigation methods.

\begin{figure}
    \centering
    \includegraphics[width=\linewidth]{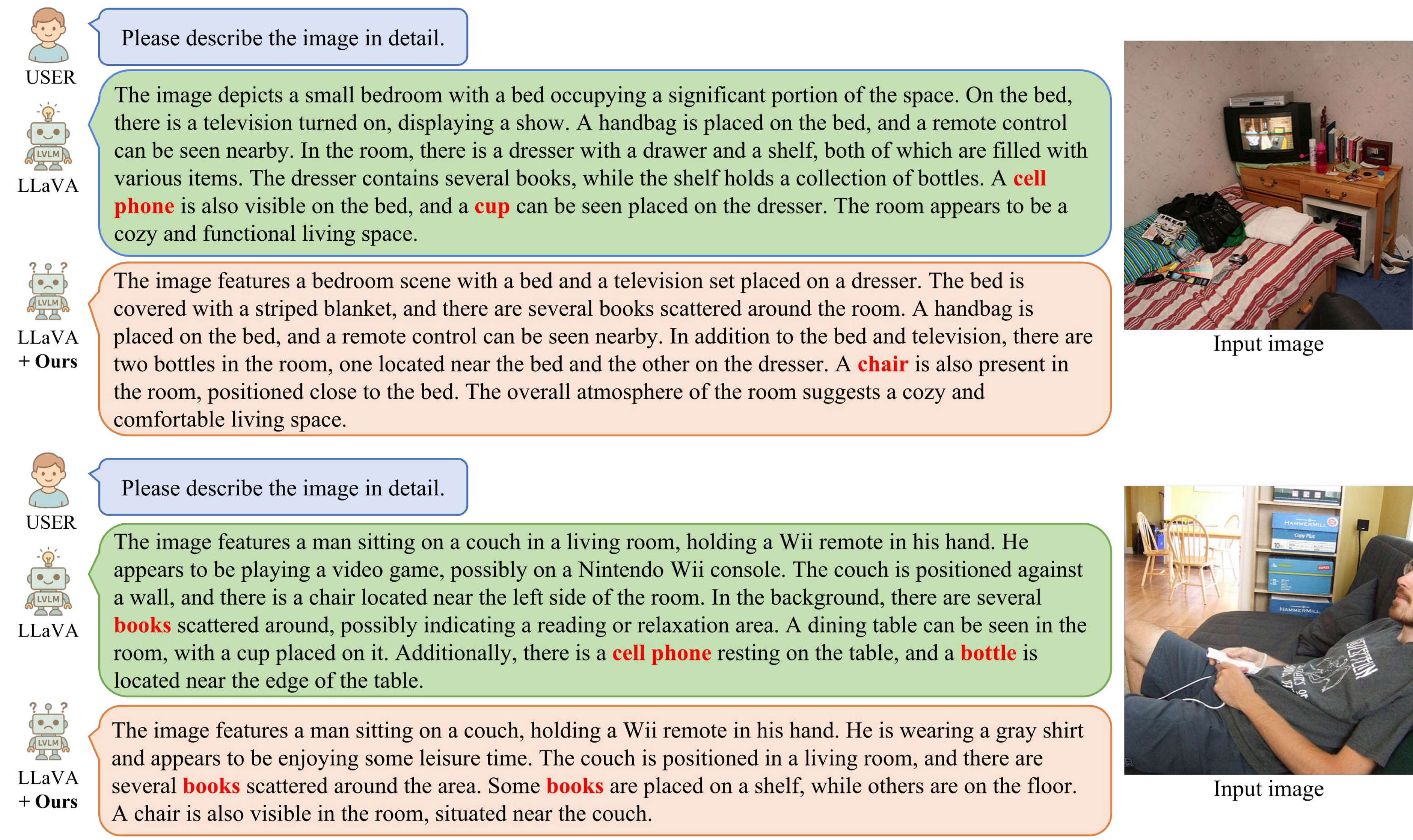}
    \caption{\textbf{Failure cases of our method on LLaVA-1.5-7B with greedy decoding.} On the CHAIR dataset, our method effectively reduces object hallucinations but fails to completely prevent the generation of non-existent objects. Hallucinated object words are highlighted in red.}
    \label{append_fig:llava7b_failure}
\end{figure}

\paragraph{Qualitative examples of failure cases.}
Fig.~\ref{append_fig:llava7b_failure} presents qualitative examples of failure cases from our proposed method. Although our method consistently mitigates hallucinated responses, it occasionally fails to prevent all hallucinations.

\section{Discussion}
We statistically demonstrate that epistemic uncertainty within the vision encoder contributes to object hallucination and address this issue through self-attention masking at intermediate layers. To understand how LVLMs change their integration of visual information after applying our method, we measured the entropy of the LLM’s attention distribution over image tokens across all layers and heads. Entropy serves as an indicator of whether the model attends broadly or narrowly, with higher entropy reflecting the use of a wider range of visual evidence rather than reliance on a small subset of tokens. Using 500 images, we found that the average entropy of LLaVA increased from 1.5746 in the original model to 1.9717 with our method. This increase suggests that our approach encourages broader and more balanced attention over reliable visual tokens, enabling the model to integrate visual information more effectively while reducing over-reliance on uncertain inputs, consistent with findings from prior work~\cite{liu2024paying}.

\section{Broader Impacts}\label{append_broader}
We proposed a method to improve the reliability of Large Vision-Language Models (LVLMs) by identifying and masking uncertain visual tokens in the vision encoder, a key source of object hallucination. In contrast to existing approaches that intervene at the language model level, our method operates solely on the vision encoder and demonstrates effectiveness across a variety of models and settings.

Our method offers significant societal benefits by improving safety and reliability in critical applications such as medical imaging, assistive technologies, and autonomous systems. However, it may also inadvertently suppress valid but ambiguous visual information, which could disproportionately affect underrepresented groups and reinforce existing dataset biases, raising important concerns about potential negative societal impacts.

\section{Limitations}\label{append_limitation}
Despite its advantages, our method has several limitations. First, while it effectively reduces hallucinations, the proposed masking strategy within self-attention process of vision encoder may result in the loss of visual information, leading to slight performance degradation in certain perception-oriented tasks. Second, the reliance on PGD-based adversarial attacks to estimate uncertainty increases inference time. Third, the masking strategy lacks a formal theoretical foundation, instead relying on a trace-based approximation of uncertainty as a surrogate for the determinant of the covariance matrix. Finally, the method is less effective for models like MiniGPT-4, which utilize a Q-Former to abstract visual information before integrating it with language. In these cases, masking at the vision encoder has limited downstream impact. Addressing these issues is a promising direction for future research.

\clearpage

\end{document}